\documentclass[a4paper,fleqn]{cas-dc}

\usepackage[numbers]{natbib}

\def\tsc#1{\csdef{#1}{\textsc{\lowercase{#1}}\xspace}}
\tsc{WGM}
\tsc{QE}

\usepackage{algorithm}%
\usepackage{algorithmicx}%
\usepackage{algpseudocode}%

\usepackage[inline]{enumitem}

\definecolor{pblue}{RGB}{75, 139, 190}%
\definecolor{pyellow}{RGB}{204, 163, 0}%

\usepackage{dsfont}

\DeclareMathOperator*{\argmin}{arg\,min}
\DeclareMathOperator*{\argmax}{arg\,max}

\newcommand{\Ball}{\text{Ball}}

\newcommand{\TU}{\mathit{TU}}
\newcommand{\AU}{\mathit{AU}}
\newcommand{\EU}{\mathit{EU}}
\newcommand{\ENT}{\mathit{ENT}}
\newcommand{\AUi}[1]{\mathit{AU_{\mathnormal{#1}}}}
\newcommand{\EUi}[1]{\mathit{EU_{\mathnormal{#1}}}}

\newcommand{\cpos}{{\checkmark}}%
\newcommand{\cneg}{{\mathds{\times}}}%

\newcommand{\xCF}{x^{\cpos}}
\newcommand{\xF}{x^{\cneg}}

\newcommand{\cplus}{{c^\cpos}}%
\newcommand{\yCF}{\cplus}%
\newcommand{\cminus}{{c^\cneg}}%
\newcommand{\yF}{\cminus}%

\newcommand{\CF}{\mathit{CF}}

\newcommand{\MAD}{\mathit{MAD}}

\newcommand{\mCF}{m_{\CF}}%

\usepackage{subcaption}
\graphicspath{{./fig/}}

\usepackage[capitalize,nameinlink]{cleveref}
\crefname{figure}{Figure}{Figures}
\Crefname{equation}{Equation}{Equations}
\Crefname{item}{Item}{Items}
\crefname{item}{Item}{Items}
\crefname{enumi}{case}{cases}%
\Crefname{enumi}{case}{cases}%

\crefname{appendix}{Appendix}{Appendices}
\Crefname{appendix}{Appendix}{Appendices}

\newcommand{\sref}[1]{\hyperref[#1]{\S\ref{#1}}}

\newcommand{\github}{\href{https://github.com/Advueu963/Uncertainty-aware-Counterfactuals}{github.com/Advueu963/Uncertainty-aware-Counterfactuals}}

\makeatletter
\@afterindentfalse
\let\@afterindenttrue\@afterindentfalse
\makeatother

\usepackage{afterpage}
\usepackage{placeins}

\begin{document}
\let\WriteBookmarks\relax
\def\floatpagepagefraction{1}
\def\textpagefraction{.001}

\shorttitle{Uncertainty-aware Explainable Artificial Intelligence}    

\shortauthors{K.\ Sokol, S.M.A.R.\ Thies and E.\ H{\"u}llermeier}%

\title[mode = title]{%
Uncertainty Quantification as a Principled Foundation for Explainable Artificial Intelligence: %
A Case Study of Counterfactual Explanations%
}

\author[1,2,3]{Kacper Sokol}[orcid=0000-0002-9869-5896]
\cormark[1]

\ead{kacper.sokol@usi.ch}

\credit{Conceptualisation, Methodology, Investigation, Formal Analysis, Visualisation, Writing -- Original Draft, Writing -- Review \& Editing, Supervision, Funding Acquisition}

\affiliation[1]{organization={Department of Informatics, USI Lugano},
            country={Switzerland}}

\affiliation[2]{organization={Department of Computer Science, ETH Zurich},
            country={Switzerland}}

\affiliation[3]{organization={Institute of Informatics, LMU Munich},
            country={Germany}}

\author[2]{Santo M.A.R. Thies}[orcid=0009-0006-4163-6699]%
\ead{s.thies@campus.lmu.de}

\credit{Methodology, Investigation, Formal Analysis, Visualisation, Software, Writing -- Original Draft, Writing -- Review \& Editing}

\author[2]{Eyke H{\"u}llermeier}[orcid=0000-0002-9944-4108]%
\ead{eyke@lmu.de}

\credit{Writing -- Review \& Editing, Supervision}

\cortext[1]{Corresponding author}

\begin{abstract}
In this paper we argue %
that, to its detriment, transparency research overlooks many foundational concepts of artificial intelligence. %
As an illustrating example %
we focus on uncertainty quantification %
in the context of counterfactual explainability, %
demonstrating that its broader adoption could address key challenges in the field. %
To this end, we show how uncertainty can provide %
a principled unifying framework for counterfactual explainability %
by %
expressing the core counterfactual properties in terms of uncertainty, %
allowing us to build %
two variants of an explainer upon them -- %
one based solely on uncertainty estimates and %
another pairing them with distance measured in the feature space. %
Our comprehensive experiments illustrate %
highly competitive performance of our framework %
when compared to many state-of-the-art methods %
despite its radically simple design. %
More broadly, %
the paper demonstrates that %
integrating artificial intelligence fundamentals into transparency research promises to yield more reliable, robust and understandable predictive models. %
We posit that %
making artificial intelligence explainability truly uncertainty-aware is the first step towards this goal. %
\end{abstract}

\begin{keywords}
 \sep \sep
explainable artificial intelligence \sep interpretable machine learning \sep counterfactual explainability \sep uncertainty quantification \sep aleatoric uncertainty \sep epistemic uncertainty
\end{keywords}

\maketitle

\section{Uncertainty and Transparency}\label{sec:intro}%

Artificial intelligence (AI) models can achieve impressive results across many domains, %
but their deployment is often stymied by their opaqueness, unreliability and lack of robustness~\citep{rudin2019stop}. %
Consequently, two paradigms have emerged to alleviate such issues: %
\emph{ante hoc} interpretability envisages building inherently transparent models whose functioning adheres to domain-specific constraints, whereas %
\emph{post hoc} explainability delivers tools that elucidate the operation of predictive models through independent explanatory mechanisms~\citep{sokol2020explainability}. %
Additionally, \emph{uncertainty quantification} has been proposed to improve the accountability of AI by looking beyond crisp (deterministic) classification as well as model evaluation solely based on predictive performance; %
instead, it sets out to %
provide truthful representation of models' \emph{aleatoric} -- %
which is %
inherent to data-generating processes -- and \emph{epistemic} -- %
i.e., arising %
due to observation sparsity -- %
uncertainty~\citep{hullermeier2021aleatoric}. %

However, methods for reliably estimating uncertainty %
remain somewhat underdeveloped, with such considerations often being neglected. %
Moreover, while \emph{post hoc} approaches are at the forefront of \emph{human-centred} explainable AI (XAI) %
-- enabling diverse audiences, both with and without technical expertise, to peer inside predictive models~\citep{miller2019explanation} -- %
these techniques are usually unable to faithfully capture the operation of AI systems, possibly offering misleading insights~\citep{rudin2019stop}. %
As a result, \emph{ante hoc} interpretable models are preferred in (high stakes) real-world applications %
even though %
their functioning %
may remain largely %
opaque to non-technical stakeholders, who nonetheless
tend to be their primary users~\citep{sokol2023reasonable}. %

In this paper %
we demonstrate that %
AI transparency techniques are largely oblivious to various notions of uncertainty %
despite %
the two being fundamentally interconnected. %
Consequently, %
XAI methods by and large dismiss or overlook any notion of uncertainty that a predictive model may provide -- %
usually fixating on %
its crisp predictions -- %
which can result in subpar explanations. %
More broadly, %
we note that XAI research tends to neglect the rich tapestry of foundational AI concepts, %
which can lead to reinventing what already exists; %
admittedly, this sentiment is more pertinent to \emph{post hoc} than \emph{ante hoc} approaches since the latter draw upon decades of work on classic AI models~\citep{rudin2022interpretable}. %
We %
support these observations by %
showing that connecting AI transparency and uncertainty quantification can address many open challenges in XAI. %
Specifically, we use %
the example of \emph{counterfactual explanations} %
given that they are %
considered the gold standard of human-centred XAI~\citep{miller2019explanation}. %

To this end, we first review relevant topics in \autoref{sec:related}, namely: %
counterfactual explainability~(\sref{sec:related:exp}), uncertainty quantification~(\sref{sec:related:uq}) and the intersection of these two fields~(\sref{sec:related:exp+uq}). %
Next, in \autoref{sec:methodology}, we motivate and present our uncertainty-aware XAI framework for counterfactual explanations. %
We begin by summarising open challenges~(\sref{sec:methodology:challenges}), which include: %
misguided or overlooked modelling assumptions resulting in AI that lacks sound technical design and functioning; %
overemphasis on crisp classification and inadequate XAI evaluation practice that, among others, disregards the quality of the explained models; %
\emph{ad hoc} fixes and unreliable proxies used in lieu of principled uncertainty handling; %
counterfactual generation's dependence on distance measured in the feature space despite inherent difficulties posed by its meaningful formalisation; %
prevalence %
of neural models applied to image and text data leading to neglect of %
classic (inherently transparent) AI methods; %
overlooked synergy between uncertainty quantification and \emph{ante hoc} interpretability; %
and %
missing human-centred perspective in the latter field. %

As a first step towards addressing these %
shortcomings, %
we show %
how %
uncertainty quantification %
could provide %
a rigorous %
unifying framework for constructing state-of-the-art counterfactuals. %
In particular, we demonstrate %
that %
the fundamental properties of this explanation type can be expressed in terms of two principal, and orthogonal, concepts: %
\emph{uncertainty} %
decomposed into its aleatoric and epistemic components %
as well as %
\emph{distance} %
measured in the feature space~(\sref{sec:methodology:cf}). %
Then, %
we %
combine our streamlined %
definitions of counterfactual desiderata %
into a coherent optimisation objective that forms the backbone of our novel uncertainty-aware explainer~(\sref{sec:methodology:gen}); %
specifically, %
we introduce its two variants: %
one that relies exclusively on uncertainty estimates and another that complements them with distance measurements. %

\autoref{sec:experiments} %
reports evaluation results for a comprehensive set of experiments. %
First, we %
assess the individual efficacy of our uncertainty-based definitions of counterfactual desiderata to %
better understand their strengths and weaknesses; %
next, we %
compare the performance of %
the two variants of our explainer %
to state-of-the-art benchmarks. %
We restrict our investigation to \emph{tabular} data sets (popular in XAI research) given the prevalence of this data type in %
real-life applications~\citep{shwartz2022tabular} and success of counterfactual explanations in this domain~\citep{wachter2017counterfactual}. %
We look primarily at evaluation outcomes aggregated across data sets, %
but also inspect specific results for completeness. %
These experiments %
clearly show %
that our %
uncertainty-based formalisation of counterfactual desiderata is overall well aligned with evaluation metrics currently used for this explanation type %
and that %
both variants of our explainer are highly competitive %
despite their fundamental simplicity. %

\autoref{sec:discussion} %
complements our quantitative analysis with a comprehensive discussion of the uncertainty-based approach to explainability. %
In particular, we %
revisit the strong connection between %
\emph{ante hoc} interpretability and uncertainty quantification~(\sref{sec:methodology:ante}). %
Then, we %
explore additional counterfactual desiderata that could be integrated into our framework %
and %
review the implications
of using distance (measured in the feature space) when generating explanations~(\sref{sec:methodology:extra}). %
We wrap up this section by %
investigating %
the broader implications of the proposed uncertainty-based approach to explainability~(\sref{sec:methodology:anti}). %
We conclude the paper in \autoref{sec:conclusion} by summarising our findings and outlining future research directions. %
In short, %
our novel framework %
delivers a more principled counterfactual generation process at negligible cost, %
but %
further research is necessary to overcome the identified challenges and facilitate progress in this area. %

\section{Background and Related Work}\label{sec:related}%

We begin with an overview of %
latest research in \emph{counterfactual explainability}~(\sref{sec:related:exp}) and \emph{uncertainty quantification}~(\sref{sec:related:uq}) %
as well as %
work at the \emph{intersection of these two fields}~(\sref{sec:related:exp+uq}). %

\subsection{Counterfactual Explainability}\label{sec:related:exp}%

Counterfactuals capture hypothetical situations where the outcome of interest is different (usually more desirable) than what has been observed. %
They are considered the gold standard of \emph{human-centred} XAI given their strong foundations in social sciences, natural familiarity to lay and expert audiences, and regulatory compliance~\citep{wachter2017counterfactual,miller2019explanation,guidotti2022counterfactual}. %
In their simplest form, %
given a \emph{crisp} AI model $f: \mathcal{X} \mapsto \mathcal{C}$, %
they are generated by minimising the distance $d: \mathcal{X} \times \mathcal{X} \mapsto \mathds{R}^+$ in the $n$-dimensional feature space $\mathcal{X}$ between the current instance $\xF \in \mathcal{X}$ %
and a hypothetical data point $\xCF \in \mathcal{X}$ %
that is assigned the desired prediction, i.e., %
$f(\xCF) = \yCF \in \mathcal{C}$ is favoured over $f(\xF) = \yF \in \mathcal{C}$~\citep{wachter2017counterfactual}. %
Simply put, %
we search for %
the most \emph{similar} (distance- and feature-wise) instance of a different class %
-- an approach formalised in \autoref{eq:cf-optimisation}, %
where %
$d_H: \mathcal{X} \times \mathcal{X} \mapsto \mathds{R}^+$ is the normalised Hamming distance $d_H(x, x^\prime) = \frac{1}{n} \sum_{i=1}^n \mathds{1}_{x_i \neq x^\prime_i}$ counting the number of changed features and %
$\lambda \in [0, 1]$ is a user-specified parameter controlling the trade-off between $d(\cdot, \cdot)$ and $d_H(\cdot, \cdot)$. %

\begin{multline}\label{eq:cf-optimisation}
    \xCF = \argmin_{x \in \mathcal{X}} %
        \lambda d(\xF, x) %
        + (1 - \lambda) d_H(\xF, x)
        \\
        \text{s.t.} \quad %
        f(\xF) \neq f(x)
\end{multline}

\begin{figure*}[t]%
  \centering
  \begin{subfigure}[t]{0.31\textwidth}
      \centering
      \includegraphics[scale=0.36]{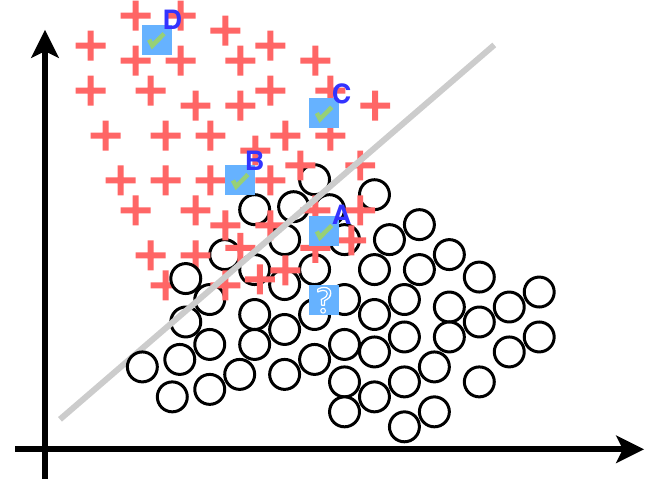}%
      \caption{%
          Visualisation of the three \emph{fundamental} counterfactual desiderata. %
          Explanation~A is \textbf{invalid}; explanation~B lacks \textbf{sparsity} in comparison to explanation~C as the former requires changing two features whereas the latter only one; explanation~D lacks \textbf{similarity} when compared to explanation~C given that it is farther away. %
          }\label{fig:counterfactual:basic}%
  \end{subfigure}
  \hspace{0.01\textwidth}
  \begin{subfigure}[t]{0.63\textwidth} %
      \centering
      \includegraphics[scale=0.36]{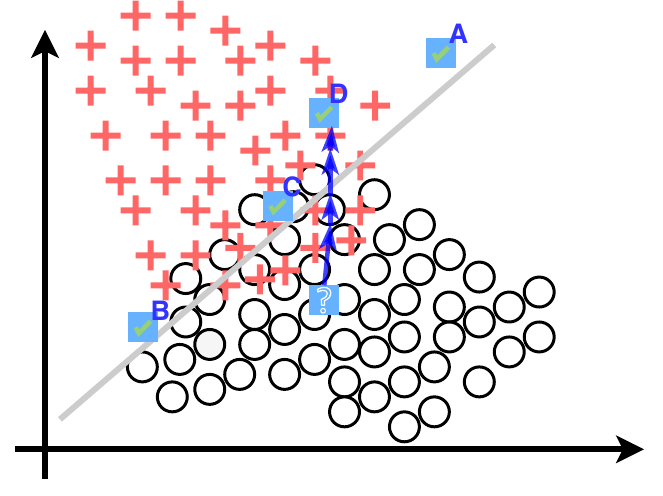}%
      \caption{%
          Visualisation of the seven \emph{extended} counterfactual desiderata. %
          Explanation~A is \textbf{implausible} as it lies outside of the data manifold. %
          Explanation~B lacks \textbf{connectedness} as its closes neighbour is classified negative. %
          Explanation~C is not \textbf{discriminative} as it lies in a region where classes overlap. %
          Explanations~A, B and C may lack \textbf{robustness} as more data points are collected to train the model and it further converges. %
          All the explanations are considered \textbf{stable} given the simplicity of this example. %
          Because of the high data density all the explanations are \textbf{feasible} as there exists a path -- such as the one shown with blue arrows for explanation~D -- connecting them with the factual instance. %
          Finally, if feature $x_1$ (horizontal axis) is assumed actionable but feature $x_2$ (vertical axis) not, only explanation~D is \textbf{actionable} as the other counterfactuals require changing both features $x_1$ and $x_2$. %
      }\label{fig:counterfactual:extended}
  \end{subfigure}
  \caption{%
  Visualisation of the (\subref{fig:counterfactual:basic}) fundamental -- \emph{validity}, \emph{similarity} and \emph{sparsity} -- and (\subref{fig:counterfactual:extended}) extended -- \emph{plausibility}, \emph{connectedness}, \emph{discriminativeness}, \emph{robustness}, \emph{stability}, \emph{feasibility} and \emph{actionability} -- properties of human-centred counterfactual explanations. %
  The question mark indicates the explained instance and the check mark represents a counterfactual data point. %
  }\label{fig:counterfactual}%
\end{figure*}

This process optimises for the following three properties shown in \autoref{fig:counterfactual:basic}: %
\begin{description}[topsep=0pt,partopsep=0pt,noitemsep]%
  \item[Validity] guarantees that the counterfactual instance is classified by the explained model with the desired class -- the $f(\xF) \neq f(x)$ part of \autoref{eq:cf-optimisation}. %
  \item[Similarity] places the counterfactual instance as close as possible to the factual data point in the feature space (according to the chosen distance metric $d$), making it highly relevant and easy to reach  -- the $d(\xF, x)$ part of \autoref{eq:cf-optimisation}. %
  \item[Sparsity] minimises the number of features changed between the factual and counterfactual instances, enhancing human comprehensibility of the explanation -- the $d_H(\xF, x)$ part of \autoref{eq:cf-optimisation}. %
\end{description}

In practice, however, this strategy often yields counterfactuals that are perceptually closer to \emph{adversarial examples}~\citep{goodfellow2014explaining} than \emph{meaningful explanations}~\citep{freiesleben2022intriguing}, i.e., the recommended change is either unrealistic or carries no intrinsic meaning. %
As a solution, %
a collection of new \emph{social}, \emph{technical} and \emph{sociotechnical} desiderata was proposed~\citep{pawelczyk2020learning,delaney2021uncertainty,schut2021generating,guidotti2022counterfactual}. %
Their formalisation, nonetheless, varies widely %
across the literature; %
examples of evaluation metrics corresponding to some of them are listed in \autoref{apx:eval-metrics} for reference. %
Seven of the most prominent such properties, visualised in \autoref{fig:counterfactual:extended}, are: %
\begin{description}[topsep=0pt,partopsep=0pt,noitemsep]%
  \item[Plausibility] requires the counterfactual state to come from the data manifold, thus be achievable in real life. %
  \item[Connectedness] extends \emph{plausibility} by ensuring that the counterfactual instance is supported by (i.e., is close in the feature space to) another data point %
  assigned to the counterfactual class (with high probability) by the explained model %
  such that there exists a direct line (or a sequence of steps through known data points) between the two that does not cross a decision boundary. %
  \item[Discriminativeness] makes the counterfactual instance \emph{unambiguous} -- i.e., clearly \emph{distinguishable} from similar instances (refer back to \emph{similarity}) that are not of the counterfactual class -- to %
  avoid \emph{confusing} humans with the like of \emph{adversarial examples}. %
  \item[Robustness] prevents \emph{realistic} data shifts and model chan\-ges from invalidating counterfactual explanations. %
  \item[Stability] ensures that the neighbours of the explained data point receive \emph{similar} (or identical) counterfactuals. %
  \item[Feasibility] %
  connects the factual and counterfactual points with a traversable path that follows pre-existing data instance, %
  thus
  building upon and further expanding \emph{plausibility}, \emph{connectedness} and \emph{stability}. %
  \item[Actionability] guarantees that counterfactuals can be implemented in real life, as some features may be immutable, e.g., ethnicity, while others must obey monotonicity, e.g., age, %
  simultaneously maintaining the congruity of these changes given that they may be incompatible. %
\end{description}

Notably, all of these properties -- illustrated in \autoref{fig:counterfactual} -- pertain to the counterfactual instance itself. %
Complementary perspectives that account for %
constructing individual explanations from \emph{sequences of discrete steps} %
as well as generating representative \emph{collections of counterfactuals} %
provide important extensions of this paradigm~\citep{poyiadzi2020face,sokol2020one}. %
We explore these ideas further in \autoref{sec:methodology:extra}, %
but leave their full analysis through the lens of uncertainty for future work %
given the focus of this paper on vanilla counterfactual explanations. %

Adhering to the properties listed above %
allows to retrieve state-of-the-art counterfactuals without the need for modelling the causal structure of the data-generating processes~\citep{karimi2022survey}, which in real life is often infeasible.%
\footnote{%
The causal perspective is outside of this paper's scope, but %
desiderata such as \emph{actionability} and \emph{plausibility} can act as its proxy.%
} %
Such explanations, nonetheless, may still lack reliability, hence be unsuitable for high stakes domains, if they do not fully account for the inner workings of the underlying predictive model~\citep{rudin2019stop}. %
In XAI, a distinction is made between \emph{ante} and \emph{post hoc} techniques. %
\emph{Post hoc} methods generate explanations using mechanisms that operate independently of predictive models, %
which makes them portable but also prevents them from accurately capturing how these models make decisions. %
\emph{Ante hoc} interpretability, in contrast, is achieved by building inherently transparent models that adhere to domain-specific constraints, %
which guarantees their reliability, but %
often at the expense of limiting their comprehensibility to AI experts, with the explanatory needs of other stakeholders left unaddressed~\citep{sokol2023reasonable}. %
While counterfactuals are ideal for this purpose, most relevant explainers are \emph{post hoc}~\citep{guidotti2022counterfactual}, leaving this problem unresolved. %

\begin{figure*}[t]%
  \centering
  \begin{subfigure}[t]{0.63\textwidth}%
      \centering
      \includegraphics[scale=0.36]{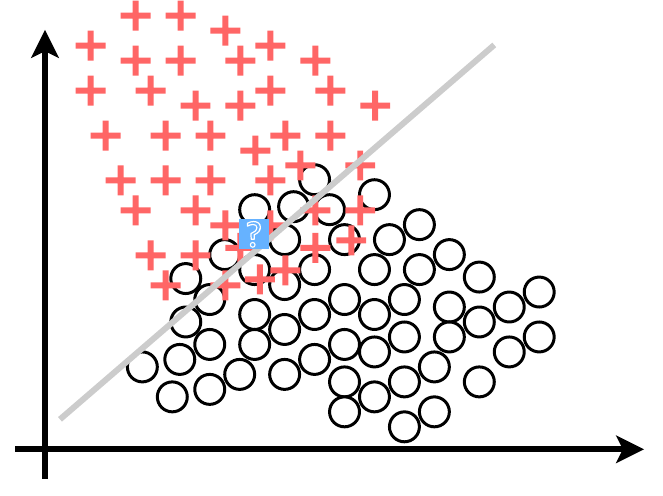}%
      \caption{The optimal model may remain \emph{aleatorically} uncertain about a prediction, e.g., the question mark, because of the intrinsic class overlap. This type of uncertainty cannot be (easily) reduced. Note, however, that collecting additional information (rather than observations), e.g., an extra feature, could allow separating the overlapping instances in the added dimension at the expense of increased \emph{epistemic} uncertainty.}\label{fig:uncertainty:aleatoric} %
  \end{subfigure}
  \hspace{0.01\textwidth}
  \begin{subfigure}[t]{0.31\textwidth}%
      \centering
      \includegraphics[scale=0.36]{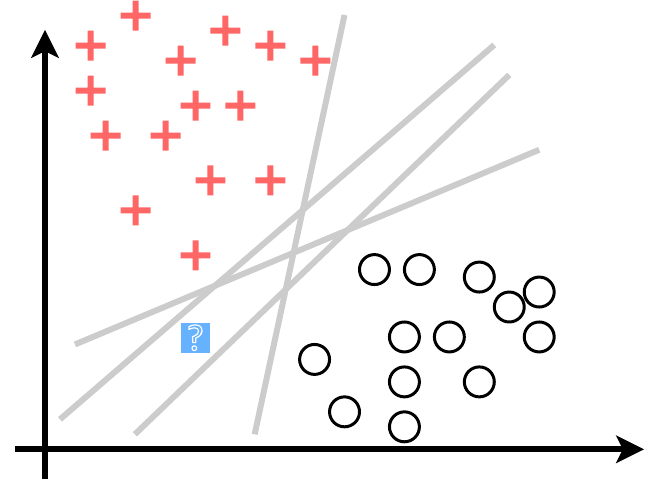}%
      \caption{The lack of knowledge about the optimal model may result in \emph{epistemic} uncertainty about a prediction, e.g., the question mark. This type of uncertainty can usually be reduced by collecting more data -- see Panel~(\subref{fig:uncertainty:aleatoric}).}\label{fig:uncertainty:epistemic}%
  \end{subfigure}
  \caption{%
Demonstration of (\subref{fig:uncertainty:aleatoric}) \emph{aleatoric} and (\subref{fig:uncertainty:epistemic}) \emph{epistemic} uncertainty %
  for a two-dimensional toy data set with the model class restricted to linear classifiers~\citep{hullermeier2021aleatoric}.%
    }\label{fig:uncertainty}%
\end{figure*}

\subsection{Uncertainty Quantification}\label{sec:related:uq}%

In addition to transparency, %
other critical properties of data-driven models -- such as their %
fairness, robustness, reliability, trustworthiness and accountability -- have gained prominence %
with the proliferation of AI. %
The foundational aspect of these desiderata is \emph{predictive uncertainty} %
given its crucial role in enhancing AI systems' awareness of their limitations~\citep{tomsett2020rapid}. %
Primarily, this involves distinguishing between \emph{aleatoric} uncertainty, which arises due to the inherent randomness of the data-generating process, and \emph{epistemic} uncertainty, %
which is caused by %
the learning algorithm's ignorance of the true underlying model~\citep{hullermeier2021aleatoric}. %

Aleatoric uncertainty can be represented by a \emph{probabilistic} model $h: \mathcal{X} \mapsto \mathds{P}[\mathcal{C}]$ mapping the input space %
to probability distributions over the outcome space $\mathcal{C}$ such as class labels. %
Capturing %
epistemic uncertainty, on the other hand, requires second-order formalisms; %
this is commonly achieved by defining %
a model $H: \mathcal{X} \mapsto \mathds{P}[\mathds{P}[\mathcal{C}]]$ mapping the input space %
to second-order probability distributions, which allows it to represent uncertainty about the true (conditional) probability distribution $\mathds{P}$ on $\mathcal{C}$ (given an instance $x \in \mathcal{X}$). %

Various methods have recently been proposed to facilitate reliable uncertainty-aware learning, but they remain in early stages of development~\citep{silva2023classifier}; %
by and large, we lack AI models that are truly uncertainty-aware, which curtails the use of data-driven tools in high stakes domains~\citep{rudin2022interpretable}. %
Notably, many such techniques are grounded in the classical Bayesian approach, where $H(x)$ is the posterior (predictive) distribution. %
The Bayesian extensions of neural networks is one prominent example in this space~\citep{depeweg2018decomposition}; %
more broadly, %
uncertainty quantification in deep learning, %
e.g., based on approximate Bayesian methods~\citep{abdar2021review}, is %
an active area of research~\citep{gawlikowski2023survey}. %
In practice, %
second-order predictors are often approximated by model ensembles, where an output $H(x)$ is given by a set of predictions $h_1(x), \ldots, h_z(x)$ %
delivered by a collection of $z$ ensemble members $h_i$ that themselves are first-order models $h_i: \mathcal{X} \mapsto \mathds{P}[\mathcal{C}]$; %
epistemic uncertainty is then captured by this diverse set of predictions~\citep{lakshminarayanan2017simple}. %

In general, aleatoric uncertainty relates to predictions that come from the regions of $\mathcal{X}$ where classes overlap. %
It %
is inherently \emph{irreducible} in the sense that further observations of a given phenomenon will not improve our ability to anticipate its outcome; %
a prototypical example is a coin toss: no matter how many data points we collect, we cannot reliably predict how the coin will land next given the intrinsic variability of this process. %
In certain cases, %
changing the problem representation -- e.g., by collecting more information, like an extra feature, as opposed to more observations -- %
can help to alleviate aleatoric uncertainty -- e.g., by separating thus far overlapping data points in the added dimension. %
Such an approach reduces aleatoric uncertainty often at the expense of its epistemic counterpart as fitting a model in higher dimensions comes with greater difficult and requires a larger data set. %

Epistemic uncertainty, on the other hand, occurs in the regions of $\mathcal{X}$ where multiple predictions are plausible because there remains a collection of admissible candidate models whose outputs disagree. %
It %
is fundamentally \emph{reducible} as further observations of a given phenomenon are likely to improve our ability to reliably anticipate its outcome; %
revisiting the coin toss example: the larger our set of observations is, the more precisely we can estimate the coin's bias. %
In the context of AI, collecting additional training data facilitates better approximation of the data-generating process, allowing us to derive the best (Bayes-optimal) model. %
\autoref{fig:uncertainty} illustrates these two types of uncertainty for a toy binary classification task in a two-dimensional feature space. %

In practice, \emph{uncertainty quantification} sets out to measure %
(as a real number) %
the %
total uncertainty reflected by a prediction, and to disentangle it into its aleatoric and epistemic parts. %
When dealing with second-order predictions, %
the total uncertainty $\TU(x)$ of an instance $x \in \mathcal{X}$ is usually quantified %
in terms of the entropy of the distribution on $\mathcal{C}$, %
which in the case of a finite ensemble approximation yields: %
\begin{equation} \label{eq:tu}
  \TU(x) = \ENT\left(\frac{1}{z} \sum_{i=1}^z h_i(x)\right)\text{,}
\end{equation}
where $\ENT(\cdot)$ denotes the Shannon entropy. %
This formulation of $\TU(x)$ %
can then be decomposed into the sum of the conditional entropy of an outcome given the (first-order) probability and the mutual information between the two~\citep{depeweg2018decomposition}. %
The former serves as a natural measure of aleatoric uncertainty: %
\begin{equation} \label{eq:au}
  \AU(x) = \frac{1}{z} \sum_{i=1}^z \ENT(h_i(x))\text{,}
\end{equation}
whereas %
the latter can be taken as a measure of epistemic uncertainty: %
\begin{equation} \label{eq:eu}
  \EU(x) = \TU(x) - \AU(x)\text{.}
\end{equation}

\begin{figure*}[t]%
    \centering
    \begin{subfigure}[t]{0.470\textwidth}%
        \centering
        \includegraphics[scale=0.36]{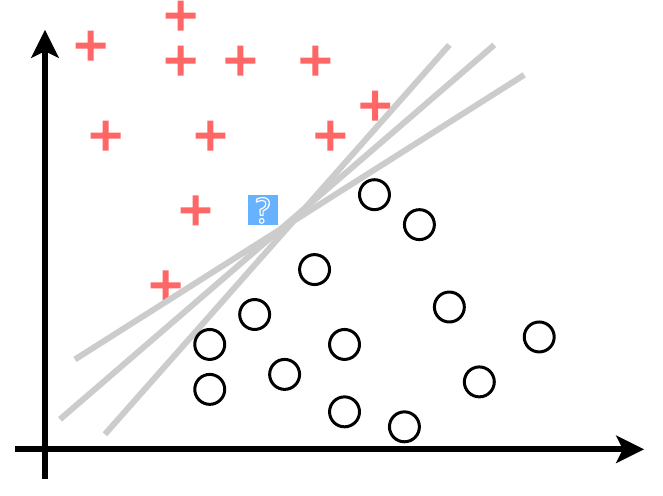}%
        \caption{Given the class of \emph{linear} models, the point represented by the question mark will be classified as positive with relatively high certainty (despite the lack of data observed in its neighbourhood that could directly support this prediction).}\label{fig:model:linear}%
    \end{subfigure}
    \hspace{0.01\textwidth}
    \begin{subfigure}[t]{0.470\textwidth}%
        \centering
        \includegraphics[scale=0.36]{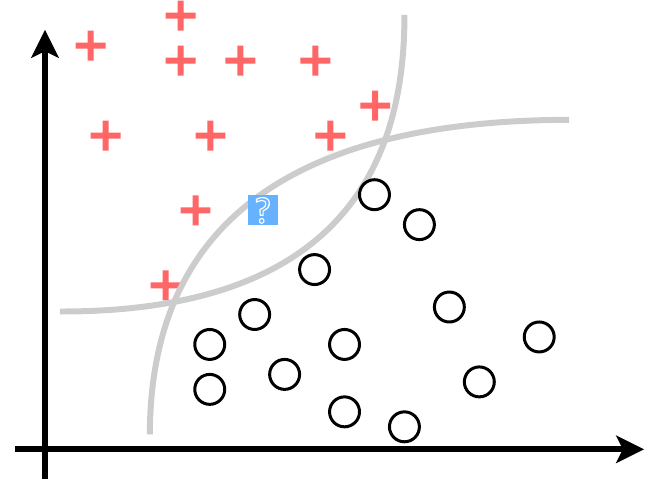}%
        \caption{When using the \emph{quadratic} model class -- which is more flexible and expressive than linear models -- the point represented by the question mark cannot be classified as positive or negative with high certainty (unless more data are collected).}\label{fig:model:quadratic}%
    \end{subfigure}
    \caption{Demonstration of how uncertainty quantification depends on the assumed model class for (\subref{fig:model:linear}) \emph{linear} and (\subref{fig:model:quadratic}) \emph{quadratic} classifiers using a two-dimensional toy data set~\citep{hullermeier2021aleatoric}.}\label{fig:model}%
\end{figure*}

Despite often being overlooked, uncertainty estimation is \emph{never} assumption-free. %
It is influenced by the underlying, and sometimes implicit, data modelling assumptions, among which the model class choice is particularly consequential. %
Notably, the degree of predictive uncertainty is determined by the flexibility of the model class -- as shown in \autoref{fig:model}. %
More restrictive families of functions result in lower uncertainty and vice versa, with those deemed universal approximators, e.g., neural networks, impacted the most~\citep{hullermeier2021aleatoric}. %
Other, model-specific, assumptions also influence uncertainty quantification. %
For example, linear models -- see \autoref{fig:model:linear} -- presuppose that the farther a data point is placed from the decision boundary, the less uncertain its prediction is. %
This holds even if such an instance comes from a sparse data region. %
While this property is inherent to linear models, its consequences may be undesirable and necessitate special handling, e.g., tweaking the uncertainty estimates \emph{post hoc} to convey low confidence outside of the data manifold~\citep{perello2016background}. %

In addition to aleatoric and epistemic, various other sources of uncertainty exist~\citep{hullermeier2021aleatoric}. %
\emph{Model uncertainty} comes from incorrect model class choice, which is common in practice; %
it cannot be easily detected and addressed, %
but when the model class is highly expressive (or considered a universal approximator), this type of uncertainty becomes inconsequential. %
\emph{Approximation uncertainty} arises when the model learnt from data does not align with the true or optimal model; %
it contributes to epistemic uncertainty and is particularly pronounced for highly expressive model classes applied to sparse training data. %
In general, uncertainty quantification is premised on the \emph{closed/stable world assumption}, violating which makes this process considerably more difficult~\citep{gigerenzer2023psychological}. %
Relevant examples are %
non-stationary data-generating processes or data distribution shifts, %
emergence of new and disappearance of known target classes, and %
noisy or imprecise data features (e.g., due to measurement errors). %

Another challenge comes from %
representing \emph{lack of knowledge}. %
The common approach of modelling total ignorance with the uniform distribution is unable to distinguish between %
\emph{precise} knowledge of an inherently random process, e.g., a fair coin toss, %
and \emph{complete lack} of knowledge about it, e.g., a possible bias of the coin. %
Overcoming this problem requires working with \emph{second-order uncertainty} given that a single distribution cannot capture uncertain knowledge~\citep{shafer1976mathematical,dubois1988possibility,walley1991statistical,smets1994transferable}. %

\subsection{Uncertainty and Counterfactuals}\label{sec:related:exp+uq} %

Integrating uncertainty quantification with XAI has recently gained interest, albeit with much left to be explored~\citep{bhatt2021uncertainty}. %
As it stands, explainability research largely overlooks probabilistic modelling -- and is particularly oblivious to the notion of uncertainty -- %
despite their synergistic relation, e.g., leading to explanations with improved reliability, robustness and trustworthiness~\citep{mehdiyev2025integrating}. %
More broadly, explaining why an AI system is uncertain can help humans to understand its limitations and act to reduce its uncertainty; %
conversely, revealing the uncertainty of an explanation can improve humans' ability to make responsible AI-assisted decisions. %

While some leap ahead and propose conceptual frameworks for integrating the two fields -- offering methods to quantify the uncertainty of explanations~\citep{salvi2025explainability} -- %
the uncertainty formalisations in XAI %
are not necessarily coherent with those in AI~\citep{chiaburu2024uncertainty}. %
As a consequence, the uncertainty of an explanation may not be directly linked to that of the underlying model, which itself can even be non-probabilistic (i.e., crisp). %
In such a setting, it remains unclear how one would even reconcile the notions of aleatoric and epistemic uncertainty reported for a model and an explainer. %

More specifically, in XAI the \emph{aleatoric} component quantifies %
explanation consistency under input variation, and %
the \emph{epistemic} part captures %
explanation stability across multiple inferences of a given instance. %
The latter relies on %
ensembles, distinct model initialisations or configurations as well as model-specific techniques such as Monte Carlo dropout, %
effectively taking advantage of the \emph{model multiplicity} phenomenon -- sometimes called the Rash\=omon effect of statistics~\citep{breiman2001statistical,rudin2024amazing} -- where %
a group of models has comparable predictive performance despite intrinsic differences (see \cref{fig:uncertainty:epistemic,fig:model}).  %
However, these definitions %
conflate %
the variability of an explanation attributed to a model %
with that introduced by an explainer, which may be non-deterministic %
as is the case with many \emph{post hoc} approaches~\citep{sokol2022what}. %
While grounding the definitions of explanation uncertainty in \emph{ante hoc} interpretable AI %
appears necessary to overcome these challenges, %
reliable uncertainty quantification remains a key hurdle in building this type of models~\citep{rudin2022interpretable}. %

Specifically at the intersection of uncertainty quantification and counterfactual explainability, %
\citet{antoran2020getting} proposed a method that instructs its users how to reduce uncertainty of a prediction. %
\citet{delaney2021uncertainty} used uncertainty to evaluate the quality of %
counterfactuals, expanding the \emph{validity} desideratum from crisp to probabilistic classification; %
to this end, they proposed \emph{trust scores}, which link (epistemic) uncertainty to counterfactual \emph{plausibility}. %
Others %
harnessed uncertainty estimates to discard out-of-distribution explanations~\citep{teufel2025improving} or %
adapted pre-existing outlier detection methods for this purpose~\citep{romashov2022baycon}. %
Probabilistically plausible counterfactuals have also been generated directly based on data density estimates computed with normalising flows~\citep{wielopolski2024probabilistically}. %
Similarly, \citet{thiagarajan2022training} %
designed an explainer that %
generates high-confidence counterfactual data points by %
minimising their uncertainty. %

\citet{duell2024quce} %
expanded the notion of counterfactual \emph{plausibility} by %
computing uncertainty of the vector connecting the factual and counterfactual instances. %
\citet{kanamori2024learning} explored %
the implicit link between epistemic uncertainty and %
the aforementioned Rash\=omon effect %
in the context of %
counterfactual \emph{availability} and \emph{robustness}. %
Likewise, \citet{christodoulou2026impact} studied the latter property in presence of increasing aleatoric and epistemic modelling uncertainty for various explainers. %
Finally, %
\citet{schut2021generating} directly coupled counterfactual explainability and uncertainty %
by connecting aleatoric and epistemic uncertainty respectively to %
counterfactual discriminativeness (which they called \emph{unambiguity}) and plausibility (referred to as \emph{realism}) %
to generate high-quality explanations. %

\section{Building Bridges: Uncertainty-aware XAI}\label{sec:methodology}%

Despite latent connections between uncertainty and various forms of AI interpretability and explainability, %
many such links remain superficial or unexplored, with the underlying challenges unaddressed. %
To fill this gap, we first summarise and discuss %
pressing open questions at the intersection of these research fields~(\sref{sec:methodology:challenges}); %
in particular, we demonstrate that, to its detriment, %
XAI %
remains uncertainty-unaware %
and, more broadly, it %
lacks a widely accepted foundation. %
Next, we show %
how uncertainty quantification can provide a \emph{principled unifying framework} for generating state-of-the-art (non-causal) counterfactuals, offering a formalisation that supersedes many \emph{ad hoc} proxies and criteria currently used to this end~(\sref{sec:methodology:cf}); %
of particular relevance here is the distinction between aleatoric and epistemic uncertainty, which allows to overcome the limitations imposed by relying on its first-order or aggregate measures. %
Then, we demonstrate how to combine our uncertainty-based definitions of counterfactual properties into a coherent optimisation objective that forms the backbone of our novel uncertainty-aware explainer~(\sref{sec:methodology:gen}); %
our approach is built upon two fundamental concepts: %
(aleatoric and epistemic) uncertainty and distance. %

\subsection{Open Challenges}\label{sec:methodology:challenges}

While a broad consensus has emerged regarding counterfactual desiderata (refer back to \autoref{sec:related:exp}), disagreement persists on their specific formalisation; %
nonetheless, %
many of these definitions are based upon some notion of distance, usually measured directly in the feature space~\citep{guidotti2022counterfactual}. %
But formalising realistic distance functions remains an open challenge, %
especially that such metrics ought to be tailored to individual application domains %
as %
to capture meaningful patterns and dependencies in the underlying data. %
Notably, this problem is not unique to counterfactuals, or XAI more broadly, and also affects fields such as AI fairness (where one may need to quantify similarity of two persons)~\citep{dwork2012fairness}. %
Researchers thus often fall back on generic distance functions -- e.g., Manhattan, Euclidean, Gower or their bespoke mixture -- %
while simultaneously %
acknowledging the inherent limitations of this approach. %

The distance metric choice %
is particularly consequential for counterfactual generation %
as its incoherence with the explained model's internal notion of similarity can result in technically valid yet impractical or unrealistic explanations; %
the same caveat applies to evaluation (see \autoref{apx:eval-metrics} for more details). %
Taking these two problematic aspects of XAI together, %
explainers that use a flawed notion of distance to generate counterfactuals and then rely on the same measure to evaluate them %
can deliver highly misleading results (and explanations). %
Additional research is therefore needed to address these challenges; %
while such considerations are outside of our paper's scope, %
\autoref{apx:dist-proxy} explores this topic further and %
outlines one promising approach that uses instance sampling likelihood as a proxy for similarity. %

Regarding uncertainty, %
\autoref{sec:related:exp+uq} showed that %
it %
has primarily been applied, albeit often indirectly, to evaluate counterfactuals and, too a much lesser extent, guide their generation. %
These studies also tend to be %
confined to probabilistic predictions output by AI models or aleatoric (first-order) uncertainty alone. %
Additionally, %
explainers usually \emph{overlook} any data modelling assumptions that are made despite %
their foundational role in robust quantification of uncertainty (as well as meaningful definition of distance) and resort to \emph{ad hoc} fixes and stopgaps when uncertainty estimates are unavailable. %
They employ proxies like %
operating within \emph{previously observed data points}, %
\emph{measuring distance} to the nearest instance, %
adapting pre-existing \emph{outlier detection} methods, %
approximating the data manifold in \emph{latent space} %
or defining \emph{custom metrics} such as the aforementioned trust scores. %
While these approaches help to ensure \emph{plausibility}, \emph{connectedness} and \emph{actionability} -- as well as other desiderata -- they rely on surrogate measures of uncertainty %
that may be unreliable, e.g., when they are incompatible with the explained class of models. %

XAI's blindness to uncertainty %
thus not only undermines the \emph{truthfulness} of explanations, but more critically it obscures the challenge of %
building reliable and robust AI systems %
that are %
founded in sound design principles %
and compatible with the envisaged application~\citep{schut2021generating,rudin2022interpretable}. %
The latter is symptomatic of a broader problem in XAI (that is particularly prominent in its \emph{post hoc} sub-field) where the quality of explainers or explanations is assessed independently of the underlying models' quality~\citep{sokol2024what}. %
Another consequence of uncertainty unawareness is the dominance of (counterfactual) explainers that are confined to \emph{crisp} predictions. %
This limitation is particularly easy to overlook as XAI tools mostly focus on %
\emph{binary} classification problems in which the notions of uncertainty or class probability may not provide clear benefit. %
Nonetheless, this ignorance, for example, %
prevents explainers from recognising the (relative) uncertainty of individual classes -- hence exacerbating explanation \emph{ambiguity} -- as well as their possible structural or hierarchical relation~\citep{sokol2020limetree}. %

More broadly, XAI rarely ventures beyond %
models that output specific predictions, like class labels; %
while it occasionally considers \emph{first-order} probabilistic predictions, %
it largely neglects \emph{second-order} uncertainty~\citep{wood2024model}. %
But %
just as focusing on a single ``optimal'' model or prediction has been criticised as insufficient or even counterproductive~\citep{hullermeier2021aleatoric,rudin2024amazing}, seeking a single ``best'' explanation appears undesirable too~\citep{sokol2020one}. %
By embracing uncertainty, both in AI models and their explainers, we can address such challenges on their most fundamental level. %
This perspective enables a shift towards considering multiple explanations with diverse and well-defined properties as well as known limitations -- a practice that is more beneficial and better aligned with the various needs and expectations of distinct end users~\citep{miller2023explainable}. %
It also enforces consistency and compatibility of various notions of uncertainty used across AI and explainability work, which, as noted in \autoref{sec:related:exp+uq}, is needed to facilitate further progress in XAI. %
Introducing uncertainty awareness, however, %
may inadvertently increase the complexity of explanations, %
making them less appealing to non-experts; %
we explore this topic further in \autoref{sec:methodology:anti}. %

Another relevant open challenge is the general neglect of %
(\emph{ante hoc}) inherently interpretable predictors~\citep{rudin2019stop}. %
Research at the intersection of XAI and uncertainty quantification is predominantly concerned with differentiable as well as neural model classes applied to images and text, thus overlooking other areas of AI. %
Similarly, whenever uncertainty estimates are required, they are usually approximated with ensemble methods that are unintelligible to humans. %
But tabular records are prevalent in real-life applications~\citep{shwartz2022tabular} and \emph{ante hoc} AI tends to achieve superior performance on structured data~\citep{rudin2019stop}. %
Such models, however, %
lack dedicated techniques for robust aleatoric and epistemic uncertainty quantification~\citep{rudin2022interpretable}. %
Additionally, %
\emph{ante hoc} interpretability may in itself be insufficient to engender understanding in non-technical stakeholders, %
but tools capable of producing %
reliable human-centred explanatory insights (like counterfactuals) into their operation %
are largely missing~\citep{sokol2023reasonable}. %

Crucially, %
\emph{ante hoc} modelling and %
uncertainty quantification are complementary and synergistic. %
Transparent AI with known capabilities and well-defined model form %
delivers a strong foundation for robust uncertainty estimation and decomposition; %
these insights, in turn, capture the predictor's quality and limitations, %
further reinforcing its %
interpretability and accountability. %
While this coupling remains largely unexplored, it is not unexpected (and has been briefly noted in the literature~\citep{rudin2022interpretable}) %
given that the model class choice is highly consequential, affecting both predictions and uncertainty estimates~\citep{hullermeier2021aleatoric}. %
For example, AI models can produce different outputs in sparse (training) data regions depending on whether they account for the underlying data distribution~\citep{perello2016background}. %
Native quantification of uncertainty %
also allows to avoid numerous pitfalls of its \emph{post hoc} estimation -- %
a practice that is notably at odds with the core principles of, and detrimental to, \emph{ante hoc} modelling. %
We further elaborate on some of these ideas and provide illustrative examples in \autoref{sec:methodology:ante}. %

In addition to benefiting uncertainty quantification, %
\emph{ante hoc} %
AI can facilitate %
better understanding of the notion of similarity used by the model, %
helping %
to alleviate some of the aforementioned challenges with defining meaningful distance metrics. %
Uncertainty awareness, on the other hand, %
creates a systematic pathway for adoption of human-centred developments from across XAI, thus extending the comprehensibility of \emph{ante hoc} AI systems %
to non-technical stakeholders, e.g., through counterfactual insights as we demonstrate next. %

\subsection{Uncertainty Quantification as a Unifying Framework}\label{sec:methodology:cf}

Given the appeal and widespread adoption of counterfactuals, %
improving their retrieval is crucial. %
We demonstrate that uncertainty quantification is uniquely suited to this end and provides a principled unifying framework for counterfactual explainability, subsuming various \emph{ad hoc} definitions of the desiderata listed in \autoref{sec:related:exp}. %
Specifically, %
we express %
them through constraints on \emph{aleatoric} and \emph{epistemic uncertainty} -- rather than its aggregate measure -- %
which, as we show below, delivers %
many benefits. %
Additionally, %
our formalisation of selected counterfactual properties %
necessarily %
relies on instance similarity measured with distance in the feature space %
given %
that this concept is orthogonal to the notion of uncertainty; %
\autoref{apx:dist-proxy} explores the link between them in more detail. %

\paragraph{Validity} %
Going beyond crisp classification allows to better differentiate explanations, which can be especially insightful when dealing with more than two classes, where the interplay of their individual probabilities may be nuanced~\citep{sokol2020limetree}. %
A \emph{valid} counterfactual $\xCF$ can, %
for example, be achieved by maximising the conditional probability $p(\cplus | x)$ of the desired class $\cplus$~\citep{schut2021generating}: %
\begin{equation}\label{eq:ucf}
    \xCF = \argmax_{x \in \mathcal{X}} \; p(\cplus | x) %
    \text{.}
\end{equation}
Increasing $p(\cplus | x)$ will decrease the total uncertainty (\autoref{eq:tu}), hence its aleatoric (\autoref{eq:au}) and epistemic (\autoref{eq:eu}) parts; %
however, lack of validity due to either of the two components communicates different information about the explanation. %
For example, the total uncertainty $\TU(\xCF)$ of a counterfactual $\xCF$ %
can either come mostly from %
the aleatoric part $\AU(\xCF)$ -- indicating that the instance is located in a region with mixed labels -- %
or from the epistemic component $\EU(\xCF)$ -- signalling its placement in a sparse data region where the exact decision boundary shape is (yet) unclear. %
While optimising \autoref{eq:ucf} arguably improves \emph{plausibility} and \emph{discriminativeness} too -- since the former can be linked to epistemic and the latter to aleatoric uncertainty~\citep{schut2021generating} -- %
this is not necessarily the case as we demonstrate below. %
A more comprehensive perspective on \emph{validity} is therefore necessary, %
which we introduce %
through fine-grained constraints on other counterfactual desiderata. %

\paragraph{Similarity and Sparsity} %
The %
standard, %
distance-based %
interpretation of counterfactual \emph{similarity} and \emph{sparsity} %
positions these properties as complementary to the notion of uncertainty. %
They can thus be implemented independently of it, e.g., by minimising the Euclidean and Manhattan distance metrics. %
Note, however, that these characteristics can, to a degree, be reflected in uncertainty estimates, for example, when the data manifold shape captures feature (inter)dependence; %
we explore this idea further in \autoref{apx:dist-proxy}. %
While a \emph{similar} counterfactual %
can be interpreted as a point lying very close to a decision boundary, on its opposite side to the factual instance, %
hence having high (total) uncertainty, i.e., %
$$
\xCF = \argmax_{x \in \mathcal{X}} \; p(\cplus | x) \quad \text{s.t.} \quad \TU(x) \approxeq 1
\text{,}
$$
this na\"ive approach has obvious shortcomings. %

\paragraph{Plausibility}
Epistemic uncertainty of a counterfactual %
instance %
quantifies the instability of the decision boundary in its vicinity; %
in many cases this phenomenon arises due to sparse training data in that region. %
Epistemic uncertainty thus captures %
how close a counterfactual is %
to the data manifold, offering a direct measure of its %
\emph{plausibility}~\citep{schut2021generating}. %
While the \emph{validity} formalisation given in \autoref{eq:ucf} indirectly minimises epistemic uncertainty, %
expressing \emph{plausibility} explicitly in its terms offers more control and flexibility: %
$$
    \xCF = \argmax_{x \in \mathcal{X}} \; p(\cplus | x) \quad \text{s.t.} \quad \EU(x) \approxeq 0
    \text{.}
$$
Notably, this definition relaxes the strong assumption that \emph{atypical} observations are impossible -- which is frequently made in counterfactual research, e.g., through direct reliance on data observed before~\citep{poyiadzi2020face} -- %
instead quantifying their \emph{improbability}. %
Generating %
explanatory insights that span a spectrum of \emph{plausibility} %
is highly beneficial from a human-centred perspective as %
it increases novelty and informativeness of explanations (by capturing insights that users may not know) as well as mitigates confirmation bias~\citep{jin2023rethinking}. %
This approach also %
improves explanation \emph{privacy} %
given that it does not operate directly on previously seen data~\citep{small2023counterfactual}. %

\begin{figure}
    \centering
    \includegraphics[scale=0.36]{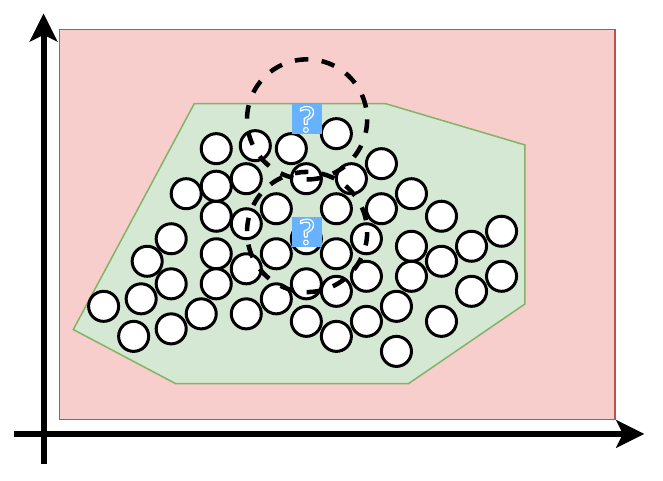}
    \caption{%
        According to \emph{connectedness}, %
        a counterfactual -- depicted by the question mark -- located centrally %
        in an area of low epistemic uncertainty -- indicated by the green background -- %
        is preferred %
        given its better alignment with the data manifold; %
        the red background represents high epistemic uncertainty, i.e., a sparse (training) data region. %
    }\label{fig:ill_connectedness}
\end{figure}

\paragraph{Connectedness}
Our definition of \emph{plausibility} %
is also the foundation of \emph{connectedness}. %
Here, we quantify %
epistemic uncertainty %
in the \emph{area} surrounding the counterfactual instance, %
seeking %
a point placed \emph{centrally} in a region of low epistemic uncertainty. %
This is desirable as %
the mean epistemic uncertainty of the neighbourhood of such an instance is lower -- %
and the predicted probability of the counterfactual class $\cplus$ higher -- %
than that of a point situated at the fringes of this region; %
\autoref{fig:ill_connectedness} illustrates these two scenarios for a toy example. %
While either location by itself %
ensures \emph{plausibility} given its closeness to the data manifold, %
the former placement makes the counterfactual more \emph{connected} than the latter. %

Formally, we minimise the epistemic uncertainty of $\Ball(\xCF, \rho)$ %
rather than $\xCF$ alone, %
where $x^\prime \sim \Ball(x, \rho)$ is a uniformly sampled instance from a hyper-sphere of radius $\rho > 0$ centred on $x$, i.e., $\Ball(x, \rho) = \{ x^\prime \in \mathcal{X} \; | \; d(x, x^\prime) \leq \rho\}$ for a selected distance metric $d$. %
We can implicitly specify the desired behaviour of epistemic uncertainty around $\xCF$ %
by choosing the function that aggregates %
its measures for the sampled data; %
for example, %
the minimum %
$\min[\cdot]$ and %
maximum %
$\max[\cdot]$ operators could be the extreme choices, %
with %
the standard deviation $\sqrt{\mathds{V}[\cdot]}$ and %
expectation $\mathds{E}[\cdot]$ being more moderate options. %
While we pick $\mathds{E}[\cdot]$ in our definition of \emph{connectedness} %
as the balanced approach: %
$$
    \xCF = %
    \argmax_{x \in \mathcal{X}} \; p(\cplus | x) %
    \quad %
    \text{s.t.}
    \quad %
    \mathds{E}_{x^\prime \sim \Ball(x, \rho)} \left[ \EU(x^\prime) \right] \approxeq 0 \text{,} %
$$
choosing $\max[\cdot]$ can, for instance, yield better results if the model struggles to capture epistemic uncertainty, thus outputting its low values. %

\paragraph{Discriminativeness}
High predicted probability of the desired class $\cplus$ is a prerequisite of \emph{discriminativeness} but can be insufficient in itself; %
for example, %
one may wish to %
explicitly differentiate the counterfactual instance from \emph{all} the other classes instead of only the factual one, %
which requires a more nuanced approach than permitted by the \emph{validity} formalisation given in \autoref{eq:ucf}. %
Depending on how the remaining probability mass is allocated among different classes we can distinguish two broad scenarios: %
\begin{enumerate*}[label=(\roman*)]
    \item \label{list:disc:one}
    one class, e.g., the factual $\cplus$, receives most of it, or %
    \item \label{list:disc:many}
    it is distributed roughly equally. %
\end{enumerate*}
Regarding \cref{list:disc:one}, %
low aleatoric uncertainty of the counterfactual curbs its \emph{ambiguity} across all classes, inspiring the following formalisation of discriminativeness: %
$$
\xCF = \argmax_{x \in \mathcal{X}} \; p(\cplus | x) \quad \text{s.t.} \quad \AU(x) \approxeq 0
\text{.}
$$
Regarding \cref{list:disc:many}, %
we can explicitly reduce the second-highest probability, redistributing its mass among the remaining classes like so: %
$$
    \xCF = %
    \argmax_{x \in \mathcal{X}} \; p( \cplus | x) %
    \quad %
    \text{s.t.} \quad \max_{\cplus \neq c \in \mathcal{C}} \; p(c | x) \; \text{is minimal.}%
$$

\paragraph{Robustness} %
As more training data are collected, especially in regions that were sparsely populated thus far, the decision boundary of a model may shift, %
negatively impacting \emph{validity} as well as other properties of counterfactuals. %
Since epistemic uncertainty quantifies the instability of the decision boundary in this context %
(as discussed earlier for \emph{plausibility}), %
minimising it will boost explanation \emph{robustness}; %
this improvement will be especially pronounced for subspaces where only relatively few observations are available. %
Striving for low aleatoric uncertainty is also beneficial: %
when new data arrive, the decision boundary in regions with class overlap may too be further refined. %

Drawing inspiration from the formalisation of \emph{validity} given in \autoref{eq:ucf}, %
we can thus minimise the \emph{total} uncertainty %
instead of separately optimising $\AU(\xCF)$ and $\EU(\xCF)$. %
Moreover, following reasoning similar to that outlined for \emph{connectedness}, %
rather than minimising $\TU(\xCF)$ directly, we can do so in the neighbourhood of the counterfactual $\xCF$. %
Consequently, we define uncertainty-based \emph{robustness} as: %
$$
    \xCF
    =
    \argmax_{x \in \mathcal{X}} \; p(\cplus | x)
    \quad
    \text{s.t.}
    \quad
    \mathds{E}_{x^\prime \sim \Ball(x, \rho)} \left[ \TU(x^\prime) \right] \approxeq 0 %
\text{.}
$$
Notably, this approach comes with numerous benefits from the human-centred perspective. %
For example, optimising this objective %
can alleviate adverse effects of imprecision with which humans may implement counterfactuals -- a factor that they do not necessarily fully control~\citep{xuan2024perfect}. %

\paragraph{Stability} %
While \emph{stability} is primarily considered a property of the counterfactual generation \emph{process} rather than the explanation itself~\citep{guidotti2022counterfactual}, %
our formalisation of \emph{robustness} is well aligned with this desideratum. %
This is highly beneficial %
given that \emph{stability} %
-- as it is characterised by its evaluation metric $m_{\text{sta}}(\cdot, \cdot)$ outlined in \autoref{apx:eval-metrics} -- %
is built upon distance measured in the feature space, %
making its uncertainty-based definition impractical. %

\paragraph{Feasibility}
A counterfactual $\xCF$ can be constructed upon $n > 1$ steps $\xCF_1, \ldots, \xCF_n$, %
with the first being the factual point, i.e., $\xCF_1 \equiv \xF$, and the last being the counterfactual instance, i.e., $\xCF_n \equiv \xCF$. %
The %
proximity of these individual points to the data manifold, or their presence in the training set, %
ensures \emph{feasibility} of the explanation~\citep{poyiadzi2020face}. %
Building atop our formalisation of \emph{plausibility}, %
we can use epistemic uncertainty to quantify \emph{feasibility} of each such step; %
additionally, %
following reasoning similar to that outlined for \emph{connectedness}, %
instead of minimising $\EU(\xCF)$ directly, we can do so in the neighbourhood of the counterfactual $\xCF$. %
The resulting definition of \emph{feasibility} is: %
\begin{multline*}
    \forall i \in \{1,\ldots,n\} \;\; \xCF_i = %
    \argmax_{x \in \mathcal{X}} \; p(\cplus | x) %
    \\%\quad %
    \text{s.t.}
    \quad
    \mathds{E}_{x^\prime \sim \Ball(x, \rho)} \left[ \EU(x^\prime) \right] \approxeq 0
\text{.}
\end{multline*}
Note that other counterfactual desiderata can be easily adapted to this path-based paradigm -- a topic we explore further in \autoref{sec:methodology:extra}. %

\paragraph{Actionability} %
Without access to sequential or time-series data, inferring \emph{mutability} of features %
as well as %
the \emph{directionality}, \emph{monotonicity} and \emph{rate} of their change %
-- including any such relations between individual attributes -- %
is largely infeasible. %
\emph{Actionability} thus requires dedicated (manual) annotations, which could possibly be formalised in terms of uncertainty constraints. %

\subsection{Uncertainty-aware Counterfactual Generation}\label{sec:methodology:gen}

Having introduced our definitions of counterfactual desiderata, %
we now propose a method for generating \emph{uncertainty-aware explanations} %
that aggregates these properties within a single objective function. %
Specifically, we formalise two distinct realisations of our explainer: %
one that is solely based on \emph{uncertainty} estimates, and another that supplements them with \emph{distance} measured in the feature space given the complementarity of these two concepts. %
Our optimisation loss $\ell$ is thus defined as: %
\begin{equation}\label{eq:u-all}
\ell = %
\begin{cases}
  \begin{aligned}
    \log p(\cplus | \xCF) -  &\lambda_1 \exp \EUi{B} \\ %
            \color{pblue}{+} &\color{pblue}{\lambda_2 \exp \AUi{\max}}
  \end{aligned}
    \\\qquad\qquad\qquad\qquad\qquad\qquad%
    \text{if} \;\; p(\cplus | \xCF) < 0.5 \\\\ %
  \begin{aligned}
    \log p(\cplus | \xCF) -  &\lambda_1 \exp \EUi{B} %
                          -  \lambda_2 \exp \AUi{B} \\ %
            \color{pblue}{+} &\color{pblue}{\lambda_2 \exp \AUi{\max}} \\
          \color{pyellow}{-} &\color{pyellow}{\lambda_2 (||\xF-\xCF||_1 + ||\xF-\xCF||_2 )}
  \end{aligned}
    \\\qquad\qquad\qquad\qquad\qquad\qquad%
    \text{if} \;\; p(\cplus | \xCF) \geq 0.5 %
\text{,}
\end{cases}
\end{equation}
where the terms in black are used by both variants, the expressions in \textcolor{pblue}{blue} only apply to \emph{pure uncertainty-driven counterfactuals}, and %
the formulation in \textcolor{pyellow}{yellow} %
underpins the \emph{uncertainty plus distance} explainer. %

We introduce the latter approach since counterfactual desiderata such as \emph{similarity} and \emph{sparsity} cannot be directly quantified with uncertainty as discussed earlier in \autoref{sec:methodology:cf}; %
it replaces the implicit representation of distance via maximum aleatoric uncertainty $\AUi{\max}$ -- marked in \textcolor{pblue}{blue} in \autoref{eq:u-all} -- with the $||\cdot||_1$ (Manhattan or $L_1$) and $||\cdot||_2$ (Euclidean or $L_2$) norms -- shown in \textcolor{pyellow}{yellow} -- commonly used to capture these two properties. %
The meaning of each individual component found in our loss $\ell$ is clarified in \cref{alg:uncertainty-cf}, which %
formalises the optimisation procedure underlying our explainers. %

\begin{algorithm}[t]
\caption{%
    Uncertainty-based counterfactual generation. %
}\label{alg:uncertainty-cf}
\begin{algorithmic}[1]
    \Statex \textbf{Input:} %
    factual instance $\xF$; %
    counterfactual target class $\cplus$; %
    desired probability $p^\cpos$ of the target class $\cplus$; %
    explained probabilistic model $H$ (as defined in \autoref{sec:related:uq}) that %
    outputs probability $p(c|x)$ of class $c$ for data point $x$ based on its total uncertainty $\TU$ estimate (\autoref{eq:tu}) and %
    also delivers %
its decomposition into aleatoric $\AU$ (\autoref{eq:au}) and epistemic $\EU$ (\autoref{eq:eu}) parts; %
    neighbourhood sampling function $\Ball(x, \rho)$ (as defined in \autoref{sec:methodology:cf}); %
    neighbourhood size $n$; %
    iteration limit $t_{\max}$; and %
    step size $\eta$. %
    \Statex \textbf{Output:} counterfactual data point $\xCF$.
    \State initialise $\xCF \gets \xF$ and $t \gets 0$
    \While{$p(\cplus | \xCF) < p^\cpos$ and $t < t_{\max}$ } %
        \State sample $B \gets \{ x \; | \; x \sim \Ball(\xCF, \rho) \}$ with $|B| = n$ %
        \State compute aggregated epistemic uncertainty $\EUi{B} \gets \frac{1}{n} \sum_{x \in B} \EU(x)$ %
        \State compute aggregated aleatoric uncertainty $\AUi{B} \gets \frac{1}{n} \sum_{x \in B} \AU(x)$ %
        \State compute maximum aleatoric uncertainty $\AUi{\max} \gets \max_{x \in B} \AU(x)$ %
        \State compute loss $\ell$ (\autoref{eq:u-all}) %
        \State compute gradients $g \gets \nabla_{\xCF} \ell$ %
        \State compute $\rho$-Ball gradients $g_B \gets \sum_{x \in B} \nabla_{\xCF} \ell$ %
        \State update counterfactual $\xCF \gets \xCF + \eta (g + g_B)$
        \State update step count $t \gets t + 1$
    \EndWhile
    \State \Return $\xCF$
\end{algorithmic}                       
\end{algorithm}

Note that %
the %
uncertainty estimates %
in \autoref{eq:u-all} are \emph{exponentiated} %
since in the early steps of the optimisation loop %
the probabilities $p(\cplus | \xCF)$ output by the explained model for the candidate counterfactual $\xCF$ take extremely low values, %
hence they would otherwise %
dominate the gradients given that they are passed through the logarithmic function. %
The added benefit of this procedure is having epistemic uncertainty $\EUi{B}$ become the biggest gradient contributor for its large values, %
thus helping to better steer counterfactual generation. %

Formulating the loss as a two-step optimisation procedure %
further allows us to preserve %
individual counterfactual desiderata while maintaining low epistemic and aleatoric uncertainty overall. %
Specifically, in the beginning of the optimisation loop -- when $p(\cplus | \xCF) < 0.5$ -- %
the loss focuses on maximising the target class probability $p(\cplus | \xCF)$ while simultaneously minimising epistemic uncertainty $\EUi{B}$; %
afterwards -- when 
$p(\cplus | \xCF) \geq 0.5$ -- aleatoric uncertainty $\AUi{B}$ is being minimised too. %
At both stages these uncertainty estimates are evaluated in the neighbourhood of the counterfactual candidate $\xCF$ rather than at its exact location, %
using the $\Ball(\xCF, \rho)$ function (defined in \autoref{sec:methodology:cf}) to this end. %
In essence, %
reducing epistemic uncertainty from the outset delivers \emph{feasibility}, \emph{plausibility} and \emph{connectedness}; %
then, lowering aleatoric uncertainty additionally helps to achieve \emph{discriminativeness} and \emph{robustness}. %
$\lambda_1$ and $\lambda_2$ are weighting hyperparameters that control the trade-off between the two. %

\section{Experiments}\label{sec:experiments}%

We take a two-step evaluation approach. %
We first assess our uncertainty-based definitions of counterfactual desiderata; %
then, we proceed to validating the two variants of our uncertainty-aware explainer. %
We compare the latter %
to four distinct state-of-the-art methods: %
Growing Spheres (GS)~\citep{laugel2018comparison}, %
DICE~\citep{mothilal2020explaining}, %
FACE~\cite{poyiadzi2020face} and %
CLUE~\citep{antoran2020getting}. %
The first three are ignorant of uncertainty, whereas the last one %
minimises the total uncertainty of a counterfactual using a variational autoencoder to this end. %
We evaluate these explainers as well as the desiderata with six quantitative metrics commonly used in counterfactual research: %
validity, similarity, sparsity, plausibility, discriminativeness and stability~\citep{guidotti2022counterfactual}; %
their definitions are provided in \autoref{apx:eval-metrics}. %
In our experiments, we %
use %
nine data sets %
that are popular in XAI studies: %
bank, (breast) cancer, churn, COMPAS, diabetes, FICO, home, housing and titanic~\citep{guidotti2022counterfactual}. %
We pre-process the data by discarding their non-continuous features; %
the hyperparameters of FACE, CLUE and our uncertainty-based approach %
are optimised %
over 50 sweeps, %
choosing a setting that achieves the best average performance across the aforementioned metrics. %
\autoref{apx:tables} provides detailed evaluation results, %
with this section reporting findings aggregated across the data sets. %

To deliver reliable and robust aleatoric and epistemic uncertainty estimates %
we employ the \emph{DARE} ensemble method, %
which leverages weight regularisation to this end~\citep{mathelin2023deep}. %
As a form of ablation we also test the \emph{deep}~\citep{lakshminarayanan2017simple} and \emph{adversarial}~\citep{schut2021generating} ensembles, with the corresponding results -- reported in \autoref{apx:exp-full} -- aligning with those provided by \emph{DARE}. %
The ensembles consist of 20 neural networks with two hidden layers, each composed of 100 neurons. %
For the implementation of the GS, DICE, FACE and CLUE counterfactual explainers we rely on the CARLA library~\citep{pawelczyk2021carla}. %
Our experiments are coded in PyTorch, using the \emph{adam} optimiser with its default settings and %
\emph{cosine annealing} as the scheduler~\citep{kingma2015adam,paszke2019pytorch}; %
we train for maximum 50 epochs with the batch size set to 256. %
Our code is available on GitHub.\footnote{\github{}} %

\begin{figure*}%
    \centering
    \includegraphics[width=\linewidth]{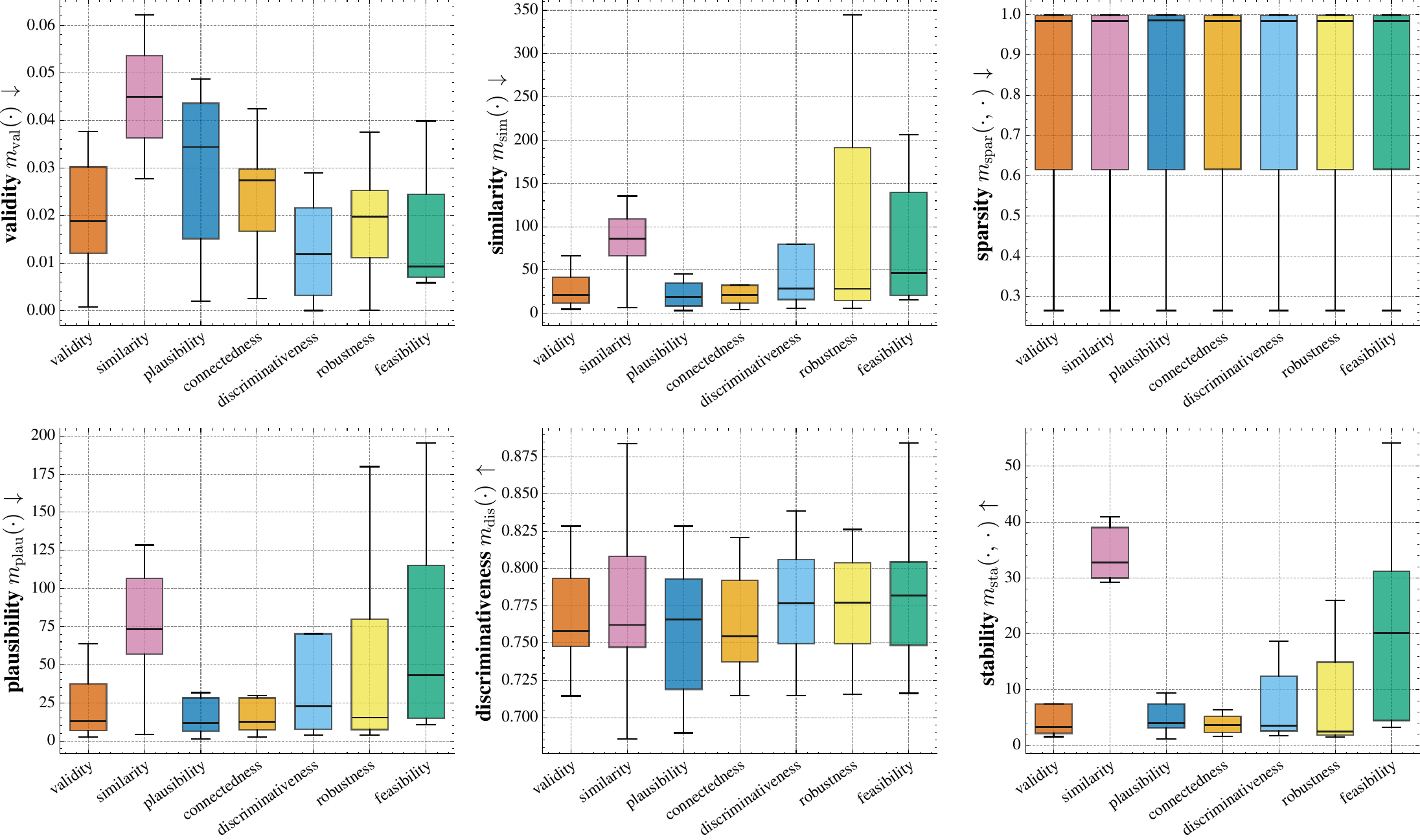}
    \caption{%
      Evaluation of our %
      uncertainty-based counterfactual desiderata (x-axis) -- as defined in \autoref{sec:methodology:cf} -- %
      for the six metrics (one per pane, y-axis) listed in \autoref{apx:eval-metrics}. %
      The experiments use the DARE ensemble, with the results aggregated across the nine data sets. %
    }\label{fig:dare_properties}
\end{figure*}

\begin{figure*}%
    \centering
    \includegraphics[width=\linewidth]{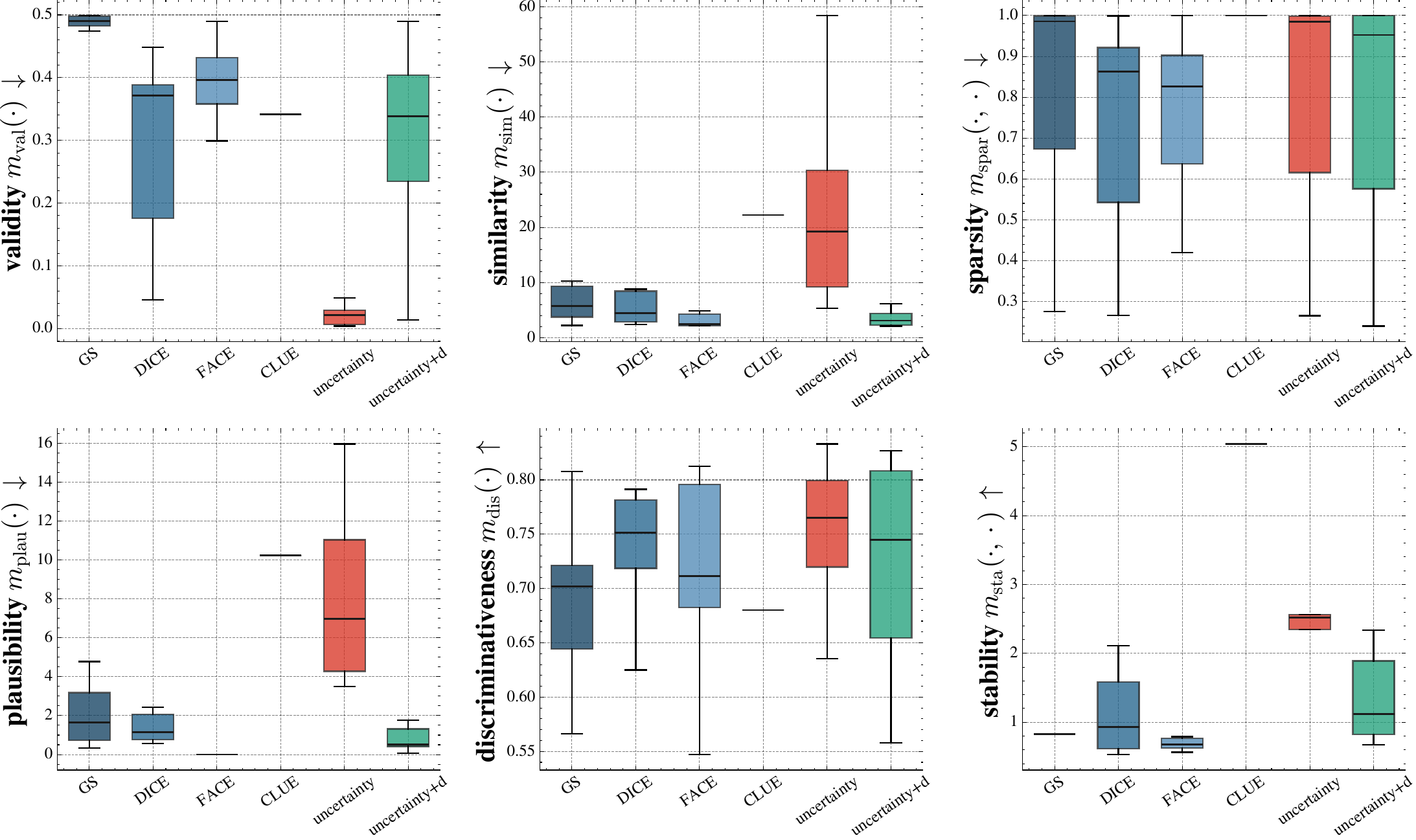}
    \caption{%
      Evaluation of counterfactual explainers (x-axis) -- with \emph{uncertainty} and \emph{uncertainty plus distance} (uncertainty+d) as defined in \autoref{sec:methodology:gen} -- %
      for the six metrics (one per pane, y-axis) listed in \autoref{apx:eval-metrics}. %
      The experiments use the DARE ensemble, with the results aggregated across the nine data sets. %
    }\label{fig:overall_dare_ensemble}
\end{figure*}

\paragraph{Property Validation}%
\autoref{fig:dare_properties} presents evaluation outcomes -- aggregated across the data sets -- for our uncertainty-based formalisation of each individual counterfactual de\-sid\-e\-ra\-tum, %
with the %
full results provided in \autoref{tab:dare_ensemble_results_properties} found in \autoref{apx:tables}. %
The definitions of \emph{connectedness} and \emph{robustness} show improved \emph{stability} scores given that they explicitly account for the neighbourhood of $x$ through the $\Ball(x, \rho)$ function. %
While our notion of \emph{feasibility} uses $\Ball(x, \rho)$ too, %
it does not show this behaviour %
because %
it pertains to a collection of points constituting a counterfactual path rather than a specific instance. %
As anticipated, our attempt to approximate the \emph{similarity} property -- which intrinsically relies on distance -- through total uncertainty %
severely underperforms on the \emph{similarity} metric %
since %
there is an inherent trade-off in simultaneously maximising both $p(\cplus | \xCF)$ and $\TU(\xCF)$. %
This observation motivates the \emph{uncertainty plus distance} variant of our explainer, which, as we show below, achieves better overall performance. %

As expected, %
the properties formalised through epistemic uncertainty, namely, \emph{plausibility} and \emph{connectedness}, outperform all the others in terms of the \emph{plausibility} metric. %
Accounting for the epistemic component explicitly yields better results than doing so implicitly through total uncertainty, %
with the latter reflected in the \emph{plausibility} score for the %
\emph{validity} desideratum (\autoref{eq:ucf}). %
An exception here is \emph{feasibility} for the exact same reason as described in the previous paragraph. %
In terms of raw discriminative power -- captured by the \emph{discriminativeness} metric -- %
the properties that rely purely on epistemic uncertainty, i.e., \emph{plausibility} and \emph{connectedness}, fare somewhat better than those accounting for its aleatoric component. %
Based on these results we can see that our formalisation of counterfactual desiderata is overall well aligned with evaluation metrics currently used for this explanation type. %

\begin{figure*}[t]
    \centering
    \includegraphics[width=\linewidth]{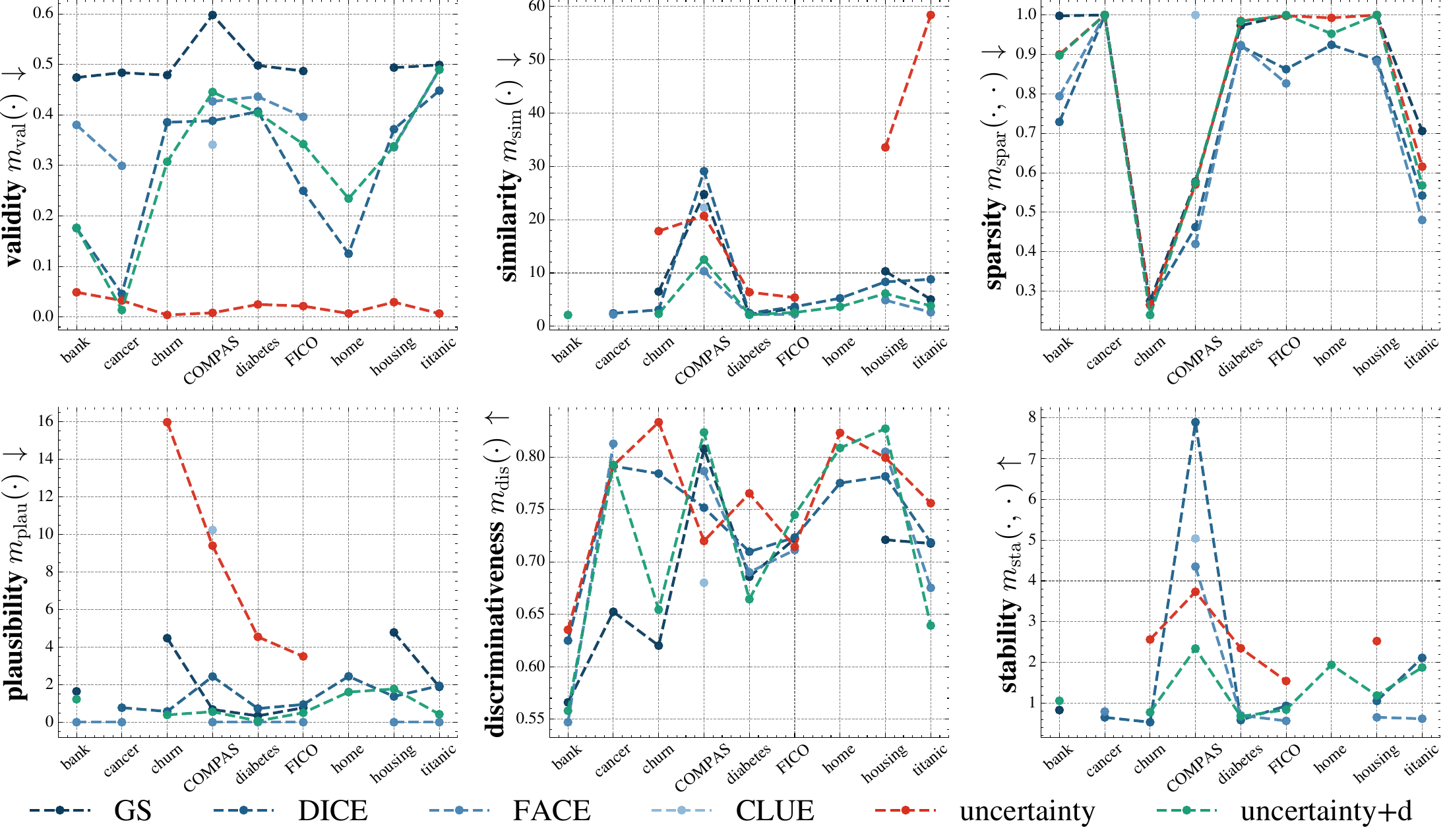}
    \caption{%
      Evaluation of counterfactual explainers %
      -- as described by the legend, with \emph{uncertainty} and \emph{uncertainty plus distance} (uncertainty+d) as defined in \autoref{sec:methodology:gen} -- %
      across the nine data sets (x-axis) %
      for the six metrics (one per pane, y-axis) listed in \autoref{apx:eval-metrics}. %
      The experiments use the DARE ensemble. %
      Note the missing results; %
      for CLUE these are due to it failing to find a counterfactual %
      and for the other explainers because of discarded outliers (i.e., values outside the interquartile range), %
      which were filtered out to improve the readability of the plots.
    }\label{fig:dare_individual_datasets}
\end{figure*}

\paragraph{Explainer Validation}%
\autoref{fig:overall_dare_ensemble} shows evaluation outcomes -- aggregated across the data sets -- for the two variants of our uncertainty-based counterfactual explainer, comparing them to the four selected baseline methods; %
the full results are provided in %
\cref{tab:dare_ensemble_results_explainers} located in \autoref{apx:tables}. %
We find %
that our \emph{uncertainty-only} method %
yields better overall \emph{validity} and \emph{discriminativeness} %
but exhibits lacklustre \emph{similarity}, \emph{sparsity} and \emph{stability}, %
which is expected given the dependence of the latter three metrics on the notion of distance in the feature space (refer to \autoref{apx:eval-metrics}). %
This variant of our explainer also %
fares poorly on \emph{plausibility} %
since this evaluation criterion too relies on distance, albeit indirectly through the $\Ball(\xCF, \rho)$ function. %
Our %
various %
attempts to %
adjust uncertainty measures to compensate for the lack of distance measurements %
to further optimise \emph{similarity} and \emph{discriminativeness} did not produce meaningful improvements in either metric. %

The \emph{uncertainty plus distance} variant of our explainer, however, %
makes up for these shortcomings, %
markedly improving \emph{similarity}, \emph{plausibility} and \emph{stability}; %
while \emph{sparsity} remains unaffected, %
the value of this metric is on a par with the baselines. %
Notably, %
achieving this boost in performance requires trading off some %
\emph{validity} and \emph{discriminativeness}, nonetheless the values of these metrics remain %
in line with the baselines. %
Overall, despite its fundamental simplicity, this variant of our explainer is highly competitive and delivers a more principled counterfactual generation process at negligible cost. %

Investigating the explainer evaluation results on the level of each individual data set -- shown in \autoref{fig:dare_individual_datasets} -- paints a more complicated picture. %
On \emph{stability}, the distance-aware variant of our explainer is the best; %
with respect to \emph{plausibility}, it is comparable to DICE while sometimes matching %
FACE, which sources counterfactuals directly from the training set. %
Disregarding the \emph{housing} data set, %
both variants of our explainer demonstrate unparalleled \emph{discriminativeness}. %
While our uncertainty-only approach exhibits somewhat high variability, %
making it distance-aware stabilises its performance. %

\section{Discussion}\label{sec:discussion}

Our framework for uncertainty-based counterfactual desiderata and explainability shows considerable potential despite the inherent simplicity of its high-level design. %
In this section we complement our quantitative experiments with %
a comprehensive discussion of this approach. %
First, we draw attention to the %
strong connection between uncertainty quantification and \emph{ante hoc} interpretability, %
which promises to enhance the techniques introduced throughout this paper~(\sref{sec:methodology:ante}). %
Then, we explore %
additional counterfactual desiderata that could be integrated into our uncertainty-based framework to further improve explanation quality; %
we also review the roles and implications of as well as alternatives for using distance measured in the feature space when generating counterfactuals~(\sref{sec:methodology:extra}). %
Lastly, we analyse %
the broader implications %
of the proposed approach, %
focusing on various topics spanning %
uncertainty quantification, %
\emph{ante hoc} interpretable modelling and %
human-centred XAI~(\sref{sec:methodology:anti}). %
Taken together, %
these insights %
show that additional research is needed to overcome %
the identified challenges and facilitate progress in this area. %

\subsection{Uncertainty Quantification and \emph{Ante Hoc} Interpretability}\label{sec:methodology:ante}%

Given their technological nature, %
AI models %
should not be accepted based on the trust they engender, %
e.g., after prolonged interactions, %
but rather due to their operational reliability and robustness~\citep{ryan2020ai}; %
the need for %
their sound technical design and functioning %
is clearly visible in high stakes domains~\citep{rudin2019stop}. %
Notably, %
these principles are embodied by %
(\emph{ante hoc}) interpretability and uncertainty quantification, %
which allow humans to develop correct understanding of models' capabilities and limitations~\citep{schut2021generating}, leading to accountability and trust~\citep{tomsett2020rapid}. %
While uncertainty is generally used to capture the quality of AI models and their predictions, %
it is important not to overlook its second-order notion as well as decomposition into the aleatoric and epistemic parts %
since they convey fundamentally distinct information. %

Based on our %
brief introduction of \emph{ante hoc} interpretability in \autoref{sec:related:exp} and %
comprehensive %
review of %
uncertainty in \autoref{sec:related:uq}, %
we posit here that %
these concepts %
are not only complementary~\citep{schut2021generating} %
but rather constitute different views of the same underlying idea -- %
with the former being necessary for a strong notion of the latter and the latter enhancing the former. %
Recall that \emph{ante hoc} interpretability offers %
inherently robust, accountable and transparent AI models %
by constraining their form~\cite{rudin2019stop}. %
Specifying the modelling assumptions explicitly, in turn, allows for %
reliable uncertainty quantification and decomposition %
given how brittle these processes can be (refer back to \autoref{fig:model}). %
While \citet{rudin2022interpretable} have briefly noted the link between these two concepts, there is still much left to be explored. %

In particular, we find %
this shared foundation %
to be critical given that %
\emph{ad hoc} uncertainty quantification %
could produce %
incorrect %
measurements %
due to its possible incompatibility with (implicit) data modelling assumptions and technical functioning of AI models -- %
akin to how \emph{post hoc} XAI may not %
truthfully capture how the predictor being inspected actually operates. %
Consequently, explainers deployed in such settings need to compensate for the shortcomings or lack of (native) uncertainty estimates, %
which often increases the complexity of these tools and degrades explanation quality. %
This approach %
would %
be especially harmful, and counterproductive, when dealing with \emph{ante hoc} interpretable models. %

For example, consider the fundamental differences in the decision boundary shape learnt by linear models and decision trees as well as their impact on the resulting predictions and uncertainty estimates. %
For the former model type, these properties depend largely on the distance from the decision boundary, whereas for the latter, %
the relation is more nuanced and comes down to the hyper-rectangle partition of the feature space as well as the quantity and class distribution of the training data used to learn it. %
A slight modification to a model class %
-- e.g., bounding the hyper-rectangles of a decision tree from all sides -- %
could %
thus %
yield drastically different predictions and uncertainty estimates. %

Likewise, %
some model classes capture the training data distribution, whereas other do not, leading to different handling of instances located in sparse data regions. %
While post-processing techniques can enhance a particular model class in this respect~\citep{perello2016background}, %
such an approach introduces asymmetry between the modelling assumptions and properties of the final predictions. %
If the latter are used in downstream tasks, this may be undesirable; %
more broadly, it signals that another model class fulfilling one's requirements ought to be considered instead~\citep{rudin2024amazing}. %
Similarly, reducing epistemic uncertainty of models deemed to be universal approximators can be particularly challenging exactly because of their inherent expressiveness. %
Here, a popular solution is to introduce strong model form regularisation -- an approach that bears the hallmarks of \emph{ante hoc} interpretability. %
Current XAI research, nonetheless, often overlooks the quality (including predictive power) of a model as well as the importance and implications of a model class choice; %
this is especially true for \emph{post hoc} methods but also affects, albeit to a much lesser extent, \emph{ante hoc} interpretability. %

Crucially, reliable uncertainty estimates open the door to a broad range of explanatory insights; %
\autoref{sec:methodology:cf} demonstrates this for the case of counterfactuals.
Such %
uncertainty-guided tools that produce human-centred explanatory insights are of particular relevance %
for \emph{ante hoc} interpretable models %
given their incomprehensibility to non-technical stakeholders, which likely harms their real-life adoption. %
Another %
big upside of this approach is %
explanation consistency across models (whether from the same or different model classes) given that modelling assumptions are directly accounted for in explanation generation. %
This is arguably more desirable than explanation consistency with respect to an explainer (applied to distinct model classes) %
since its operational characteristics are unlikely to be compatible with a broad range of unique modelling assumptions. %

\begin{figure*}[t]%
  \centering
  \begin{subfigure}[t]{0.470\textwidth}%
      \centering
      \includegraphics[scale=0.36]{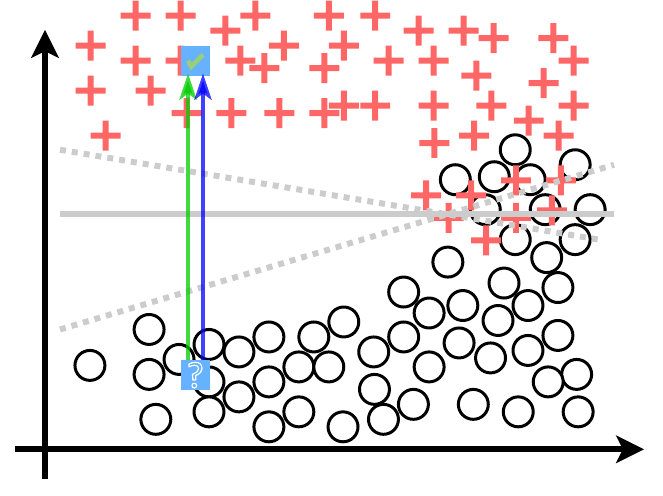}%
      \caption{%
        Linear model.
      }\label{fig:example-p:linear}%
  \end{subfigure}
  \hspace{0.01\textwidth}
  \begin{subfigure}[t]{0.470\textwidth}%
      \centering
      \includegraphics[scale=0.36]{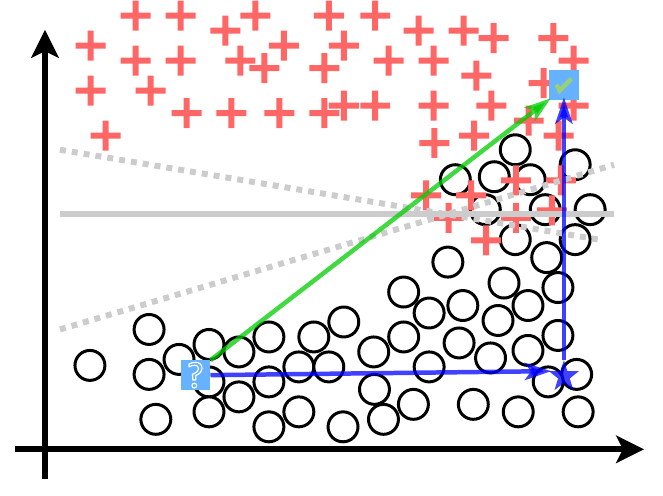}%
      \caption{%
        Data manifold-aware linear model. %
      }\label{fig:example-p:linear+bg}%
  \end{subfigure}
  \caption{%
    Illustration of uncertainty-driven instance-based (green) and path-based (blue) counterfactuals for %
    (\subref{fig:example-p:linear})~a standard linear model and %
    (\subref{fig:example-p:linear+bg})~one enhanced with a background class~\citep{perello2016background}. %
    For the former, %
    following the direction perpendicular to the decision boundary is the optimal approach; %
    for the latter, %
    the counterfactual vector and path are skewed since epistemic uncertainty additionally accounts for the shape of the data manifold. %
  }\label{fig:example-p}%
\end{figure*}

Optimal, in terms of uncertainty, explanations may thus necessarily be dissimilar in absolute terms across distinct modelling scenarios -- e.g., different counterfactual instances -- but at the same time consistent with regard to their high-level desiderata as demonstrated in \autoref{fig:example-p}. %
Additionally, %
founding explanation retrieval upon uncertainty %
somewhat %
alleviates the aforementioned challenges of defining a meaningful (for the problem at hand) distance metric that is also consistent with %
how the underlying model quantifies similarity in the feature space (see \autoref{sec:methodology:challenges} for more details). %
Consequently, building AI systems that are uncertainty-aware, reliable and robust because they are fundamentally suitable for a particular application appears more important than creating ``universal'', thus overly complex, explainers. %

This section's focus %
on \emph{ante hoc} interpretability is rooted in its rich history of success and reported superiority for structured data~\citep{rudin2019stop,shwartz2022tabular}; %
while the latter claim can be challenged~\citep{wang2021hybrid}, dismissing inherently transparent AI altogether without further evidence seems na\"ive and premature. %
Additionally, %
one could argue that %
using auxiliary explanatory mechanisms to inspect \emph{ante hoc} interpretable models inadvertently breaks this property, %
but %
in our case %
this position can be easily refuted since the main focus remains on robust data modelling, with counterfactuals simply being a by-product of access to reliable uncertainty estimates. %
Crucially, counterfactual explainability can be viewed through frameworks other than uncertainty quantification, most notably causality. %
This perspective promises more reliable explanations and is compatible with \emph{ante hoc} interpretability, however causal modelling is often unrealistic in practice as mentioned earlier in \autoref{sec:related:exp}. %
Lastly, %
while due to practical constraints our experiments relied on %
model ensembles to quantify uncertainty, %
our findings clearly demonstrate %
the foundational role of principled uncertainty handling and its relation to transparent modelling practices (see \autoref{sec:experiments} as well as \cref{apx:tables,apx:exp-full} for more details). %

\subsection{Extended Properties and Evaluation of Counterfactuals}%
\label{sec:methodology:extra}

All the properties described in \autoref{sec:related:exp}, except \emph{feasibility} (see its formalisation in \autoref{sec:methodology:cf} for more details), %
pertain to the counterfactual instance itself. %
Additional desiderata apply to collections of such explanations, which are employed because a single counterfactual is unlikely to satisfy the distinct needs and expectations of different explainees~\citep{sokol2020one}. %
In this context, \emph{diversity} %
ensures that a set of counterfactuals is representative of the available explanations while at the same time remaining minimal, i.e., as small as possible; %
note that, to a degree, \emph{similarity} naturally lends itself to generating \emph{diverse} counterfactuals. %
From the model-selection perspective, explanation \emph{availability} should also be considered~\citep{kanamori2024learning}. %
It maximises the number of individuals for whom at least one acceptable counterfactual can be generated, %
which is crucial when there exist multiple predictors (from a single model class) with comparable performance~\citep{sokol2024cross,rudin2024amazing}. %

Additionally, %
it is important to recognise %
a prominent extension of counterfactuals %
that %
builds %
action sequences %
guiding an explainee step-by-step towards the desired outcome -- an approach that delivers many human-centred and technical benefits~\citep{poyiadzi2020face,sokol2023navigating}. %
The aforementioned \emph{feasibility} is one relevant property that captures and formalises the characteristics %
of the entire path connecting the factual and counterfactual instances (rather than those of the latter point alone). %
Its exact operationalisation -- described in \autoref{sec:related:exp} -- %
entails building this link by following pre-existing data points, %
such as those found in the training set; %
note that this notion of \emph{feasibility} tends to imply \emph{plausibility}, \emph{connectedness} and \emph{stability} as well. %
Our uncertainty-based interpretation of this desideratum %
-- introduced in \autoref{sec:methodology:cf} -- %
relaxes the strict requirement of constructing counterfactual paths upon training data, %
instead imposing epistemic uncertainty constraints on these instances. %
This approach places those individual steps %
on the data manifold, making them highly likely rather than simply observed before. %

While frequently overlooked, %
many %
other %
counterfactual %
properties %
-- such as \emph{plausibility}, \emph{connectedness} and \emph{robustness} -- %
can be %
easily generalised to %
path-based explanations %
by %
simply applying them to %
the points constituting their individual steps; %
their further extension could rely on the neighbourhood of each of these data points via the $\Ball(\xCF_i, \rho)$ function (see \autoref{sec:methodology:cf}) rather than these instances themselves. %
From the uncertainty perspective, %
a desirable %
profile of a counterfactual path can %
be characterised by low epistemic uncertainty along its entirety, low aleatoric uncertainty towards its end and decision boundary crossings in regions of increased aleatoric uncertainty or a balanced mixture of both uncertainty types. %
Further specification %
could prevent %
sharp changes and discontinuities in per-class uncertainty measured along the path and %
ensure %
that it obeys monotonicity constraints such that a decision boundary between any two classes is only crossed once. %

\begin{figure*}[t]%
  \centering
    \begin{subfigure}[t]{0.310\textwidth}
      \centering
      \includegraphics[scale=0.36]{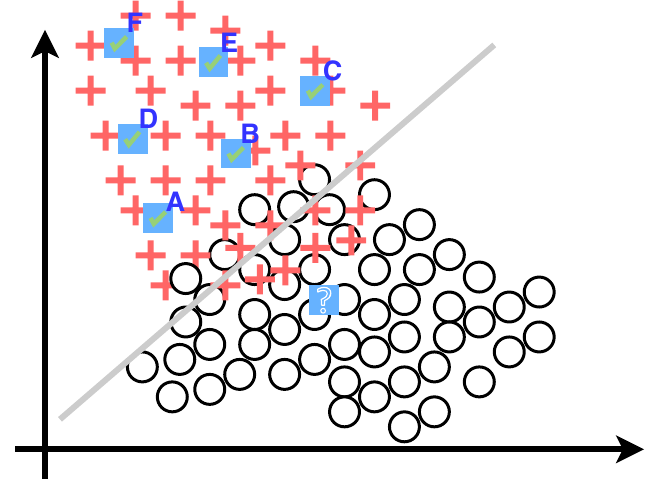}%
      \caption{%
          Generating a \textbf{diverse} (representative and minimal) collection of counterfactuals helps %
          to satisfy disparate needs of different explainees. %
          Explanation \textbf{availability} is not a problem in this example, especially that the depicted linear model is near-optimal and unlikely to change significantly. %
      }\label{fig:cf:group}%
  \end{subfigure}
  \hspace{0.01\textwidth}
  \begin{subfigure}[t]{0.310\textwidth}
      \centering
      \includegraphics[scale=0.36]{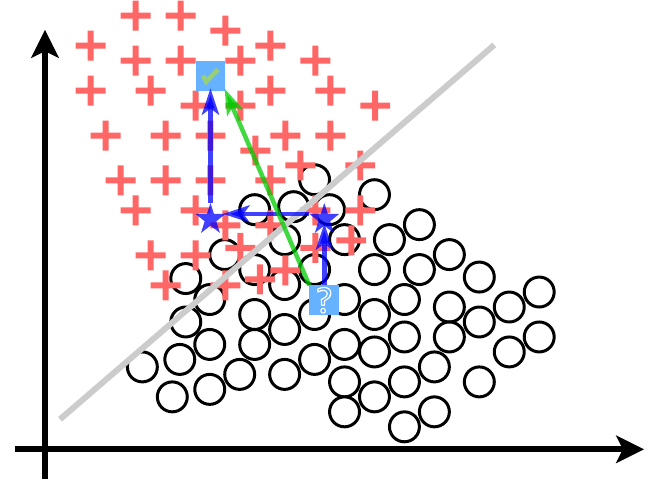}%
      \caption{%
          A counterfactual explanation -- captured by the green vector -- communicates the entire feature set change as a single action; %
          its path-based alternative -- depicted by the blue segments -- splits it into three simpler steps. %
          The latter approach becomes particularly useful in high-dimensional feature spaces, where it promotes a discrete sequence of sparse actions. %
      }\label{fig:cf:cl+rec}%
  \end{subfigure}
  \hspace{0.01\textwidth}
  \begin{subfigure}[t]{0.310\textwidth}
      \centering
      \includegraphics[scale=0.36]{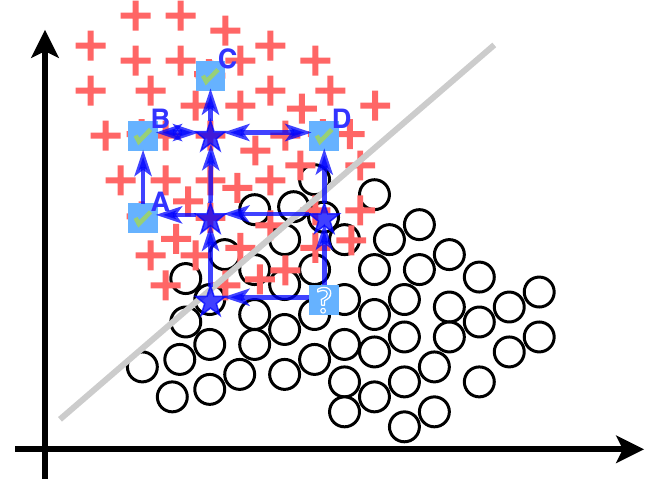}%
      \caption{%
          The \textbf{feasibility} of the shown path-based counterfactuals (which also entails their \emph{plausibility}, \emph{connectedness} and \emph{stability}) is ensured by constructing them upon epistemically-certain (or pre-existing) instances. %
          Accounting for the spatial properties additionally allows to capture the \textbf{affinity}, \textbf{branching} and \textbf{divergence} of these paths. %
      }\label{fig:cf:path}%
  \end{subfigure}
  \caption{%
  Visualisation of counterfactuals in relation to: %
  (\subref{fig:cf:group}) explanation groups and model selection; %
  (\subref{fig:cf:cl+rec}) path-based explainability; and %
  (\subref{fig:cf:path}) geometry of the feature space density. %
  The question mark indicates the explained instance, the star symbolises an intermediate step of a counterfactual path and the check mark represents a counterfactual data point. %
  }\label{fig:cf}%
\end{figure*}

Notably, %
the path-based explainability paradigm also provides the foundation for %
implementing desiderata such as %
\emph{affinity}, \emph{branching} and \emph{divergence}, %
which %
imbue counterfactual paths with spatial awareness %
by capturing the geometry of the feature space density~\citep{sokol2023navigating}. %
These properties account for the order in which feature changes need to be implemented, group explanations based on the (directional) similarity of their paths and differentiate incompatible counterfactuals; %
taken together, these desiderata %
additionally %
offer a richer perspective on explanation \emph{actionability}, \emph{diversity} and \emph{availability} (as well as a principled mechanism to realise these notions). %
In this context, %
uncertainty estimates %
could help to identify steps where paths are likely to branch (aleatoric) and robust decision boundary crossings (epistemic). %
A selection of the counterfactual properties discussed throughout this section is %
illustrated in \autoref{fig:cf}; %
we leave their in-depth uncertainty-oriented exploration for future work. %
Other popular XAI desiderata, which are not specific to this explanation type, %
can be found in resources such as the \emph{Explainability Fact Sheets}~\citep{sokol2020explainability}. %

Another aspect of counterfactual properties %
that warrants further scrutiny is how they are evaluated, %
or more precisely how they are measured for the final explanation. %
\autoref{apx:eval-metrics} reviews common counterfactual evaluation metrics; %
as discussed in \autoref{sec:methodology:challenges}, many of them rely on distance measured in the feature space, %
typically adopting its Manhattan and Euclidean or Gower definitions since formalising its meaningful notion for a specific modelling problem is inherently difficult. %
Consequently, %
any counterfactual explainer that uses a bespoke similarity function -- one tailored specifically to the underlying data -- is bound to score poorly %
on such benchmarks. %
As noted earlier in the overview of our experimental results (\autoref{sec:experiments}), %
we have encountered a similar challenge %
when evaluating the techniques introduced throughout this paper %
given their grounding in uncertainty, which, as we argued, is conceptually orthogonal to the notion of distance measured in the feature space (\autoref{sec:methodology:challenges}). %

These observations point to the fundamental limitations of many counterfactual evaluation metrics, %
in particular those built atop distance. %
This problem is further compounded by the widely documented unreliability of many qualitative approaches to XAI evaluation~\cite{sokol2024what}, %
demonstrating an urgent need for %
more flexible metrics that can be easily adapted to each unique modelling setting. %
As a step in this direction, %
\autoref{apx:dist-proxy} explores how uncertainty could act as a proxy of distance. %
In case of the \emph{similarity} desideratum, %
this perspective would facilitate %
comparing %
explanations to each other through their uncertainty in addition to the gap between them in the feature space as well as their separation from the explained instance, %
with the benefits of such a strategy outlined earlier in \autoref{sec:methodology:ante} (refer to \autoref{fig:example-p}). %
For path-based explanations, \emph{similarity} (as well as \emph{sparsity}) %
could %
further be extended to %
account for the quantity and size of individual steps as well as the number of features affected by %
each of these %
segments. %

\subsection{Implications and Challenges of Uncertainty-aware XAI}%
\label{sec:methodology:anti}%

Reliable uncertainty quantification is undoubtedly challenging, and in some cases even impossible~\citep{hullermeier2021aleatoric}. %
Using it as the foundation of XAI may %
thus curtail the progress of this field -- at least in the short term. %
Embracing this perspective will also likely %
require a fundamental shift %
in how explanations are formalised, in some cases increasing the technical complexity of XAI to the detriment of lay explainees~\citep{chau2023explaining}; %
for example, such algorithmic insights might become probabilistic altogether %
as can be seen in \autoref{fig:model:linear} for any counterfactual placed in the span of possible classifiers (top-right or bottom-left). %
With this in mind, %
for many (low stakes) domains, crisp (\emph{ante hoc} interpretable) modelling may suffice, %
making uncertainty quantification redundant. %
However, simply considering this aspect of data-driven systems can catalyse important insights even if %
(reliable) uncertainty estimates are ultimately deemed unnecessary. %

While black-box AI systems %
appear to dominate current practice, %
the techniques presented throughout this paper %
could %
greatly benefit from %
access to and meaningful control over model internals. %
Our framework is thus %
highly %
relevant %
to %
inherently interpretable (\emph{ante hoc}) models or scenarios where full model transparency is a prerequisite %
as argued earlier in \autoref{sec:methodology:ante}. %
When this is not the case, %
uncertainty proxies and \emph{post hoc} explainability are largely unavoidable; %
such methods are also valuable when %
dealing with legacy systems or where explainability must be retrofitted. %
As demonstrated by our extensive experiments (\autoref{sec:experiments}), the techniques proposed in this paper remain highly competitive in these settings despite their inherent simplicity. %

The uncertainty-centred framework presented here %
explicitly targets counterfactuals. %
Nonetheless, the broader premise of grounding XAI methods in foundational AI concepts, like uncertainty, is valid for other explanation types. %
Feature importance and attribution techniques as well as rule- and exemplar-based explanations could all benefit from uncertainty awareness. %
For example, %
feature importance measures may be grounded in their robustness determined by epistemic uncertainty; %
selection of prototypes and criticisms can rely on identifying regions of low and high uncertainty; and %
rule-based explanations may use uncertainty-guided pruning to discard brittle or ambiguous statements. %

Uncertainty-aware explanations can also serve as a mechanism %
for probing model multiplicity that arises due to the Rash\=omon effect, %
thus revealing the stability or fragility of predictions from across the hypothesis space~\citep{rudin2024amazing}. %
Aggregating or contrasting such explanatory insights %
can help to uncover structural patterns in model behaviour, support fairness audits and identify regions of epistemic uncertainty warranting closer inspection. %
More broadly, uncertainty awareness can aid with identifying issues pertaining to data shift, coverage, quality and bias, %
with the corresponding explanations becoming a convenient instrument to discover and rectify them. %
Overall, enhancing data curation in this way could lead to lower epistemic uncertainty and further improve the robustness, interpretability and equitability of AI systems. %

One key concern, mentioned earlier, with regard to uncertainty-aware explanations is their accessibility. %
While such insights are more comprehensive, %
they impose additional cognitive burden on end users, especially those lacking technical expertise. %
Further research into human-centred communication of uncertainty and its derivative concepts is thus vital. %
Approaches that adjust the information being presented based on user expertise, current cognitive load and task importance can bridge the gap between the level of detail contained in an explanation and its practical usability. %
Dedicated user interfaces could, for example, offer layered explanatory insights, only revealing complex uncertainty concepts upon request, %
or adapt methods from risk communication to make probabilistic pieces of information more intuitive. %

Further inspiration can come from the growing field of interactive and iterative XAI. %
Rather than providing users with a single explanation or a collection thereof, such insights can be delivered in a dialogue, enabling explainees to request and review alternative scenarios, refine constraints or explore hypothetical futures~\citep{sokol2023navigating}. %
In such settings, access to reliable uncertainty estimates could allow users to %
better judge which trade-offs are acceptable, which actions are feasible and which risks are tolerable. %
The interactive delivery mechanism also promises to facilitate better trust calibration %
and to empower various stakeholders to co-create their explanations in a way that addresses their unique needs and helps them to achieve their desired goals. %

\section{Conclusion and Future Work}\label{sec:conclusion}%

In this paper we demonstrated that %
uncertainty quantification offers a sound \emph{unifying framework} for \emph{counterfactual explainability} %
-- an often overlooked connection that %
facilitates rigorous generation of trustworthy explanatory insights for different model classes. %
Consequently, we observed that %
counterfactual explainers tend to be overcomplicated since they attempt to \emph{compensate for various shortcomings} of AI models. %
Grounding these tools in uncertainty allows them to remain relatively simple, %
instead shifting the engineering burden towards \emph{building more principled predictive models} %
that are reliable, robust and, most importantly, uncertainty-aware. %

Specifically, we identified the \emph{ante hoc} interpretability paradigm as a promising candidate for achieving this goal %
given that, as we showed, its tenets are fundamentally intertwined with the basic principles of uncertainty quantification. %
In brief, %
recognising this synergistic relation benefits the comprehensibility of this type of predictive models as well as improves the resulting aleatoric and epistemic uncertainty estimates; %
it also %
opens %
a pathway for \emph{bringing human-centred explanatory insights}, like counterfactuals, \emph{to ante hoc interpretable models}, whose intelligibility %
has thus far been largely limited to technical experts. %
More broadly, %
this paper %
charted a new direction for XAI research that strives to strengthen its foundations through integration of core AI concepts. %
Our uncertainty-aware counterfactuals are a case in point, exemplifying the benefits of such a synthesis; %
they offer technically grounded explanatory insights that are also inherently human-centred. %
As the field matures, we anticipate that this integration will become not only advantageous but necessary for developing AI systems that are truly transparent, robust and capable of addressing real-life challenges. %

Given their prevalence, %
this paper focused on counterfactuals, %
nonetheless in future work we will expand our uncertainty-centred perspective to other explanation types. %
Since our approach is premised on access to reliable aleatoric and epistemic uncertainty estimates, %
we will also investigate %
critical \emph{modelling assumptions} that impact their quality as well as %
latest \emph{uncertainty quantification} and \emph{probability calibration} methods, %
beginning with \emph{ante hoc} interpretable models because of their strong connection to this topic. %
We additionally plan to examine the role that other \emph{types} and \emph{sources} of uncertainty play in AI transparency, %
which will guide our further exploration of uncertainty-aware XAI. %
In particular, we will look into second-order uncertainty %
and suitable (probabilistic) formalisations of explanations %
as well as %
feature measurement uncertainty and %
variability in the human implementation of counterfactual actions. %
Such a broad %
perspective can, among others, improve the \emph{robustness} and \emph{stability} of predictions and explanations, %
benefiting (high stakes) domains like healthcare, %
where decisions are often made based on incomplete and uncertain information~\citep{sokol2025artificial}. %
More generally, we will study other fundamental AI concepts that are overlooked in XAI research, %
striving to strengthen the foundations of %
explainable artificial intelligence. %

\appendix

\section{Counterfactual Evaluation Metrics}\label{apx:eval-metrics}

For the evaluation of counterfactual explanations we adopt a range of metrics commonly employed in XAI research~\citep{guidotti2022counterfactual}. %
Specifically, we use their variant for evaluating a \emph{single} counterfactual instance (rather than their collection). %
These metrics and their definitions are as follows: %
\begin{description}[topsep=0pt,partopsep=0pt,noitemsep]

    \item [Validity $m_{\text{val}}(\cdot)$ $\downarrow$]%
    quantifies \emph{whether} (for crisp models) or \emph{to what extent} (for probabilistic models) a counterfactual instance $\xCF$ is classified with the desired counterfactual class $\cplus$. %
    For a probabilistic model $h$ invalidity is defined as: %
    $$
        m_{\text{val}}(\xCF; \; \cplus) = 1 - h(\cplus | \xCF) %
        \text{.}
    $$

    \item [Similarity $m_{\text{sim}}(\cdot)$ $\downarrow$]%
    measures how dissimilar a counterfactual instance $\xCF$ is from the source factual point $\xF$ by quantifying the distance $d(\xCF, \xF)$ between them, thus capturing how far away they are in the feature space: %
    $$
        m_{\text{sim}}(\xCF; \; \xF) = d(\xCF, \xF)%
        \text{.}
    $$

    \item[Sparsity $m_{\text{spar}}(\cdot,\cdot)$ $\downarrow$]%
    measures how many features were changed when transforming the factual $\xF$ to the counterfactual $\xCF$ (for $n$-dimensional feature space $\mathcal{X}$): %
    $$
        m_{\text{spar}}(\xF, \xCF) = \frac{1}{n} \sum_{i=1}^n \mathds{1}_{\xF_i \neq \xCF_i}%
        \text{.}
    $$

    \item [Plausibility $m_{\text{plau}}(\cdot)$ $\downarrow$]%
    measures how plausible a counterfactual instance $\xCF$ is %
    by quantifying its distance $d(\xCF, \cdot)$ from the data manifold given by a representative reference population $X \subseteq \mathcal{X}$: %
    $$
        m_{\text{plau}}(\xCF; \; X) = \min_{x \in X} d(\xCF, x)%
        \text{.}
    $$

    \item [Discriminativeness $m_{\text{dis}}(\cdot)$ $\uparrow$]%
    measures how well a counterfactual instance $\xCF$ retrieved for a factual point $\xF$ can be distinguished %
    from a set of nearby instances. %
        To this end, a crisp $1$-nearest neighbour classifier $g: \mathcal{X} \mapsto \mathcal{C}$ based on the chosen distance function $d(\cdot, \cdot)$ is trained on $X = \{\xCF, \xF\}$; %
    then, the value of the metric $m_{\text{dis}}(\cdot)$ is %
    given by the accuracy of $g$ evaluated on %
    a set of $n >0$ data points from each class $\cplus$ and $\cminus$ -- denoted respectively by $X^\cpos$ and $X^\cneg$, where $n = |X^\cpos| = |X^\cneg|$ -- that are closest to $\xF$ (these instances can, for example, be sourced from the training set): %
    \begin{multline*}
        m_{\text{dis}}(g; \; X^\cpos, X^\cneg) = %
            \frac{1}{|X^\cpos| + |X^\cneg|}
        \\
            \left(
            \sum_{x \in X^\cpos}
            \mathds{1}_{g(x) = \cplus}
            +
            \sum_{x \in X^\cneg}
            \mathds{1}_{g(x) = \cminus}
            \right)
        \text{.}
    \end{multline*}
    Note that explanations with high \emph{similarity} are likely to have good \emph{discriminativeness} as well -- according to this formalisation of the metric -- since the instances in $X^\cpos$ are located close to a decision boundary; %
    however, a counterfactual placed near a decision boundary is, by definition, not discriminative, demonstrating the deficiency of this particular metric formalisation. %

    \item [Stability $m_{\text{sta}}(\cdot, \cdot)$ $\uparrow$]%
    measures how unstable a counterfactual instance $\xCF$ generated for a factual point $\xF$ is. %
     Given an instance ${\xF}^\prime$ that is similar to the explained point $\xF$ and whose counterfactual is ${\xCF}^\prime$, %
     this metric quantifies how dissimilar the two counterfactuals $\xCF$ and ${\xCF}^\prime$ are according to the chosen distance function $d(\cdot, \cdot)$: %
     $$
         m_{\text{sta}}(\xCF, {\xCF}^\prime; \; \xF, {\xF}^\prime) = %
         \frac{1}{1 + d(\xF, {\xF}^\prime)} d(\xCF, {\xCF}^\prime)
        \text{.}
     $$

\end{description}

All of these metrics except \emph{validity} and \emph{sparsity} %
are constructed upon a distance function $d: \mathcal{X} \times \mathcal{X} \mapsto \mathds{R}^+$ defined on the feature space $\mathcal{X}$. %
But, as noted earlier in \autoref{sec:methodology:cf}, formalising distance in a meaningful way is usually non-trivial (and application-specific). %
For this reason, the implementation of these (and other) counterfactual evaluation metrics %
-- as well as (the corresponding) counterfactual properties used for generating these explanations -- %
usually falls back on generic measures such as the Euclidean of Gower's distance~\citep{guidotti2022counterfactual}. %
Given this paper's focus on making explainability uncertainty-aware, one interesting direction for future research is revising XAI evaluation metrics and properties such that they rely less on distance and more on uncertainty estimates -- see \autoref{apx:dist-proxy} for further exploration of this topic. %

In our experiments reported in \autoref{sec:experiments}, we adopt a distance function commonly used in counterfactual research defined as~\citep{wachter2017counterfactual}: %
$$
d(x, x^\prime) = \frac{1}{2} \left( \frac{1}{|C|} \sum_{i\in C} \frac{|x_i - x_i^\prime|}{\MAD_i} + \frac{1}{|D|} \sum_{i \in D} \mathds{1}_{x_i \neq x_i^\prime}
\right)
\text{,}
$$ 
where $C$ and $D$ respectively capture the indices of \emph{continuous} and \emph{categorical} (discrete) features in $\mathcal{X}$, and %
$\MAD_i$ stands for the \emph{median absolute deviation} of the $i$\textsuperscript{th} feature computed based on the training data set. %
Notably, this choice %
inadvertently links \emph{dissimilarity} with \emph{discriminativeness} %
since highly dissimilar counterfactuals are bound to have low discriminativeness given that the instances in $X^\cneg$ are by definition chosen to lie as close as possible to $\xF$. %

\section{Uncertainty-based Proxy of Distance}\label{apx:dist-proxy}%

The implementation of many %
counterfactual properties (including those reviewed in \autoref{sec:related:exp}) and evaluation metrics (like the ones listed in \autoref{apx:eval-metrics}) %
relies on a distance function $d$, often defined on the input feature space $\mathcal{X}$, i.e., $d: \mathcal{X} \times \mathcal{X} \mapsto \mathds{R}^+$. %
It is the principal building block of \emph{similarity} and \emph{sparsity}, %
but other %
properties such as \emph{actionability} are frequently expressed in terms of distance too %
-- e.g., modifying the underlying measure to assign infinity to counterfactuals built upon unactionable features -- %
to facilitate their direct optimisation. %
While our uncertainty-based framework proposed in \autoref{sec:methodology:cf} demonstrates that %
most counterfactual properties can be unified under its umbrella, %
capturing the notion of distance is fundamentally beyond its scope. %
Given that the goal of our approach is to ensure reliability and robustness of counterfactual explanations, %
it is important not to neglect the impact of the chosen %
distance metric %
on their generation and evaluation, %
especially that %
it may be misspecified vis-\`a-vis the measure of similarity used internally by the explained model. %

But, as discussed earlier in \autoref{sec:methodology:challenges}, formalising distance that is meaningful for a particular application domain is usually non-trivial -- a challenge that is also widely recognised in AI fairness work~\citep{dwork2012fairness}. %
Because of this, XAI research frequently resorts to %
generic distance metrics~\citep{guidotti2022counterfactual} despite their known shortcomings, some of which we discussed earlier in \autoref{sec:methodology:challenges} in the context of counterfactual generation and evaluation. %
More broadly, reliance on \emph{any} distance measure may be unsound, hence detrimental, %
whenever counterfactual desiderata are misaligned with %
the fundamental properties of distance functions. %

For example, consider the \emph{symmetry} of distance, i.e., $d(x, x^\prime) = d(x^\prime, x)$; %
features that can only change unidirectionally, like age, are incompatible with this assumption, %
necessitating \emph{ad hoc} fixes like the aforementioned ``infinite distance'' stopgap. %
Therefore, a \emph{pseudo-distance} $\mCF : \mathcal{X} \times \mathcal{X} \mapsto \mathds{R}$ that behaves like a \emph{similarity} measure and only adheres to $\mCF(x, x) > \mCF(x^\prime, x) \; \forall x^\prime \neq x \in \mathcal{X}$ %
as well as (with a slight abuse of notation) %
$\mCF(x + \epsilon, x) > \mCF(x + 2\epsilon, x) \; \forall \epsilon \neq 0$ %
appears flexible enough and sufficient for counterfactual generation purposes. %
Note that for any proper distance metric $d$, $\mCF(x^\prime, x) = -d(x^\prime, x)$ fulfils these criteria %
given that the primary requirement of $\mCF$ is to decrease as both more and bigger changes are applied to the factual data point when it gets gradually transformed into the counterfactual instance. %

But designing the $\mCF$ metric is clearly no easier than finding a suitable distance candidate $d$ %
when both are defined directly on the feature space $\mathcal{X}$; %
therefore, identifying %
other proxies for similarity that are meaningful, robust and reliable %
is an important open research question. %
Since this paper is concerned with uncertainty-aware XAI, %
one promising idea is to express $\mCF$ in probabilistic terms. %
Specifically, we could use a \emph{location-scaled probability distribution} $p(x^\prime | x)$ to capture the likelihood of sampling $x^\prime$ given $x$, with $\mCF( x^\prime, x) = p(x^\prime| x)$. %
Instead of quantifying the similarity of $x$ and $x^\prime$ through a direct measurement in the feature space, %
this notion of $\mCF$ conveys how likely observing $x^\prime$ is when being located at $x$. %
Consequently, %
while distance-based formalisations of $\mCF$ communicate that a particular change to the feature vector is more or less ``difficult'' (or ``easy''), %
the probabilistic definition of $\mCF$ reframes this as the attribute adjustment being more or less ``unlikely'' (or ``likely''). %

In practice, continuous features can be modelled with the multivariate Gaussian distribution. %
Despite its inherent reliance on the Euclidean distance, %
it directly accounts %
for attribute interactions by encoding this information in the shape of the underlying probability distribution (given by the covariance matrix). %
Since probability distributions can capture arbitrary structures, this approach can also naturally %
and faithfully model dependencies between categorical features, %
for example, reflecting that an attribute value change from $\mathtt{A}$ to $\mathtt{B}$ is different than its adjustment from $\mathtt{B}$ to $\mathtt{C}$. %
This formalisation of $\mCF$ thus improves %
upon the 0--1 loss (with each feature treated independently and weighted equally) commonly used in XAI when dealing with this type of variables~\citep{guidotti2022counterfactual}. %
Combining the two gives us the overall distribution that becomes our similarity measure: %
$$%
\mCF(x^\prime, x) = p(x^\prime | x) = p_C(x^\prime | x) p_D (x^\prime | x)%
\text{,}
$$%
where $p_C$ and $p_D$ respectively refer to the (location-scaled) probability distributions of \emph{continuous} and \emph{categorical} (discrete) attributes. %

{
\begin{figure*}[t]
    \renewcommand\thesubfigure{Step~\Roman{subfigure}}%
    \captionsetup[subfigure]{labelformat=parens, labelsep=space}%
    \centering
    \begin{subfigure}[t]{0.31\textwidth}
        \includegraphics[scale=0.36]{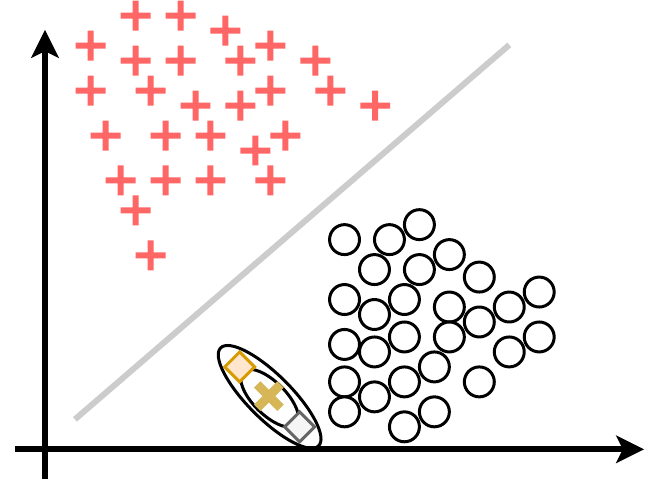}%
    \caption{The yellow cross is the factual instance $\xF$, thus the origin of sampling in the first step of our counterfactual generation process; the orange square is the best sample at this stage.}%
    \label{fig:prob_cf_a}%
    \end{subfigure}
    \hspace{0.01\textwidth}%
    \begin{subfigure}[t]{0.31\textwidth}
        \includegraphics[scale=0.36]{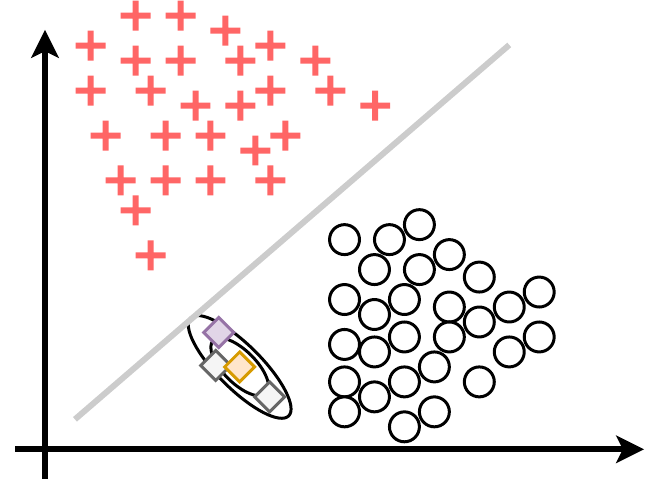}%
    \caption{The orange square -- corresponding to the instance with the same marker in \subref{fig:prob_cf_a} -- is the origin of sampling in the second step; the violet square is the best sample at this stage.}%
    \label{fig:prob_cf_b}%
    \end{subfigure}
    \hspace{0.01\textwidth}%
    \begin{subfigure}[t]{0.31\textwidth}
        \includegraphics[scale=0.36]{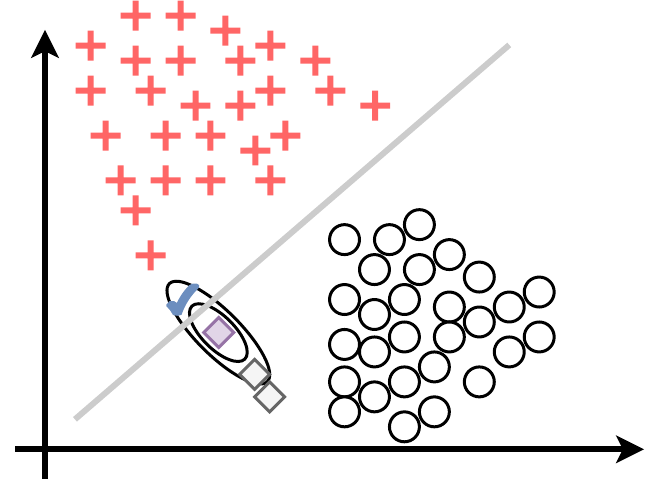}%
    \caption{The violet square %
    is the origin of sampling in the third step; the blue check mark is the best sample and our counterfactual $\xCF$ given that it received the desired prediction.} %
    \label{fig:prob_cf_c}%
    \end{subfigure}
    \caption{%
    Illustration of a counterfactual generation process that foregoes distance measured directly in the feature space as a measure of instance similarity, instead relying on the Markov chain Monte Carlo approach -- here, with three steps -- introduced in \autoref{apx:dist-proxy}. %
    In \subref{fig:prob_cf_a}, two candidates -- represented by squares -- are sampled from the $p(x | \xF)$ distribution centred on the factual point $\xF$ depicted by the yellow cross; %
    the orange square, which is the best candidate at this stage, is chosen as the next sampling location. %
    In \subref{fig:prob_cf_b}, three samples are drawn from the new location-scaled probability distribution centred on the orange square (corresponding to the data point with the same marker in \subref{fig:prob_cf_a}); %
    the instance represented by the violet square becomes the sampling anchor for the next step. %
    Finally, in \subref{fig:prob_cf_c} the process terminates as it delivers a counterfactual explanation given by the blue check mark, which is one of the points sampled at this stage. %
    }\label{fig:prob_cfs_generation}
\end{figure*}
}

To illustrate this approach, let us consider (without loss of generality) a toy example with two continuous features %
-- see \autoref{fig:prob_cfs_generation}. %
For each data point $x$, we model $\mCF(\cdot, x) = p(\cdot | x) = \mathcal{N}(x, \Sigma)$, %
where %
the individual components $\sigma_{ij}$ of the covariance matrix $\Sigma$ implicitly encode \emph{sparsity} and \emph{actionability}. %
Crucially, these parameters %
have an intuitive interpretation, %
where a non-zero value of $\sigma_{ij}$, i.e., $\sigma_{ij} \not = 0$, indicates %
a monotonic relation (hence interaction) between features $i$ and $j$, %
with its direction determined by the sign of $\sigma_{ij}$. %
Similarly, %
an attribute $i$ whose value cannot be changed will be encoded by %
$\sigma_{ij} \approxeq 0$ for every other feature $j \neq i$. %
A clear advantage of this direct link between the elements of $\Sigma$ and feature relations is the user's ability to manually adjust each $\sigma_{ij}$ to enforce the desired requirements. %

Sampling from $\mathcal{N}(x, \Sigma)$ thus delivers (with high probability) instances that represent %
\emph{feasible} and \emph{actionable} variations of $x$; %
however, %
it is unlikely that drawing $x^\prime$ from the location-scaled distribution $\mathcal{N}(\xF, \Sigma)$ centred on the factual point $\xF$, i.e., $x^\prime \sim \mathcal{N}(\xF, \Sigma)$, %
will outright produce a valid counterfactual $\xCF$. %
We therefore employ a (greedy) Markov chain Monte Carlo approach -- illustrated in \autoref{fig:prob_cfs_generation} -- starting at the factual instance $\xF$ and terminating when a valid counterfactual $\xCF$ is found. %

Specifically, at each step $t \in [1,\ldots, t_{\text{last}})$ %
of the process %
we sample $x^{(t+1)}$ from $p(x^{(t+1)} | x^{(t)}) = \mathcal{N}(x^{(t)}, \Sigma)$, with $x^{(1)} = \xF$ and $x^{(t_{\text{last}})} = \xCF$, and where %
each new point $x^{(t+1)}$ only depends on the preceding instance $x^{(t)}$, guaranteeing the \emph{Markov property}. %
Given the symmetry of the location-scaled distribution, $x^{(t+1)}$ might inadvertently be ``farther'' from the counterfactual class $\cplus$ in terms of its predicted probability $h(\cplus | x^{(t+1)})$ as compared to that of the current point $x^{(t)}$. %
We therefore sample $w$ %
instances $x_1^{(t+1)},\ldots,x_w^{(t+1)} \sim \mathcal{N}(x^{(t)}, \Sigma)$ at each step $t$ and %
select the one with the highest probability of the counterfactual class, i.e., $x^{(t+1)} = \argmax_{x \in \{x_1^{(t+1)},\ldots,x_w^{(t+1)}\}} h(\cplus | x)$, %
making our approach \emph{greedy}. %

\section{Full Experimental Results\label[appendix]{apx:tables}}%

\autoref{sec:experiments} presents an overview of our experimental results for the \emph{DARE} ensemble, with \autoref{apx:exp-full} providing the corresponding findings for the \emph{deep} and \emph{adversarial} ensembles. %
Here, we complement this analysis by offering %
detailed evaluation outcomes %
for %
our uncertainty-based definitions of counterfactual desiderata (proposed in \autoref{sec:methodology:cf}) %
as well as %
the two variants of our uncertainty-aware explainer (introduced in \autoref{sec:methodology:gen}) and the selected baseline methods. %
Respectively, these can be found in %
\cref{tab:dare_ensemble_results_properties,tab:dare_ensemble_results_explainers} for the \emph{DARE} ensemble, %
\cref{tab:deep_ensemble_results_properties,tab:deep_ensemble_results_explainers} for the \emph{deep} ensemble and %
\cref{tab:adversarial_ensemble_results_properties,tab:adversarial_ensemble_results_explainers} for the \emph{adversarial} ensemble. %
The \textbf{bold}, \underline{underlined} and \textit{italicised} entries in these tables indicate the \textbf{best}, \underline{second-best} and \textit{third-best} metric scores, in that order, across approaches -- i.e., counterfactual property definitions in \cref{tab:dare_ensemble_results_properties,tab:adversarial_ensemble_results_properties,tab:deep_ensemble_results_properties} and explainers in \cref{tab:dare_ensemble_results_explainers,tab:adversarial_ensemble_results_explainers,tab:deep_ensemble_results_explainers} -- for each individual data set. %
Notably, %
the distance-aware variant of our uncertainty-based explainer (see \emph{uncertainty+d} in \cref{tab:dare_ensemble_results_explainers,tab:adversarial_ensemble_results_explainers,tab:deep_ensemble_results_explainers}) often achieves the \textbf{best} or \underline{second-best} result, especially on \emph{similarity}, \emph{plausibility} and \emph{stability}; %
overall, it demonstrates highly competitive performance despite its fundamentally simple design. %

\afterpage{

\clearpage

\begin{table*}[p]%
\centering
    \scriptsize%
    \setlength{\tabcolsep}{2.55pt}%
\caption{%
      Detailed evaluation results %
      based on the \emph{DARE} ensemble %
      of our %
      uncertainty-based counterfactual desiderata, as defined in \autoref{sec:methodology:cf}, %
      for the six metrics listed in \autoref{apx:eval-metrics}. %
}
\label{tab:dare_ensemble_results_properties}
\begin{tabular}{@{}l l r r r r r r@{}}
\toprule
\multirow{2.5}{*}{Data set} & \multirow{2.5}{*}{Property} & \multicolumn{6}{c}{Metric} \\
\cmidrule(lr){3-8}
 &  & validity $\downarrow$ & similarity $\downarrow$ & sparsity $\downarrow$ & plausibility $\downarrow$ & discriminativeness $\uparrow$ & stability $\uparrow$ \\ 
\midrule
\multirow{7}{*}{bank} & validity & 0.003 & \underline{23.191} & \textbf{0.900} & \underline{13.788} & 0.479 & \textbf{3.712} \\ 
 & connectedness & \textit{0.003} & \textbf{22.282} & \textbf{0.900} & \textbf{13.550} & 0.473 & \underline{4.876} \\ 
 & robustness & 0.038 & 220.877 & \textit{0.900} & 128.137 & \textit{0.576} & 72.891 \\ 
 & feasibility & 0.054 & 206.141 & \underline{0.900} & 113.984 & 0.562 & 54.128 \\ 
 & discriminativeness & \textbf{0.002} & \textit{44.323} & \textbf{0.900} & \textit{28.428} & \underline{0.584} & 10.368 \\ 
 & plausibility & \underline{0.002} & 45.369 & \textbf{0.900} & 31.610 & 0.496 & \textit{9.394} \\ 
 & similarity & 0.286 & 397.598 & \underline{0.900} & 370.509 & \textbf{0.591} & 131.601 \\ 
\midrule
\multirow{7}{*}{cancer} & validity & 0.038 & \textit{834.910} & \textbf{1.000} & \textit{833.164} & 0.791 & \underline{95.386} \\ 
 & connectedness & 0.042 & \underline{826.251} & \textbf{1.000} & \underline{824.358} & 0.790 & \textit{104.612} \\ 
 & robustness & \textit{0.033} & \textbf{753.015} & \textbf{1.000} & \textbf{751.358} & \textit{0.792} & 146.110 \\ 
 & feasibility & \textbf{0.024} & 868.934 & \textbf{1.000} & 867.017 & 0.790 & 330.676 \\ 
 & discriminativeness & \underline{0.029} & 916.483 & \textbf{1.000} & 914.749 & \underline{0.792} & \textbf{93.026} \\ 
 & plausibility & 0.033 & 865.110 & \textbf{1.000} & 863.310 & 0.792 & 131.771 \\ 
 & similarity & 0.116 & 28865.025 & \textbf{1.000} & 28863.963 & \textbf{0.793} & 9316.862 \\ 
\midrule
\multirow{7}{*}{churn} & validity & 0.012 & \underline{14.102} & \textbf{0.265} & \underline{12.152} & 0.760 & \underline{2.339} \\ 
 & connectedness & 0.017 & \textbf{13.506} & \textbf{0.265} & \textbf{11.629} & 0.743 & \textit{2.491} \\ 
 & robustness & \underline{0.005} & 17.229 & \textbf{0.265} & 15.299 & 0.820 & \textbf{2.216} \\ 
 & feasibility & \textit{0.006} & 121.042 & \textbf{0.265} & 118.685 & \textbf{0.884} & 34.435 \\ 
 & discriminativeness & \textbf{0.003} & 19.008 & \textbf{0.265} & 17.068 & \textit{0.839} & 2.783 \\ 
 & plausibility & 0.009 & \textit{15.246} & \textbf{0.265} & \textit{13.331} & 0.781 & 3.175 \\ 
 & similarity & 0.138 & 89.644 & \textbf{0.265} & 87.231 & \underline{0.884} & 33.424 \\ 
\midrule
\multirow{7}{*}{COMPAS} & validity & 0.016 & \textbf{19.489} & \textbf{0.570} & \textbf{7.920} & \underline{0.723} & \underline{3.964} \\ 
 & connectedness & 0.015 & \underline{19.849} & \textbf{0.570} & \underline{8.311} & \textit{0.721} & \textit{4.146} \\ 
 & robustness & \underline{0.011} & \textit{20.501} & \textbf{0.570} & 9.164 & 0.721 & \textbf{3.853} \\ 
 & feasibility & \textbf{0.008} & 21.748 & \textbf{0.570} & 10.515 & 0.716 & 4.377 \\ 
 & discriminativeness & \textit{0.012} & 20.521 & \textbf{0.570} & \textit{8.919} & 0.720 & 4.240 \\ 
 & plausibility & 0.155 & 22.314 & \textbf{0.570} & 9.963 & \textbf{0.751} & 6.807 \\ 
 & similarity & 0.113 & 82.356 & \underline{0.570} & 56.180 & 0.686 & 32.184 \\ 
\midrule
\multirow{7}{*}{diabetes} & validity & 0.030 & \textit{4.599} & \underline{0.985} & \textit{2.642} & 0.756 & \textit{1.573} \\ 
 & connectedness & 0.030 & \underline{4.457} & \underline{0.985} & \underline{2.532} & 0.755 & 1.654 \\ 
 & robustness & \textit{0.024} & 7.562 & \underline{0.985} & 5.681 & \underline{0.763} & \underline{1.549} \\ 
 & feasibility & \underline{0.016} & 15.394 & \underline{0.985} & 13.434 & \textbf{0.774} & 4.512 \\ 
 & discriminativeness & \textbf{0.008} & 6.313 & \underline{0.985} & 4.279 & \textit{0.762} & 2.145 \\ 
 & plausibility & 0.358 & \textbf{3.162} & \textit{0.987} & \textbf{1.307} & 0.690 & \textbf{1.182} \\ 
 & similarity & 0.062 & 6.258 & \textbf{0.984} & 4.405 & 0.755 & 2.862 \\ 
\midrule
\multirow{7}{*}{FICO} & validity & 0.019 & \textbf{5.311} & \underline{0.998} & \textbf{3.476} & 0.715 & \underline{1.661} \\ 
 & connectedness & \textit{0.017} & \underline{5.506} & \underline{0.998} & \underline{3.690} & 0.715 & \textit{1.766} \\ 
 & robustness & 0.018 & \textit{5.657} & \underline{0.998} & \textit{3.772} & 0.716 & \textbf{1.547} \\ 
 & feasibility & \textbf{0.006} & 17.380 & \underline{0.998} & 15.203 & \textbf{0.727} & 13.012 \\ 
 & discriminativeness & \underline{0.014} & 5.694 & \underline{0.998} & 3.799 & 0.715 & 1.779 \\ 
 & plausibility & 0.046 & 8.783 & \underline{0.998} & 6.843 & \textit{0.722} & 3.684 \\ 
 & similarity & 0.123 & 115.552 & \textbf{0.998} & 112.945 & \underline{0.724} & 40.911 \\ 
\midrule
\multirow{7}{*}{home} & validity & 0.031 & \underline{103.274} & \textit{0.993} & \underline{101.922} & \underline{0.828} & \underline{19.706} \\ 
 & connectedness & 0.027 & \textit{106.143} & 0.993 & \textit{104.777} & 0.821 & 22.039 \\ 
 & robustness & \underline{0.020} & 181.278 & \textit{0.993} & 179.955 & \textit{0.826} & 25.971 \\ 
 & feasibility & \textbf{0.009} & 196.558 & \underline{0.993} & 195.266 & 0.823 & 27.253 \\ 
 & discriminativeness & \textit{0.022} & 186.881 & \textit{0.993} & 185.564 & \textit{0.826} & \textbf{18.663} \\ 
 & plausibility & 0.049 & 155.113 & \textit{0.993} & 153.759 & \underline{0.828} & \textit{21.516} \\ 
 & similarity & 0.277 & \textbf{60.997} & \textbf{0.990} & \textbf{59.687} & \textbf{0.854} & 29.272 \\ 
\midrule
\multirow{7}{*}{housing} & validity & 0.030 & 33.051 & \textbf{1.000} & 28.536 & \textbf{0.799} & \textit{2.948} \\ 
 & connectedness & \textit{0.028} & \textit{32.354} & \textbf{1.000} & \textit{27.829} & \underline{0.799} & 3.139 \\ 
 & robustness & \underline{0.025} & 35.919 & \textbf{1.000} & 31.675 & \textit{0.799} & \textbf{2.467} \\ 
 & feasibility & 0.040 & \textbf{30.358} & \textbf{1.000} & \textbf{25.986} & 0.798 & 3.303 \\ 
 & discriminativeness & \textbf{0.023} & 36.527 & \textbf{1.000} & 31.900 & \textbf{0.799} & \underline{2.928} \\ 
 & plausibility & 0.036 & \underline{31.247} & \textbf{1.000} & \underline{27.177} & 0.797 & 4.433 \\ 
 & similarity & 0.118 & 534.760 & \textbf{1.000} & 511.605 & 0.763 & 99.423 \\ 
\midrule
\multirow{7}{*}{titanic} & validity & \textit{0.001} & 66.537 & \textbf{0.615} & 63.873 & 0.757 & \textit{17.862} \\ 
 & connectedness & 0.060 & \underline{32.609} & \textit{0.616} & \underline{29.607} & 0.754 & \underline{6.387} \\ 
 & robustness & \underline{0.000} & 344.883 & \textbf{0.615} & 342.470 & \underline{0.759} & 108.794 \\ 
 & feasibility & 0.007 & \textit{63.088} & \textit{0.616} & \textit{60.522} & 0.756 & 30.139 \\ 
 & discriminativeness & \textbf{0.000} & 205.373 & \textbf{0.615} & 201.544 & \textit{0.759} & 64.846 \\ 
 & plausibility & 0.442 & \textbf{7.052} & \underline{0.615} & \textbf{5.025} & 0.711 & \textbf{3.181} \\ 
 & similarity & 0.028 & 135.702 & \textbf{0.615} & 128.490 & \textbf{0.761} & 63.714 \\ 
\bottomrule
\end{tabular}
\end{table*}

\clearpage

\begin{table*}[p]%
\centering
    \scriptsize%
    \setlength{\tabcolsep}{2.55pt}%
\caption{%
      Detailed evaluation results %
      based on the \emph{DARE} ensemble %
      of counterfactual explainers, with \emph{uncertainty} and \emph{uncertainty plus distance} (uncertainty+d) as defined in \autoref{sec:methodology:gen}, %
      for the six metrics listed in \autoref{apx:eval-metrics}. %
}
\label{tab:dare_ensemble_results_explainers}
\begin{tabular}{@{}l l r r r r r r@{}}
\toprule
\multirow{2.5}{*}{Data set} & \multirow{2.5}{*}{Explainer} & \multicolumn{6}{c}{Metric} \\
\cmidrule(lr){3-8}
 &  & validity $\downarrow$ & similarity $\downarrow$ & sparsity $\downarrow$ & plausibility $\downarrow$ & discriminativeness $\uparrow$ & stability $\uparrow$ \\ 
\midrule
\multirow{6}{*}{bank} & GS & 0.474 & 257.296 & 0.998 & \textit{1.642} & \textit{0.566} & \textbf{0.830} \\ 
 & DICE & \underline{0.176} & 3083.723 & \textbf{0.729} & 652.803 & \underline{0.625} & nan \\ 
 & FACE & 0.380 & \textit{202.883} & \underline{0.794} & \textbf{0.000} & 0.547 & 62.889 \\ 
 & CLUE & nan & nan & nan & nan & nan & nan \\ 
 \cmidrule{2-8}
 & uncertainty & \textbf{0.049} & \underline{169.764} & 0.900 & 104.727 & \textbf{0.635} & \textit{56.904} \\ 
 & uncertainty+d & \textit{0.177} & \textbf{2.118} & \textit{0.897} & \underline{1.223} & 0.558 & \underline{1.057} \\ 
\midrule
\multirow{6}{*}{cancer} & GS & 0.484 & \textit{287.329} & \underline{1.000} & \textit{286.025} & 0.652 & nan \\ 
 & DICE & \textit{0.046} & \underline{2.431} & \textbf{0.999} & \underline{0.771} & 0.791 & \textbf{0.651} \\ 
 & FACE & 0.299 & \textbf{2.208} & \underline{1.000} & \textbf{0.000} & \textbf{0.813} & \underline{0.792} \\ 
 & CLUE & nan & nan & nan & nan & nan & nan \\ 
 \cmidrule{2-8}
 & uncertainty & \underline{0.032} & 782.271 & \underline{1.000} & 780.417 & \textit{0.792} & \textit{146.421} \\ 
 & uncertainty+d & \textbf{0.014} & 754.978 & \underline{1.000} & 753.606 & \underline{0.792} & 175.207 \\ 
\midrule
\multirow{6}{*}{churn} & GS & 0.479 & \textit{6.509} & 0.275 & \textit{4.472} & 0.620 & nan \\ 
 & DICE & \textit{0.385} & \underline{3.076} & \textit{0.266} & \underline{0.572} & \underline{0.784} & \textbf{0.532} \\ 
 & FACE & nan & nan & nan & nan & nan & nan \\ 
 & CLUE & nan & nan & nan & nan & nan & nan \\ 
 \cmidrule{2-8}
 & uncertainty & \textbf{0.004} & 17.845 & \underline{0.265} & 15.964 & \textbf{0.833} & \textit{2.562} \\ 
 & uncertainty+d & \underline{0.307} & \textbf{2.339} & \textbf{0.239} & \textbf{0.388} & \textit{0.655} & \underline{0.775} \\ 
\midrule
\multirow{6}{*}{COMPAS} & GS & 0.598 & 24.771 & 0.578 & \textit{0.677} & \underline{0.808} & nan \\ 
 & DICE & \textit{0.388} & 29.084 & \underline{0.462} & 2.432 & 0.752 & 7.894 \\ 
 & FACE & 0.427 & \textbf{10.321} & \textbf{0.419} & \textbf{0.000} & \textit{0.786} & \textit{4.350} \\ 
 & CLUE & \underline{0.341} & 22.244 & 1.000 & 10.229 & 0.680 & 5.040 \\ 
 \cmidrule{2-8}
 & uncertainty & \textbf{0.008} & \textit{20.715} & \textit{0.570} & 9.395 & 0.720 & \underline{3.733} \\ 
 & uncertainty+d & 0.445 & \underline{12.519} & 0.576 & \underline{0.556} & \textbf{0.823} & \textbf{2.340} \\ 
\midrule
\multirow{6}{*}{diabetes} & GS & 0.498 & \textit{2.243} & \textit{0.973} & \textit{0.337} & 0.686 & nan \\ 
 & DICE & \textit{0.407} & 2.471 & \textbf{0.921} & 0.726 & \underline{0.710} & \textbf{0.583} \\ 
 & FACE & 0.436 & \textbf{2.161} & \underline{0.924} & \textbf{0.000} & \textit{0.690} & \textit{0.699} \\ 
 & CLUE & nan & nan & nan & nan & nan & nan \\ 
 \cmidrule{2-8}
 & uncertainty & \textbf{0.025} & 6.397 & 0.985 & 4.542 & \textbf{0.765} & 2.346 \\ 
 & uncertainty+d & \underline{0.404} & \underline{2.170} & 0.984 & \underline{0.066} & 0.664 & \underline{0.674} \\ 
\midrule
\multirow{6}{*}{FICO} & GS & 0.487 & \textit{3.349} & 0.999 & \textit{0.767} & \textit{0.722} & nan \\ 
 & DICE & \underline{0.250} & 3.666 & \underline{0.863} & 0.943 & \underline{0.723} & \textit{0.935} \\ 
 & FACE & 0.396 & \textbf{2.269} & \textbf{0.827} & \textbf{0.000} & 0.711 & \textbf{0.564} \\ 
 & CLUE & nan & nan & nan & nan & nan & nan \\ 
 \cmidrule{2-8}
 & uncertainty & \textbf{0.021} & 5.384 & \textit{0.998} & 3.501 & 0.714 & 1.542 \\ 
 & uncertainty+d & \textit{0.342} & \underline{2.586} & 1.000 & \underline{0.502} & \textbf{0.745} & \underline{0.840} \\ 
\midrule
\multirow{6}{*}{home} & GS & nan & nan & nan & nan & nan & nan \\ 
 & DICE & \underline{0.125} & \underline{5.253} & \textbf{0.924} & \underline{2.440} & \textit{0.775} & nan \\ 
 & FACE & nan & nan & nan & nan & nan & nan \\ 
 & CLUE & nan & nan & nan & nan & nan & nan \\ 
 \cmidrule{2-8}
 & uncertainty & \textbf{0.007} & \textit{191.290} & \textit{0.993} & \textit{189.957} & \textbf{0.823} & \underline{20.652} \\ 
 & uncertainty+d & \textit{0.234} & \textbf{3.658} & \underline{0.952} & \textbf{1.604} & \underline{0.809} & \textbf{1.940} \\ 
\midrule
\multirow{6}{*}{housing} & GS & 0.494 & 10.314 & \textit{1.000} & 4.779 & 0.721 & nan \\ 
 & DICE & 0.372 & \textit{8.332} & \underline{0.886} & \underline{1.371} & 0.782 & \underline{1.054} \\ 
 & FACE & \underline{0.336} & \textbf{4.893} & \textbf{0.881} & \textbf{0.000} & \underline{0.805} & \textbf{0.652} \\ 
 & CLUE & nan & nan & nan & nan & nan & nan \\ 
 \cmidrule{2-8}
 & uncertainty & \textbf{0.029} & 33.556 & \textit{1.000} & 29.378 & \textit{0.799} & 2.523 \\ 
 & uncertainty+d & \textit{0.338} & \underline{6.144} & \textit{1.000} & \textit{1.769} & \textbf{0.827} & \textit{1.187} \\ 
\midrule
\multirow{6}{*}{titanic} & GS & 0.499 & \textit{5.034} & 0.706 & \textit{1.883} & \textit{0.718} & nan \\ 
 & DICE & \underline{0.448} & 8.802 & \underline{0.542} & 1.939 & \underline{0.719} & \textit{2.111} \\ 
 & FACE & \textit{0.489} & \textbf{2.618} & \textbf{0.480} & \textbf{0.000} & 0.675 & \textbf{0.618} \\ 
 & CLUE & nan & nan & nan & nan & nan & nan \\ 
 \cmidrule{2-8}
 & uncertainty & \textbf{0.007} & 58.392 & 0.616 & 55.691 & \textbf{0.756} & 9.601 \\ 
 & uncertainty+d & 0.490 & \underline{3.827} & \textit{0.568} & \underline{0.423} & 0.639 & \underline{1.875} \\ 
\bottomrule
\end{tabular}
\end{table*}

\clearpage

\begin{table*}[p]%
\centering
    \scriptsize%
    \setlength{\tabcolsep}{2.55pt}%
\caption{%
      Detailed evaluation results %
      based on the \emph{deep} ensemble %
      of our %
      uncertainty-based counterfactual desiderata, as defined in \autoref{sec:methodology:cf}, %
      for the six metrics listed in \autoref{apx:eval-metrics}. %
}
\label{tab:deep_ensemble_results_properties}
\begin{tabular}{@{}l l r r r r r r@{}}
\toprule
\multirow{2.5}{*}{Data set} & \multirow{2.5}{*}{Property} & \multicolumn{6}{c}{Metric} \\
\cmidrule(lr){3-8}
 &  & validity $\downarrow$ & similarity $\downarrow$ & sparsity $\downarrow$ & plausibility $\downarrow$ & discriminativeness $\uparrow$ & stability $\uparrow$ \\ 
\midrule
\multirow{7}{*}{bank} & validity & 0.135 & \textit{11.900} & \textbf{0.900} & 7.431 & 0.406 & \textit{1.813} \\ 
 & connectedness & 0.135 & \textbf{11.858} & \textbf{0.900} & \textit{7.426} & 0.410 & \textbf{1.809} \\ 
 & robustness & \textbf{0.101} & 12.221 & \textbf{0.900} & 7.867 & \textbf{0.436} & 2.031 \\ 
 & feasibility & 0.135 & \underline{11.891} & \textbf{0.900} & \underline{7.426} & \textit{0.410} & \underline{1.813} \\ 
 & discriminativeness & \underline{0.115} & 12.332 & \textbf{0.900} & 7.753 & 0.402 & 2.026 \\ 
 & plausibility & \textit{0.135} & 11.919 & \textbf{0.900} & 7.436 & 0.407 & 1.821 \\ 
 & similarity & 0.143 & 11.920 & \textbf{0.900} & \textbf{7.380} & \underline{0.421} & 1.848 \\ 
\midrule
\multirow{7}{*}{cancer} & validity & 0.068 & \textit{725.170} & \textbf{1.000} & \textit{723.992} & \textit{0.791} & \underline{122.874} \\ 
 & connectedness & 0.068 & 725.177 & \textbf{1.000} & 723.999 & \textit{0.791} & \textbf{122.865} \\ 
 & robustness & \textbf{0.055} & \underline{693.422} & \textbf{1.000} & \underline{692.232} & \textbf{0.791} & 155.964 \\ 
 & feasibility & \textit{0.068} & 725.738 & \textbf{1.000} & 724.607 & \textit{0.791} & 123.929 \\ 
 & discriminativeness & \underline{0.058} & 757.285 & \textbf{1.000} & 756.108 & \textbf{0.791} & 126.667 \\ 
 & plausibility & 0.068 & 725.195 & \textbf{1.000} & 724.015 & \textit{0.791} & \textit{122.903} \\ 
 & similarity & 0.080 & \textbf{603.140} & \textbf{1.000} & \textbf{601.997} & \underline{0.791} & 168.799 \\ 
\midrule
\multirow{7}{*}{churn} & validity & 0.087 & \textit{13.759} & \textbf{0.265} & \textit{11.834} & 0.796 & 3.311 \\ 
 & connectedness & 0.087 & \underline{13.759} & \textbf{0.265} & \underline{11.833} & 0.796 & \underline{3.308} \\ 
 & robustness & \underline{0.074} & 14.210 & \textbf{0.265} & 12.277 & \textbf{0.810} & \textbf{3.006} \\ 
 & feasibility & 0.087 & 13.760 & \textbf{0.265} & 11.835 & 0.796 & \textit{3.311} \\ 
 & discriminativeness & \textbf{0.068} & 14.520 & \textbf{0.265} & 12.584 & \underline{0.810} & 3.424 \\ 
 & plausibility & \textit{0.086} & 13.762 & \textbf{0.265} & 11.837 & \textit{0.796} & 3.312 \\ 
 & similarity & 0.107 & \textbf{13.498} & \textbf{0.265} & \textbf{11.561} & 0.786 & 3.527 \\ 
\midrule
\multirow{7}{*}{COMPAS} & validity & 0.030 & \underline{17.063} & \textbf{0.570} & \textit{5.496} & \textbf{0.741} & \textit{3.292} \\ 
 & connectedness & 0.030 & 17.063 & \textbf{0.570} & \underline{5.496} & \textit{0.741} & 3.292 \\ 
 & robustness & \textbf{0.022} & 19.001 & \textbf{0.570} & 7.201 & 0.729 & 3.360 \\ 
 & feasibility & 0.030 & \textbf{17.063} & \textbf{0.570} & 5.496 & \underline{0.741} & \textbf{3.291} \\ 
 & discriminativeness & \underline{0.022} & 17.687 & \textbf{0.570} & 6.108 & 0.737 & 3.331 \\ 
 & plausibility & \textit{0.030} & \textit{17.063} & \textbf{0.570} & 5.496 & \underline{0.741} & \underline{3.292} \\ 
 & similarity & 0.101 & 17.224 & \textbf{0.570} & \textbf{5.486} & 0.736 & 5.182 \\ 
\midrule
\multirow{7}{*}{diabetes} & validity & 0.044 & \underline{3.173} & \textbf{0.985} & \textit{1.383} & \textit{0.744} & 1.136 \\ 
 & connectedness & \textit{0.044} & 3.174 & \textbf{0.985} & 1.384 & \textit{0.744} & \underline{1.136} \\ 
 & robustness & \textbf{0.015} & 4.460 & \textbf{0.985} & 2.551 & \textbf{0.759} & \textbf{1.133} \\ 
 & feasibility & 0.044 & \textbf{3.173} & \textbf{0.985} & \underline{1.383} & \textit{0.744} & \textit{1.136} \\ 
 & discriminativeness & \underline{0.035} & 3.284 & \textbf{0.985} & 1.501 & \underline{0.746} & 1.185 \\ 
 & plausibility & 0.044 & \textit{3.173} & \textbf{0.985} & 1.383 & \textit{0.744} & 1.136 \\ 
 & similarity & 0.077 & 3.183 & \textbf{0.985} & \textbf{1.351} & 0.733 & 1.269 \\ 
\midrule
\multirow{7}{*}{FICO} & validity & 0.038 & \textit{5.521} & \textbf{0.998} & \underline{3.796} & 0.715 & 1.151 \\ 
 & connectedness & 0.038 & 5.521 & \textbf{0.998} & 3.796 & 0.715 & \textit{1.151} \\ 
 & robustness & \textbf{0.029} & 6.537 & \textbf{0.998} & 4.803 & \textbf{0.717} & \textbf{1.122} \\ 
 & feasibility & 0.038 & 5.521 & \textbf{0.998} & 3.796 & \textit{0.715} & \underline{1.151} \\ 
 & discriminativeness & \underline{0.030} & 5.822 & \textbf{0.998} & 4.084 & \underline{0.715} & 1.182 \\ 
 & plausibility & \textit{0.038} & \underline{5.521} & \textbf{0.998} & \textit{3.796} & 0.715 & 1.151 \\ 
 & similarity & 0.060 & \textbf{4.829} & \textbf{0.998} & \textbf{3.076} & 0.713 & 1.331 \\ 
\midrule
\multirow{7}{*}{home} & validity & 0.206 & 37.392 & \textbf{0.984} & 35.884 & \textbf{0.808} & 4.446 \\ 
 & connectedness & 0.210 & \textit{36.909} & \textbf{0.984} & \textit{35.399} & \textbf{0.808} & \textit{4.373} \\ 
 & robustness & 0.215 & \textbf{36.319} & \underline{0.988} & \textbf{34.837} & \textit{0.801} & 4.975 \\ 
 & feasibility & 0.206 & \underline{36.539} & \textbf{0.984} & \underline{35.028} & \textbf{0.808} & 4.864 \\ 
 & discriminativeness & \textbf{0.199} & 38.762 & \textbf{0.984} & 37.255 & \underline{0.808} & \underline{4.325} \\ 
 & plausibility & \underline{0.204} & 37.522 & \textbf{0.984} & 36.012 & \underline{0.808} & \textbf{4.284} \\ 
 & similarity & \textit{0.205} & 37.739 & \textbf{0.984} & 36.231 & \textbf{0.808} & 4.519 \\ 
\midrule
\multirow{7}{*}{housing} & validity & 0.051 & \textit{23.392} & \textbf{1.000} & \underline{19.390} & \textbf{0.800} & 2.528 \\ 
 & connectedness & 0.051 & \underline{23.392} & \textbf{1.000} & \textit{19.390} & \textbf{0.800} & 2.528 \\ 
 & robustness & \textbf{0.041} & 27.736 & \textbf{1.000} & 23.783 & 0.797 & \textbf{2.135} \\ 
 & feasibility & \textit{0.051} & 23.394 & \textbf{1.000} & 19.392 & \textbf{0.800} & \underline{2.527} \\ 
 & discriminativeness & \underline{0.042} & 24.797 & \textbf{1.000} & 20.793 & \underline{0.799} & 2.615 \\ 
 & plausibility & 0.051 & 23.393 & \textbf{1.000} & 19.391 & \textbf{0.800} & \textit{2.528} \\ 
 & similarity & 0.071 & \textbf{19.883} & \textbf{1.000} & \textbf{15.767} & \textit{0.798} & 3.277 \\ 
\midrule
\multirow{7}{*}{titanic} & validity & 0.040 & \textit{36.003} & \textbf{0.617} & \textit{31.924} & \textbf{0.759} & \textit{13.880} \\ 
 & connectedness & 0.040 & \underline{36.001} & \textbf{0.617} & \underline{31.923} & \textbf{0.759} & \textbf{13.879} \\ 
 & robustness & \textbf{0.020} & 42.743 & \textbf{0.617} & 38.566 & \textbf{0.759} & 14.116 \\ 
 & feasibility & \textit{0.040} & 36.005 & \textbf{0.617} & 31.927 & \textbf{0.759} & \underline{13.879} \\ 
 & discriminativeness & \underline{0.032} & 39.907 & \textbf{0.617} & 35.790 & \textit{0.758} & 15.173 \\ 
 & plausibility & 0.040 & 36.003 & \textbf{0.617} & 31.925 & \textbf{0.759} & 13.880 \\ 
 & similarity & 0.050 & \textbf{34.670} & \textbf{0.617} & \textbf{30.957} & \underline{0.759} & 14.386 \\ 
\bottomrule
\end{tabular}
\end{table*}

\clearpage

\begin{table*}[p]%
\centering
    \scriptsize%
    \setlength{\tabcolsep}{2.55pt}%
\caption{%
      Detailed evaluation results %
      based on the \emph{deep} ensemble %
      of counterfactual explainers, with \emph{uncertainty} and \emph{uncertainty plus distance} (uncertainty+d) as defined in \autoref{sec:methodology:gen}, %
      for the six metrics listed in \autoref{apx:eval-metrics}. %
}
\label{tab:deep_ensemble_results_explainers}
\begin{tabular}{@{}l l r r r r r r@{}}
\toprule
\multirow{2.5}{*}{Data set} & \multirow{2.5}{*}{Explainer} & \multicolumn{6}{c}{Metric} \\
\cmidrule(lr){3-8}
 &  & validity $\downarrow$ & similarity $\downarrow$ & sparsity $\downarrow$ & plausibility $\downarrow$ & discriminativeness $\uparrow$ & stability $\uparrow$ \\ 
\midrule
\multirow{6}{*}{bank} & GS & 0.484 & \textit{253.432} & 0.998 & \underline{2.535} & \textit{0.566} & \textbf{0.921} \\ 
 & DICE & \textit{0.132} & 8954.504 & \textbf{0.732} & 1641.651 & \textbf{0.639} & 4223.271 \\ 
 & FACE & 0.337 & 277.565 & \underline{0.876} & \textbf{0.000} & 0.540 & 132.212 \\ 
 & CLUE & nan & nan & nan & nan & nan & nan \\ 
 \cmidrule{2-8}
 & uncertainty & \underline{0.102} & \underline{12.108} & \textit{0.900} & 7.797 & 0.425 & \textit{2.015} \\ 
 & uncertainty+d & \textbf{0.075} & \textbf{6.944} & \textit{0.900} & \textit{4.988} & \underline{0.581} & \underline{1.239} \\ 
\midrule
\multirow{6}{*}{cancer} & GS & 0.482 & 358.210 & \underline{1.000} & 356.850 & 0.649 & nan \\ 
 & DICE & \textit{0.088} & \underline{2.396} & \textbf{0.999} & \underline{0.744} & 0.778 & \underline{0.699} \\ 
 & FACE & 0.295 & \textbf{2.048} & \underline{1.000} & \textbf{0.000} & \underline{0.841} & \textbf{0.629} \\ 
 & CLUE & nan & nan & nan & nan & nan & nan \\ 
 \cmidrule{2-8}
 & uncertainty & \textbf{0.056} & 695.644 & \underline{1.000} & 694.447 & \textit{0.791} & 156.913 \\ 
 & uncertainty+d & \underline{0.075} & \textit{303.835} & \underline{1.000} & \textit{302.380} & \textbf{0.899} & \textit{118.131} \\ 
\midrule
\multirow{6}{*}{churn} & GS & 0.485 & \underline{7.355} & \textit{0.278} & \underline{5.535} & \textit{0.677} & \textit{1.718} \\ 
 & DICE & nan & nan & nan & nan & nan & nan \\ 
 & FACE & nan & nan & nan & nan & nan & nan \\ 
 & CLUE & \underline{0.173} & 3895.622 & 1.000 & 3892.211 & \textbf{0.884} & \textbf{0.000} \\ 
 \cmidrule{2-8}
 & uncertainty & \textbf{0.074} & \textit{14.167} & \textbf{0.265} & \textit{12.233} & \underline{0.809} & 2.994 \\ 
 & uncertainty+d & \textit{0.385} & \textbf{2.900} & \underline{0.268} & \textbf{1.835} & 0.258 & \underline{0.950} \\ 
\midrule
\multirow{6}{*}{COMPAS} & GS & 0.622 & 15.622 & \textit{0.535} & \underline{0.416} & 0.743 & nan \\ 
 & DICE & \textit{0.432} & \underline{12.457} & \underline{0.437} & \textbf{0.000} & \textbf{0.787} & \textbf{0.192} \\ 
 & FACE & \underline{0.371} & \textbf{10.007} & \textbf{0.396} & \textbf{0.000} & \underline{0.782} & \textit{3.191} \\ 
 & CLUE & nan & nan & nan & nan & nan & nan \\ 
 \cmidrule{2-8}
 & uncertainty & \textbf{0.024} & 18.684 & 0.570 & 6.845 & 0.731 & 3.428 \\ 
 & uncertainty+d & 0.432 & \textit{14.454} & 0.590 & \textit{1.668} & \textit{0.776} & \underline{1.793} \\ 
\midrule
\multirow{6}{*}{diabetes} & GS & 0.495 & \textbf{2.107} & \textit{0.973} & \textit{0.401} & 0.674 & nan \\ 
 & DICE & \textit{0.379} & 2.619 & \textbf{0.928} & 0.782 & \textit{0.719} & \textbf{0.662} \\ 
 & FACE & 0.410 & \underline{2.279} & \underline{0.934} & \textbf{0.000} & 0.683 & \underline{0.720} \\ 
 & CLUE & nan & nan & nan & nan & nan & nan \\ 
 \cmidrule{2-8}
 & uncertainty & \textbf{0.017} & 4.525 & 0.985 & 2.605 & \textbf{0.758} & 1.244 \\ 
 & uncertainty+d & \underline{0.357} & \textit{2.506} & 1.000 & \underline{0.254} & \underline{0.756} & \textit{0.818} \\ 
\midrule
\multirow{6}{*}{FICO} & GS & 0.489 & \underline{3.855} & 0.999 & \underline{1.409} & \underline{0.748} & nan \\ 
 & DICE & \textit{0.337} & \textit{4.722} & \underline{0.858} & 1.907 & \textit{0.728} & \underline{1.070} \\ 
 & FACE & 0.388 & \textbf{2.355} & \textbf{0.833} & \textbf{0.000} & 0.724 & \textbf{0.574} \\ 
 & CLUE & nan & nan & nan & nan & nan & nan \\ 
 \cmidrule{2-8}
 & uncertainty & \textbf{0.030} & 6.453 & \textit{0.998} & 4.720 & 0.717 & \textit{1.113} \\ 
 & uncertainty+d & \underline{0.239} & 5.345 & 1.000 & \textit{1.781} & \textbf{0.867} & 1.891 \\ 
\midrule
\multirow{6}{*}{home} & GS & nan & nan & nan & nan & nan & nan \\ 
 & DICE & \underline{0.166} & \underline{4.910} & \textbf{0.916} & \underline{2.142} & 0.773 & \textbf{1.597} \\ 
 & FACE & 0.341 & \textbf{1.991} & \underline{0.925} & \textbf{0.000} & \textbf{0.838} & nan \\ 
 & CLUE & nan & nan & nan & nan & nan & nan \\ 
 \cmidrule{2-8}
 & uncertainty & \textit{0.204} & \textit{38.064} & \textit{0.984} & \textit{36.556} & \textit{0.808} & \underline{4.491} \\ 
 & uncertainty+d & \textbf{0.135} & 39.884 & 0.989 & 38.411 & \underline{0.821} & \textit{12.442} \\ 
\midrule
\multirow{6}{*}{housing} & GS & 0.489 & 11.404 & \textit{1.000} & \textit{5.614} & 0.739 & nan \\ 
 & DICE & \textit{0.337} & \underline{8.877} & \textbf{0.892} & \underline{2.281} & 0.772 & \underline{1.182} \\ 
 & FACE & \underline{0.288} & \textbf{4.940} & \underline{0.907} & \textbf{0.000} & \underline{0.811} & \textbf{0.726} \\ 
 & CLUE & nan & nan & nan & nan & nan & nan \\ 
 \cmidrule{2-8}
 & uncertainty & \textbf{0.042} & 27.355 & \textit{1.000} & 23.397 & \textit{0.798} & 2.139 \\ 
 & uncertainty+d & 0.357 & \textit{10.526} & \textit{1.000} & 6.179 & \textbf{0.866} & \textit{1.828} \\ 
\midrule
\multirow{6}{*}{titanic} & GS & nan & nan & nan & nan & nan & nan \\ 
 & DICE & \underline{0.376} & \underline{4.609} & \underline{0.568} & \textbf{0.000} & \underline{0.753} & \textbf{0.384} \\ 
 & FACE & 0.447 & \textbf{3.182} & \textbf{0.566} & \textbf{0.000} & \textit{0.732} & \underline{1.098} \\ 
 & CLUE & nan & nan & nan & nan & nan & nan \\ 
 \cmidrule{2-8}
 & uncertainty & \textbf{0.023} & 41.101 & 0.617 & \textit{37.035} & \textbf{0.759} & 14.044 \\ 
 & uncertainty+d & \textit{0.385} & \textit{5.409} & \textit{0.607} & \underline{1.957} & 0.706 & \textit{2.080} \\ 
\bottomrule
\end{tabular}
\end{table*}

\clearpage

\begin{table*}[p]%
\centering
    \scriptsize%
    \setlength{\tabcolsep}{2.55pt}%
\caption{%
      Detailed evaluation results %
      based on the \emph{adversarial} ensemble %
      of our %
      uncertainty-based counterfactual desiderata, as defined in \autoref{sec:methodology:cf}, %
      for the six metrics listed in \autoref{apx:eval-metrics}. %
}
\label{tab:adversarial_ensemble_results_properties}
\begin{tabular}{@{}l l r r r r r r@{}}
\toprule
\multirow{2.5}{*}{Data set} & \multirow{2.5}{*}{Property} & \multicolumn{6}{c}{Metric} \\
\cmidrule(lr){3-8}
 &  & validity $\downarrow$ & similarity $\downarrow$ & sparsity $\downarrow$ & plausibility $\downarrow$ & discriminativeness $\uparrow$ & stability $\uparrow$ \\ 
\midrule
\multirow{7}{*}{bank} & validity & 0.070 & 70.664 & \textbf{0.900} & \textit{65.145} & 0.498 & \underline{6.282} \\ 
 & connectedness & 0.070 & \textit{70.153} & \textbf{0.900} & 65.160 & 0.508 & \textit{6.315} \\ 
 & robustness & \underline{0.059} & 76.765 & \textbf{0.900} & 70.508 & \textit{0.545} & \textbf{5.050} \\ 
 & feasibility & \textit{0.066} & \underline{59.446} & \textbf{0.900} & \underline{52.856} & \underline{0.558} & 8.248 \\ 
 & discriminativeness & \textbf{0.045} & 78.460 & \textbf{0.900} & 71.168 & 0.490 & 6.843 \\ 
 & plausibility & 0.066 & \textbf{57.493} & \textbf{0.900} & \textbf{51.699} & 0.492 & 7.909 \\ 
 & similarity & 0.282 & 375.423 & \textbf{0.900} & 253.736 & \textbf{0.717} & 137.695 \\ 
\midrule
\multirow{7}{*}{cancer} & validity & 0.054 & \textit{669.474} & \textbf{1.000} & \textit{668.185} & \textit{0.791} & \underline{106.325} \\ 
 & connectedness & 0.054 & 669.483 & \textbf{1.000} & 668.194 & \textit{0.791} & \textit{106.329} \\ 
 & robustness & \underline{0.049} & \textbf{645.957} & \textbf{1.000} & \textbf{644.602} & \underline{0.791} & 123.946 \\ 
 & feasibility & \textit{0.054} & 670.694 & \textbf{1.000} & 669.415 & \textit{0.791} & 106.818 \\ 
 & discriminativeness & \textbf{0.048} & 687.065 & \textbf{1.000} & 685.785 & \textbf{0.791} & 109.332 \\ 
 & plausibility & 0.054 & \underline{668.657} & \textbf{1.000} & \underline{667.368} & \textit{0.791} & \textbf{105.778} \\ 
 & similarity & 0.177 & 23849.760 & \textbf{1.000} & 23848.664 & 0.790 & 8341.913 \\ 
\midrule
\multirow{7}{*}{churn} & validity & 0.015 & \underline{26.195} & \textbf{0.265} & \underline{24.163} & \textit{0.883} & \textit{7.330} \\ 
 & connectedness & 0.015 & \textit{26.195} & \textbf{0.265} & \textit{24.164} & \textit{0.883} & 7.332 \\ 
 & robustness & \underline{0.011} & 28.389 & \textbf{0.265} & 26.362 & \underline{0.884} & \textbf{6.943} \\ 
 & feasibility & \textit{0.015} & 26.235 & \textbf{0.265} & 24.206 & 0.883 & 7.336 \\ 
 & discriminativeness & \textbf{0.007} & 29.819 & \textbf{0.265} & 27.799 & \textbf{0.884} & 8.124 \\ 
 & plausibility & 0.016 & \textbf{26.022} & \textbf{0.265} & \textbf{23.996} & 0.883 & \underline{7.225} \\ 
 & similarity & 0.241 & 101.063 & \textbf{0.265} & 97.851 & \textbf{0.884} & 32.070 \\ 
\midrule
\multirow{7}{*}{COMPAS} & validity & 0.011 & \textbf{23.076} & \textbf{0.570} & \underline{11.764} & \textbf{0.718} & 3.847 \\ 
 & connectedness & 0.011 & \underline{23.076} & \textbf{0.570} & \textbf{11.764} & \underline{0.718} & \textit{3.847} \\ 
 & robustness & \underline{0.008} & 26.320 & \textbf{0.570} & 15.284 & 0.712 & 3.960 \\ 
 & feasibility & 0.011 & 23.083 & \textbf{0.570} & 11.771 & \textit{0.718} & \textbf{3.846} \\ 
 & discriminativeness & \textbf{0.007} & 24.440 & \textbf{0.570} & 13.160 & 0.716 & 3.971 \\ 
 & plausibility & \textit{0.011} & \textit{23.077} & \textbf{0.570} & \textit{11.766} & \textbf{0.718} & \underline{3.846} \\ 
 & similarity & 0.161 & 75.892 & \underline{0.571} & 55.169 & 0.695 & 32.512 \\ 
\midrule
\multirow{7}{*}{diabetes} & validity & 0.005 & \textit{5.886} & \underline{0.985} & \underline{3.974} & 0.758 & \textbf{2.117} \\ 
 & connectedness & 0.005 & \textbf{5.880} & \underline{0.985} & \textbf{3.974} & 0.758 & \textit{2.117} \\ 
 & robustness & \underline{0.005} & 6.696 & \underline{0.985} & 4.722 & \underline{0.764} & 2.369 \\ 
 & feasibility & \textit{0.005} & \underline{5.884} & \underline{0.985} & 3.977 & \textit{0.758} & 2.120 \\ 
 & discriminativeness & \textbf{0.004} & 7.080 & \underline{0.985} & 5.102 & \textbf{0.768} & 2.541 \\ 
 & plausibility & 0.005 & 5.887 & \underline{0.985} & \textit{3.974} & 0.758 & \underline{2.117} \\ 
 & similarity & 0.143 & 5.951 & \textbf{0.985} & 4.093 & 0.757 & 2.760 \\ 
\midrule
\multirow{7}{*}{FICO} & validity & 0.033 & \textit{10.922} & \textbf{0.998} & 9.118 & 0.728 & 3.615 \\ 
 & connectedness & 0.033 & \underline{10.922} & \textbf{0.998} & \textit{9.118} & 0.728 & 3.614 \\ 
 & robustness & \underline{0.031} & 11.689 & \textbf{0.998} & 9.921 & \textbf{0.729} & \textbf{3.259} \\ 
 & feasibility & 0.034 & \textbf{10.916} & \textbf{0.998} & \underline{9.113} & 0.728 & \textit{3.612} \\ 
 & discriminativeness & \textbf{0.031} & 10.996 & \textbf{0.998} & 9.192 & \textit{0.729} & 3.623 \\ 
 & plausibility & \textit{0.033} & 10.929 & \textbf{0.998} & \textbf{9.077} & 0.728 & \underline{3.595} \\ 
 & similarity & 0.218 & 185.424 & \underline{0.998} & 181.113 & \underline{0.729} & 71.490 \\ 
\midrule
\multirow{7}{*}{home} & validity & \textit{0.207} & \textit{2.903} & \textbf{1.000} & \underline{0.051} & \textbf{0.880} & nan \\ 
 & connectedness & \textit{0.207} & \underline{2.903} & \textbf{1.000} & \textbf{0.051} & \textbf{0.880} & nan \\ 
 & robustness & 0.207 & 2.903 & \textbf{1.000} & \textit{0.051} & \textbf{0.880} & nan \\ 
 & feasibility & 0.207 & 2.903 & \textbf{1.000} & 0.051 & \textbf{0.880} & nan \\ 
 & discriminativeness & \textbf{0.207} & 2.903 & \textbf{1.000} & 0.051 & \textbf{0.880} & nan \\ 
 & plausibility & \underline{0.207} & 2.903 & \textbf{1.000} & 0.051 & \textbf{0.880} & nan \\ 
 & similarity & 0.207 & \textbf{2.898} & \textbf{1.000} & 0.051 & \textbf{0.880} & nan \\ 
\midrule
\multirow{7}{*}{housing} & validity & 0.023 & \textbf{32.976} & \textbf{1.000} & \textbf{29.029} & \textbf{0.799} & \underline{3.239} \\ 
 & connectedness & 0.023 & \underline{32.976} & \textbf{1.000} & \underline{29.029} & \textbf{0.799} & \textit{3.239} \\ 
 & robustness & \underline{0.020} & 35.580 & \textbf{1.000} & 31.567 & 0.797 & \textbf{2.467} \\ 
 & feasibility & \textit{0.023} & \textit{32.979} & \textbf{1.000} & 29.034 & \underline{0.798} & 3.241 \\ 
 & discriminativeness & \textbf{0.017} & 35.061 & \textbf{1.000} & 31.045 & \textit{0.798} & 3.315 \\ 
 & plausibility & 0.023 & 32.979 & \textbf{1.000} & \textit{29.034} & \textbf{0.799} & 3.240 \\ 
 & similarity & 0.129 & 470.227 & \textbf{1.000} & 450.304 & 0.763 & 87.831 \\ 
\midrule
\multirow{7}{*}{titanic} & validity & 0.002 & \textbf{70.453} & \textbf{0.623} & \textbf{65.992} & \textit{0.751} & \underline{19.251} \\ 
 & connectedness & 0.002 & \textit{70.498} & \textbf{0.623} & \textit{66.040} & \textit{0.751} & \textit{19.265} \\ 
 & robustness & \underline{0.001} & 82.394 & \textbf{0.623} & 76.972 & \textbf{0.753} & 22.694 \\ 
 & feasibility & 0.002 & 70.615 & \textbf{0.623} & 66.169 & \textit{0.751} & \textbf{19.096} \\ 
 & discriminativeness & \textbf{0.000} & 89.771 & \textbf{0.623} & 84.835 & 0.750 & 24.053 \\ 
 & plausibility & \textit{0.002} & \underline{70.485} & \textbf{0.623} & \underline{66.019} & \underline{0.751} & 19.272 \\ 
 & similarity & 0.019 & 114.165 & \textbf{0.623} & 109.747 & 0.750 & 46.916 \\ 
\bottomrule
\end{tabular}
\end{table*}

\clearpage

\begin{table*}[p]%
\centering
    \scriptsize%
    \setlength{\tabcolsep}{2.55pt}%
\caption{%
      Detailed evaluation results %
      based on the \emph{adversarial} ensemble %
      of counterfactual explainers, with \emph{uncertainty} and \emph{uncertainty plus distance} (uncertainty+d) as defined in \autoref{sec:methodology:gen}, %
      for the six metrics listed in \autoref{apx:eval-metrics}. %
}
\label{tab:adversarial_ensemble_results_explainers}
\begin{tabular}{@{}l l r r r r r r@{}}
\toprule
\multirow{2.5}{*}{Data set} & \multirow{2.5}{*}{Explainer} & \multicolumn{6}{c}{Metric} \\
\cmidrule(lr){3-8}
 &  & validity $\downarrow$ & similarity $\downarrow$ & sparsity $\downarrow$ & plausibility $\downarrow$ & discriminativeness $\uparrow$ & stability $\uparrow$ \\ 
\midrule
\multirow{6}{*}{bank} & GS & 0.488 & \textit{264.049} & 0.998 & \underline{13.091} & \underline{0.571} & \textbf{3.999} \\ 
 & DICE & \underline{0.297} & 2648.397 & \textbf{0.749} & 315.006 & \textbf{0.588} & 762.211 \\ 
 & FACE & \textit{0.394} & \underline{143.886} & \textit{0.901} & \textbf{0.000} & 0.508 & \textit{62.557} \\ 
 & CLUE & nan & nan & nan & nan & nan & nan \\ 
 \cmidrule{2-8}
 & uncertainty & \textbf{0.079} & \textbf{65.269} & \underline{0.900} & \textit{59.877} & \textit{0.533} & \underline{12.296} \\ 
 & uncertainty+d & nan & nan & nan & nan & nan & nan \\ 
\midrule
\multirow{6}{*}{cancer} & GS & 0.486 & \textit{229.332} & \underline{1.000} & \textit{228.162} & 0.647 & \textit{83.737} \\ 
 & DICE & \textit{0.175} & \underline{2.513} & \textbf{0.999} & \underline{0.868} & 0.772 & \textbf{0.686} \\ 
 & FACE & 0.357 & \textbf{2.336} & \textbf{0.999} & \textbf{0.000} & \textit{0.790} & \underline{0.801} \\ 
 & CLUE & nan & nan & nan & nan & nan & nan \\ 
 \cmidrule{2-8}
 & uncertainty & \textbf{0.052} & 637.917 & \underline{1.000} & 636.526 & \underline{0.791} & 127.706 \\ 
 & uncertainty+d & \underline{0.123} & 299.481 & \underline{1.000} & 297.798 & \textbf{0.883} & 120.468 \\ 
\midrule
\multirow{6}{*}{churn} & GS & \textit{0.484} & \underline{9.192} & \underline{0.274} & \underline{7.397} & 0.730 & \textit{2.299} \\ 
 & DICE & \underline{0.447} & \textbf{3.481} & \textit{0.310} & \textbf{1.254} & \textit{0.830} & \underline{0.777} \\ 
 & FACE & nan & nan & nan & nan & nan & nan \\ 
 & CLUE & 0.494 & \textit{10.101} & 1.000 & \textit{7.685} & \textbf{0.884} & \textbf{0.238} \\ 
 \cmidrule{2-8}
 & uncertainty & \textbf{0.034} & 27.403 & \textbf{0.265} & 25.379 & \underline{0.883} & 10.271 \\ 
 & uncertainty+d & nan & nan & nan & nan & nan & nan \\ 
\midrule
\multirow{6}{*}{COMPAS} & GS & 0.644 & \textit{22.709} & \textit{0.566} & \underline{1.058} & \textbf{0.838} & nan \\ 
 & DICE & \textit{0.460} & 30.408 & \underline{0.446} & \textit{1.235} & 0.738 & \textit{10.297} \\ 
 & FACE & \underline{0.401} & \textbf{11.381} & \textbf{0.425} & \textbf{0.000} & \textit{0.781} & \underline{3.117} \\ 
 & CLUE & nan & nan & nan & nan & nan & nan \\ 
 \cmidrule{2-8}
 & uncertainty & \textbf{0.047} & 24.314 & 0.570 & 12.100 & 0.718 & 17.123 \\ 
 & uncertainty+d & 0.471 & \underline{22.057} & 0.578 & 1.985 & \underline{0.796} & \textbf{2.454} \\ 
\midrule
\multirow{6}{*}{diabetes} & GS & 0.497 & \textbf{2.149} & \textit{0.971} & \textit{0.421} & 0.700 & nan \\ 
 & DICE & \textit{0.383} & \textit{2.665} & \textbf{0.923} & 0.819 & \textit{0.717} & \underline{0.695} \\ 
 & FACE & 0.445 & \underline{2.221} & \underline{0.942} & \textbf{0.000} & 0.696 & \textbf{0.647} \\ 
 & CLUE & nan & nan & nan & nan & nan & nan \\ 
 \cmidrule{2-8}
 & uncertainty & \textbf{0.056} & 12.347 & 0.985 & 10.251 & \textbf{0.771} & 5.644 \\ 
 & uncertainty+d & \underline{0.382} & 2.776 & 1.000 & \underline{0.218} & \underline{0.745} & \textit{0.840} \\ 
\midrule
\multirow{6}{*}{FICO} & GS & 0.492 & 4.295 & 0.999 & 1.928 & \textbf{0.739} & nan \\ 
 & DICE & \underline{0.338} & \textit{3.892} & \underline{0.860} & \underline{1.180} & 0.727 & \underline{1.090} \\ 
 & FACE & 0.430 & \textbf{2.389} & \textbf{0.842} & \textbf{0.000} & 0.725 & \textbf{0.612} \\ 
 & CLUE & nan & nan & nan & nan & nan & nan \\ 
 \cmidrule{2-8}
 & uncertainty & \textbf{0.037} & 11.571 & \textit{0.998} & 9.821 & \underline{0.728} & 3.525 \\ 
 & uncertainty+d & \textit{0.409} & \underline{3.362} & 0.999 & \textit{1.193} & \textit{0.728} & \textit{1.155} \\ 
\midrule
\multirow{6}{*}{home} & GS & nan & nan & nan & nan & nan & nan \\ 
 & DICE & \textbf{0.195} & \textit{4.750} & \textbf{0.923} & \textit{2.063} & \underline{0.785} & \textbf{1.420} \\ 
 & FACE & nan & nan & nan & nan & nan & nan \\ 
 & CLUE & nan & nan & nan & nan & nan & nan \\ 
 \cmidrule{2-8}
 & uncertainty & \underline{0.207} & \underline{2.903} & \underline{1.000} & \underline{0.051} & \textbf{0.880} & nan \\ 
 & uncertainty+d & \textit{0.207} & \textbf{2.897} & \underline{1.000} & \textbf{0.045} & \textbf{0.880} & nan \\ 
\midrule
\multirow{6}{*}{housing} & GS & 0.490 & 12.716 & \textit{1.000} & 6.773 & 0.751 & nan \\ 
 & DICE & 0.381 & \underline{8.652} & \underline{0.888} & \underline{1.798} & \textit{0.795} & \underline{1.074} \\ 
 & FACE & \underline{0.365} & \textbf{4.577} & \textbf{0.869} & \textbf{0.000} & 0.794 & \textbf{0.557} \\ 
 & CLUE & nan & nan & nan & nan & nan & nan \\ 
 \cmidrule{2-8}
 & uncertainty & \textbf{0.022} & 33.794 & \textit{1.000} & 29.691 & \underline{0.798} & 2.725 \\ 
 & uncertainty+d & \textit{0.367} & \textit{9.142} & \textit{1.000} & \textit{5.339} & \textbf{0.832} & \textit{2.281} \\ 
\midrule
\multirow{6}{*}{titanic} & GS & 0.499 & \textit{6.100} & 0.696 & 3.695 & 0.685 & nan \\ 
 & DICE & \underline{0.162} & 10.651 & \textbf{0.538} & \textit{2.352} & \textbf{0.766} & \underline{1.859} \\ 
 & FACE & 0.479 & \textbf{3.370} & \underline{0.588} & \textbf{0.000} & \textit{0.732} & \textbf{0.634} \\ 
 & CLUE & nan & nan & nan & nan & nan & nan \\ 
 \cmidrule{2-8}
 & uncertainty & \textbf{0.008} & 115.160 & \textit{0.620} & 111.107 & \underline{0.754} & 44.337 \\ 
 & uncertainty+d & \textit{0.406} & \underline{5.756} & 0.646 & \underline{1.487} & 0.646 & \textit{2.525} \\ 
\bottomrule
\end{tabular}
\end{table*}

\clearpage

\begin{figure*}[p]%
    \centering
    \includegraphics[width=.95\linewidth]{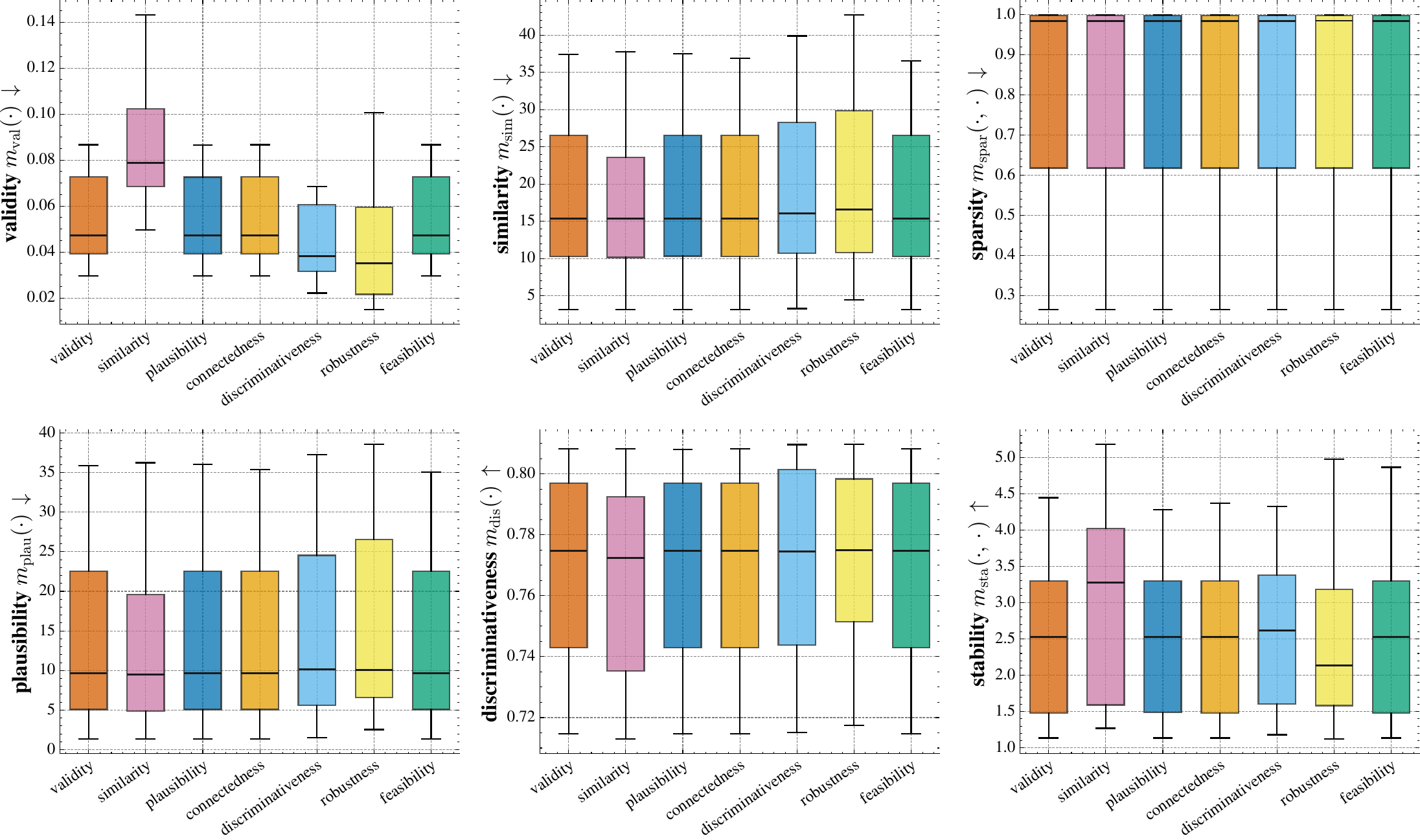}
    \caption{%
      Evaluation of our %
      uncertainty-based counterfactual desiderata (x-axis) -- as defined in \autoref{sec:methodology:cf} -- %
      for the six metrics (one per pane, y-axis) listed in \autoref{apx:eval-metrics}. %
      The experiments use the \emph{deep} ensemble, with the results aggregated across the nine data sets. %
    }
    \label{fig:deep_per_property}%
\end{figure*}

\begin{figure*}[p]%
    \centering
    \includegraphics[width=.95\linewidth]{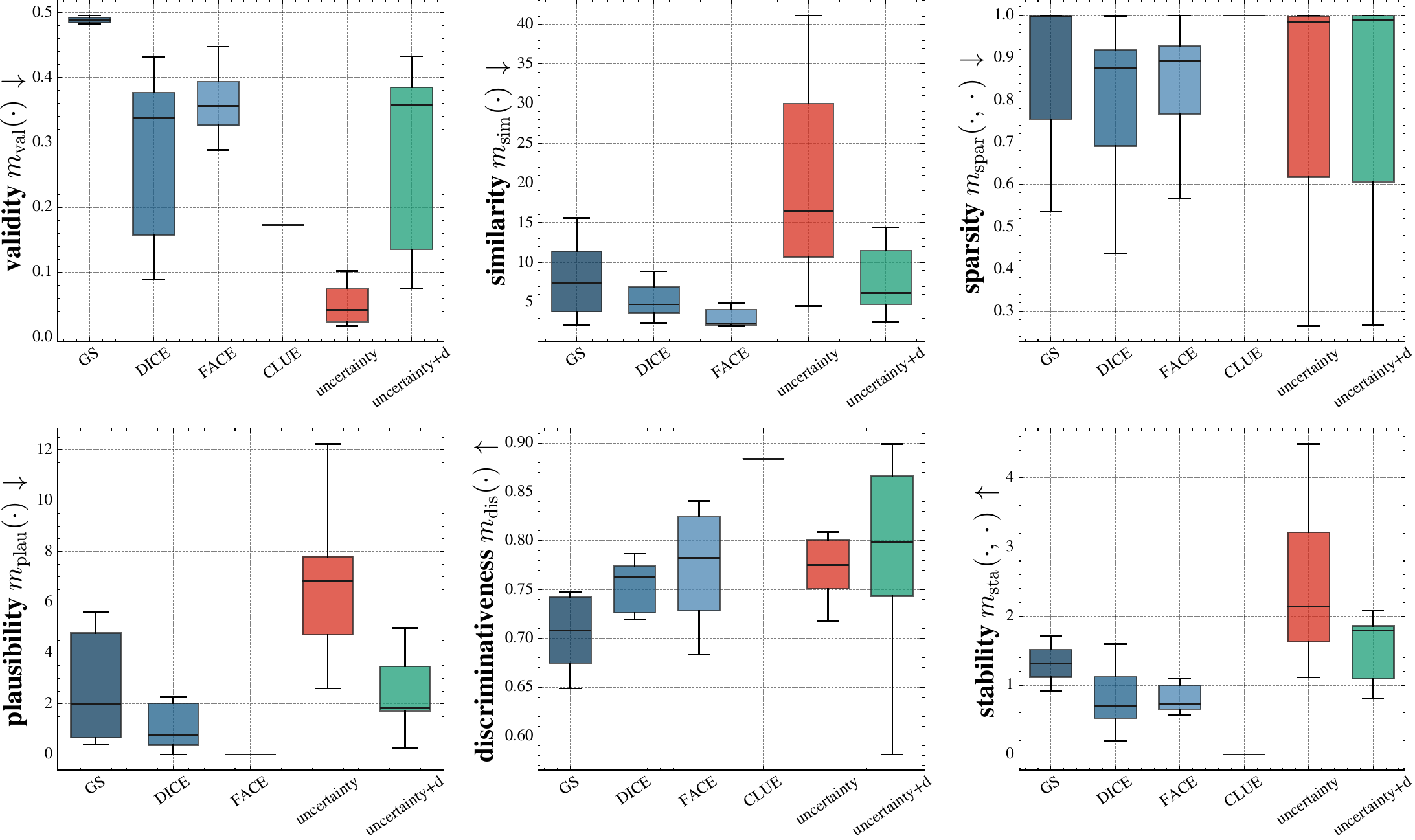}
    \caption{%
      Evaluation of counterfactual explainers (x-axis) -- with \emph{uncertainty} and \emph{uncertainty plus distance} (uncertainty+d) as defined in \autoref{sec:methodology:gen} -- %
      for the six metrics (one per pane, y-axis) listed in \autoref{apx:eval-metrics}. %
      The experiments use the \emph{deep} ensemble, with the results aggregated across the nine data sets. %
    }
    \label{fig:deep_all_datasets}%
\end{figure*}

\clearpage

\begin{figure*}[p]%
    \centering
    \includegraphics[width=.95\linewidth]{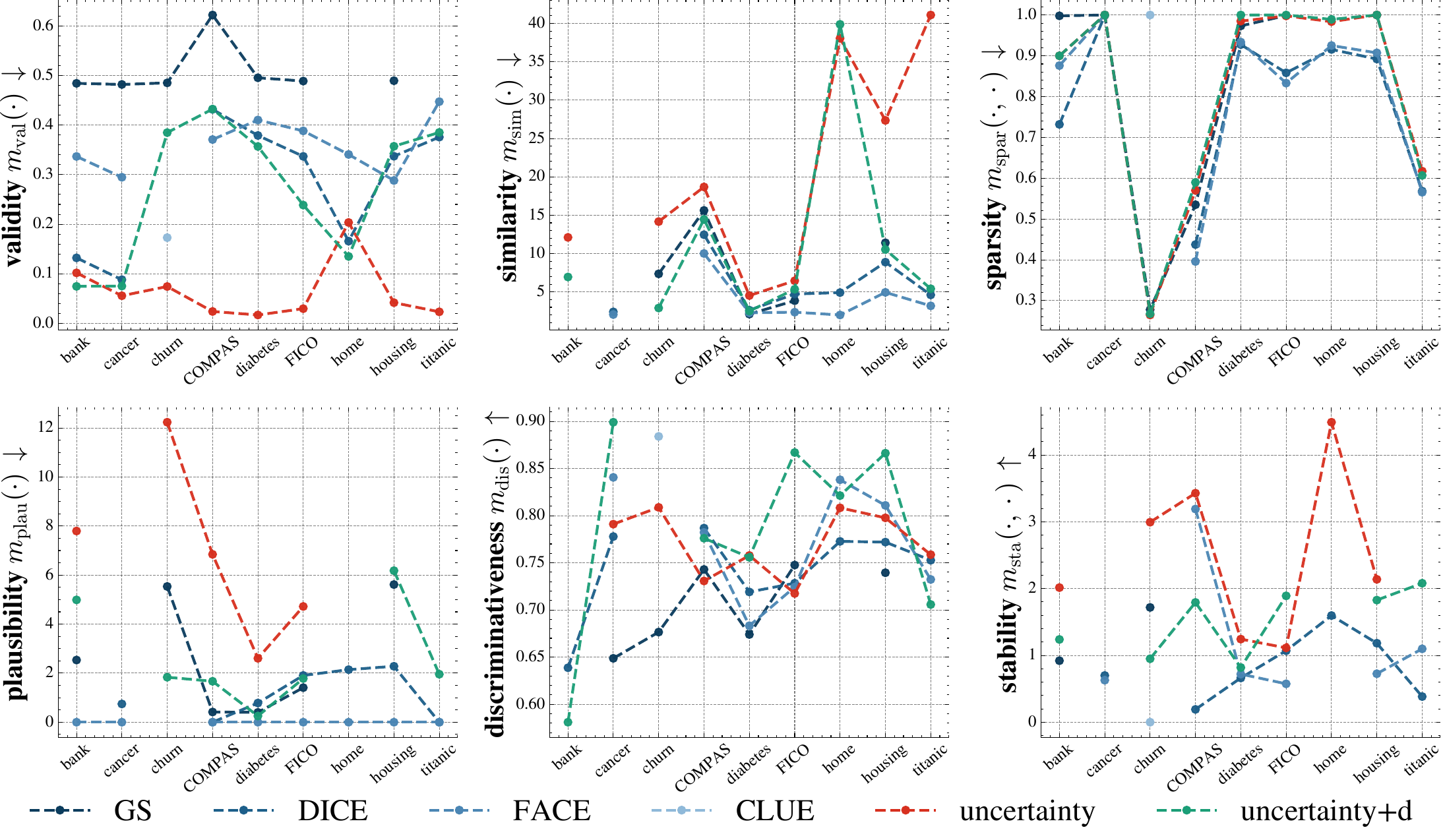}
    \caption{%
      Evaluation of counterfactual explainers %
      -- as described by the legend, with \emph{uncertainty} and \emph{uncertainty plus distance} (uncertainty+d) as defined in \autoref{sec:methodology:gen} -- %
      across the nine data sets (x-axis) %
      for the six metrics (one per pane, y-axis) listed in \autoref{apx:eval-metrics}. %
      The experiments use the \emph{deep} ensemble. %
      Note the missing results; %
      for CLUE these are due to it failing to find a counterfactual %
      and for the other explainers because of discarded outliers (i.e., values outside the interquartile range), %
      which were filtered out to improve the readability of the plots.
    }
    \label{fig:deep_individual_datasets}%
\end{figure*}

\clearpage

\begin{figure*}[t]%
    \centering
    \includegraphics[width=.95\linewidth]{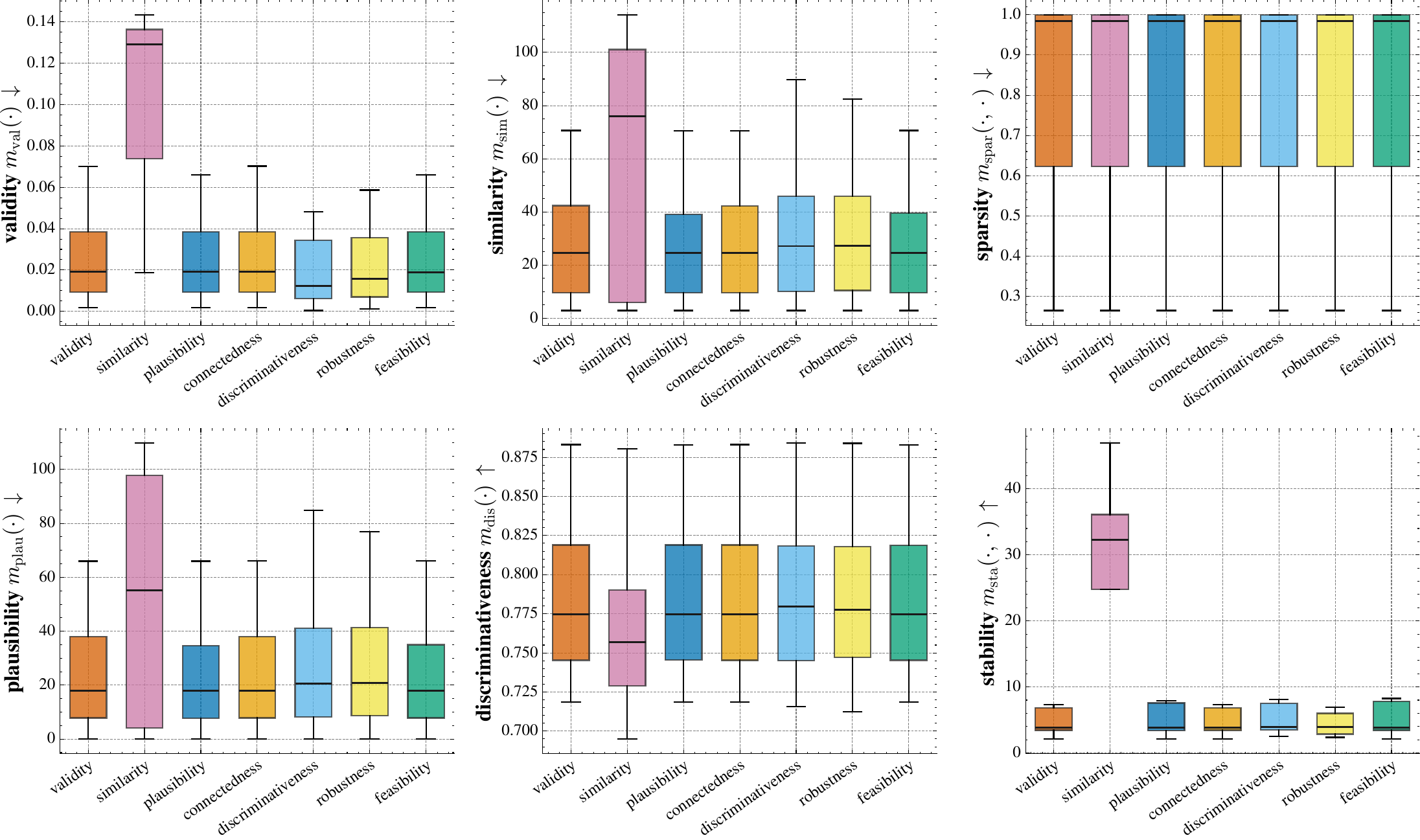}
    \caption{%
      Evaluation of our %
      uncertainty-based counterfactual desiderata (x-axis) -- as defined in \autoref{sec:methodology:cf} -- %
      for the six metrics (one per pane, y-axis) listed in \autoref{apx:eval-metrics}. %
      The experiments use the \emph{adversarial} ensemble, with the results aggregated across the nine data sets. %
    }
    \label{fig:adversarial_per_property}%
\end{figure*}

}

\section{Deep and Adversarial Ensembles\label[appendix]{apx:exp-full}}%

\autoref{sec:experiments} presents an overview of our experimental results for the \emph{DARE} ensemble; %
here, we complement this analysis %
by discussing the corresponding findings for %
the \emph{deep} and \emph{adversarial} ensembles. %
The former %
is a standard approach used to obtain uncertainty estimates based on a (deep) model ensemble, training each one of its members on a random shuffle of a data set. %
The latter uses adversarial training to further improve the resulting uncertainty estimates. %
See \autoref{apx:tables} for %
full evaluation outcomes %
of %
our uncertainty-based definitions of counterfactual desiderata (proposed in \autoref{sec:methodology:cf}) %
and %
the two variants of our uncertainty-aware explainer (introduced in \autoref{sec:methodology:gen}). %

\paragraph{Deep Ensemble} %
On the property level -- see \autoref{fig:deep_per_property} -- %
we observe that %
the deep ensemble delivers almost identical performance across the counterfactual desiderata definitions, indicating that both aleatoric and epistemic uncertainty have low impact on the optimisation procedure. %
This is because the deep ensemble has \emph{low diversity}, making the total uncertainty almost equal to its aleatoric component, with epistemic uncertainty being mostly close to $0$. %
This example clearly demonstrates that the proposed uncertainty-aware counterfactual desiderata (as well as the explainers built on top of them) can only achieve their full potential %
when the underlying uncertainty quantification and decomposition are reliable and robust. %

When evaluating explainers -- refer to \autoref{fig:deep_all_datasets} -- %
the deep ensemble provides noticeably different metric values as compared to the DARE experiments (\autoref{fig:overall_dare_ensemble}). %
Specifically, our uncertainty-only approach demonstrates poor \emph{stability} %
and also %
underperforms on \emph{similarity} and \emph{plausibility} despite the latter two metric scores exhibiting lower variance. %
Additionally, the uncertainty-only explainer performs worse than its distance-aware counterpart in terms of \emph{discriminativeness}, which is the opposite of what we observe for the DARE ensemble. %
As with our property-level findings %
these differences can be attributed to how the deep ensemble captures uncertainty. %

Across the individual data sets -- see \autoref{fig:deep_individual_datasets} -- %
both the deep and DARE (\autoref{fig:dare_individual_datasets}) ensembles deliver comparable performance. %
Note, however, the overall differences in the magnitude of metric values, except for \emph{discriminativeness}. %
For example, %
while the DARE ensemble results in \emph{similarity} of $\approx 60$ for \emph{titanic}, the deep ensemble provides a score of only $40$; %
the same behaviour can be observed for \emph{plausibility} on the \emph{churn} data set. %
In general, we still note similar trends across different data sets, with DARE showing higher overall variance. %

\begin{figure*}[t]%
    \centering
    \includegraphics[width=.95\linewidth]{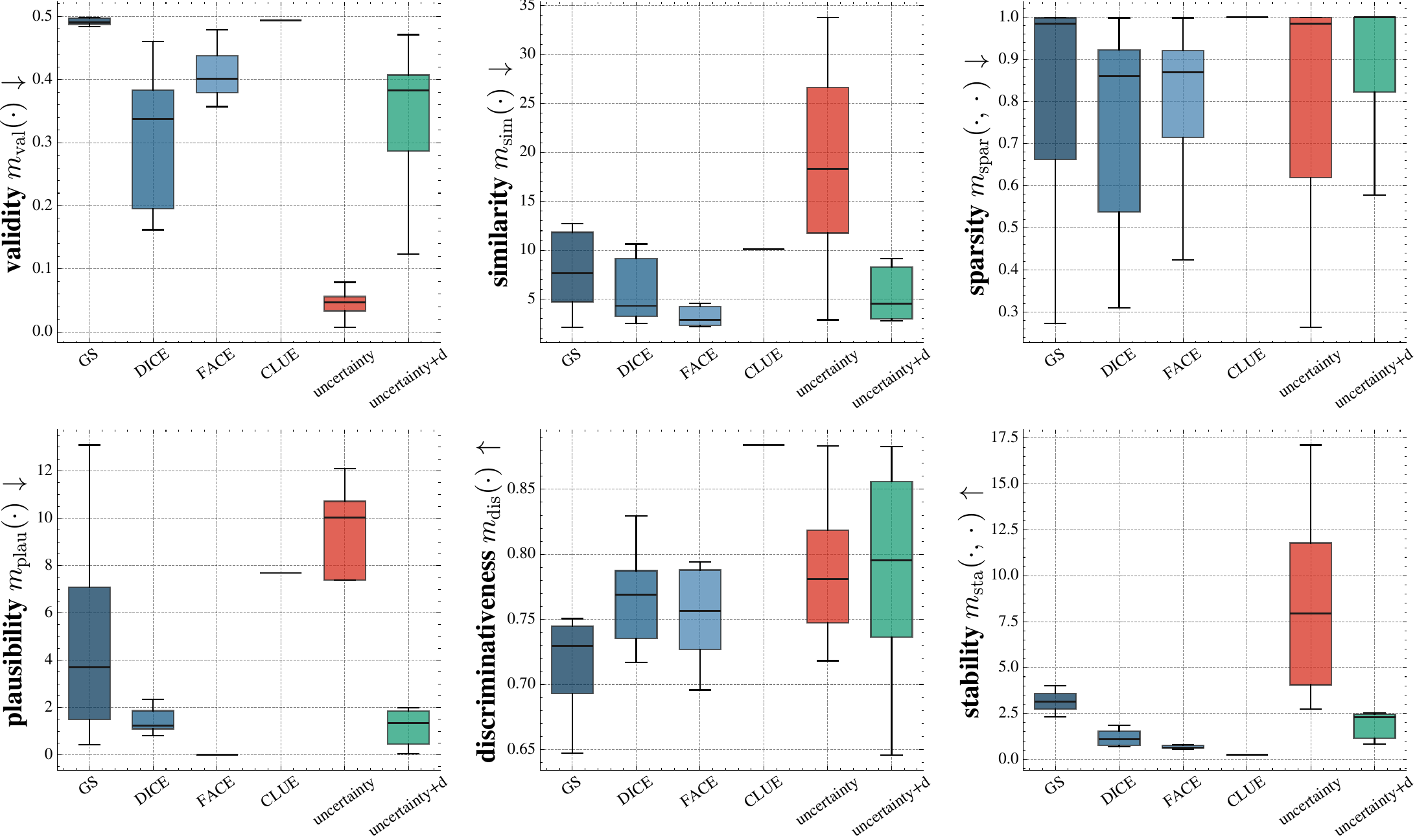}
    \caption{%
      Evaluation of counterfactual explainers (x-axis) -- with \emph{uncertainty} and \emph{uncertainty plus distance} (uncertainty+d) as defined in \autoref{sec:methodology:gen} -- %
      for the six metrics (one per pane, y-axis) listed in \autoref{apx:eval-metrics}. %
      The experiments use the \emph{adversarial} ensemble, with the results aggregated across the nine data sets. %
    }
    \label{fig:adversarial_all_datasets}
\end{figure*}

\begin{figure*}%
    \centering
    \includegraphics[width=.95\linewidth]{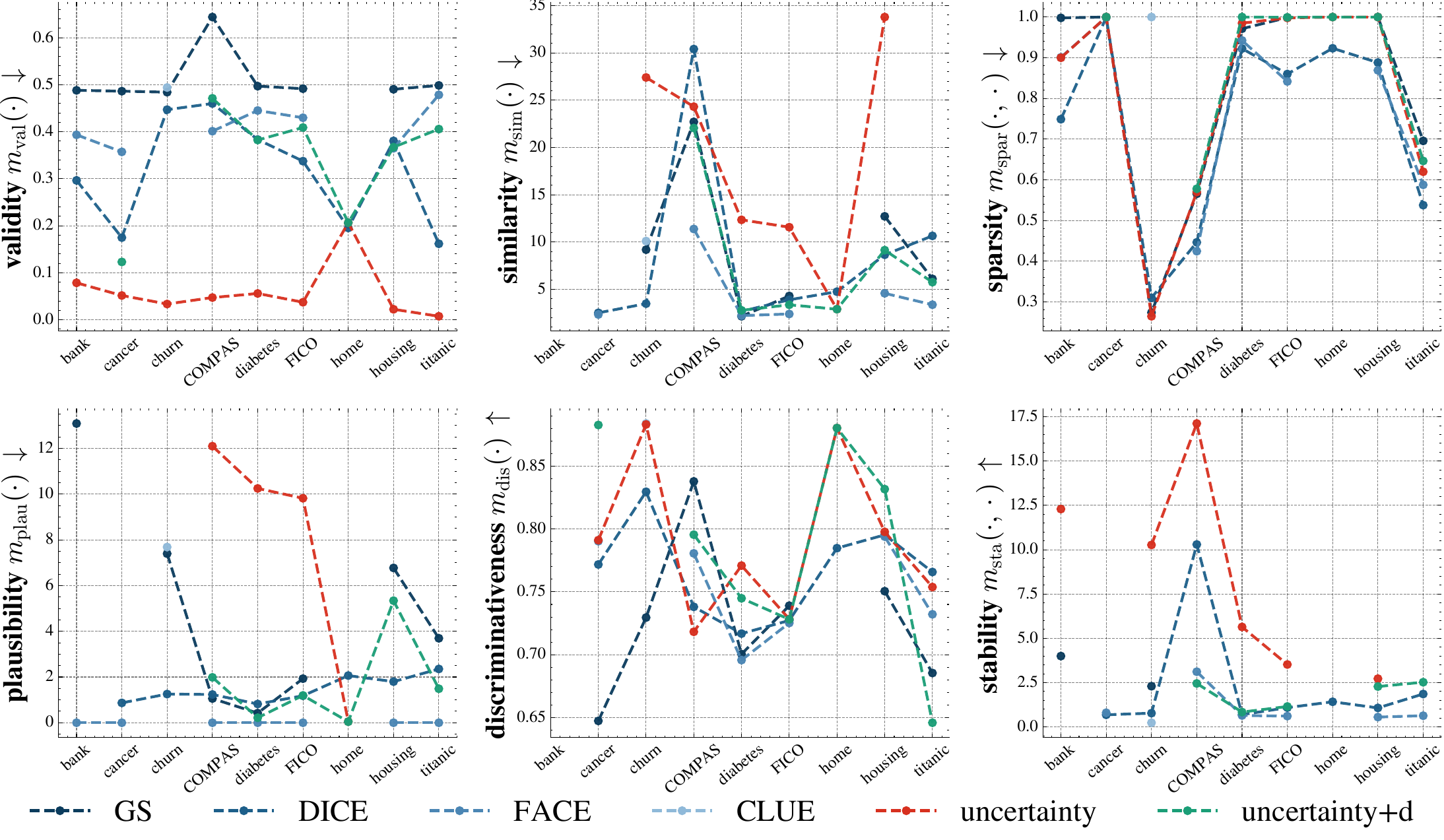}
    \caption{%
      Evaluation of counterfactual explainers %
      -- as described by the legend, with \emph{uncertainty} and \emph{uncertainty plus distance} (uncertainty+d) as defined in \autoref{sec:methodology:gen} -- %
      across the nine data sets (x-axis) %
      for the six metrics (one per pane, y-axis) listed in \autoref{apx:eval-metrics}. %
      The experiments use the \emph{adversarial} ensemble. %
      Note the missing results; %
      for CLUE these are due to it failing to find a counterfactual %
      and for the other explainers because of discarded outliers (i.e., values outside the interquartile range), %
      which were filtered out to improve the readability of the plots.
    }
    \label{fig:adversarial_individual_datasets}
\end{figure*}

\paragraph{Adversarial Ensemble} %
\autoref{fig:adversarial_per_property} %
depicts evaluation results of our uncertainty-based counterfactual desiderata %
for the adversarial ensemble. %
The most striking observation is how our definition of \emph{similarity} %
achieves markedly different metric scores (across the board, except for \emph{sparsity}) when juxtaposed with all the other properties; %
note that comparable behaviour, albeit somewhat less pronounced, can be seen for the DARE ensemble (\autoref{fig:dare_properties}). %
The metric scores for the definitions of the remaining desiderata, on the other hand, are visibly homogeneous, %
indicating that the epistemic component captured by the adversarial ensemble may not contribute much to the total uncertainty estimate. %
Both this and the deep ensemble thus appear to %
quantify epistemic uncertainty with much lower values as compared to the DARE ensemble, which adversely affects the optimisation objective that forms the backbone of our techniques. %
More research %
is clearly necessary to further explore and better understand this phenomenon. %

In terms of explainer performance -- %
see \autoref{fig:adversarial_all_datasets} -- %
the adversarial ensemble yields slight metric value changes when compared to the DARE ensemble (\autoref{fig:overall_dare_ensemble}). %
This is especially visible for the \emph{plausibility} and \emph{stability} metrics, with the remaining results being largely consistent across the two experimental set-ups. %
Regarding the former, %
the noticeable deterioration in its scores (especially for the GS explainer) %
signals broad interpolation in the feature space that %
can be attributed to its regions with low data density being assigned $p(\cplus | x) > 0.5$. %
Regarding the latter, %
one reason for this behaviour may be the reliance of the adversarial ensemble on larger feature space regions, which leads to higher estimates of the probability of the counterfactual class $p(\cplus | x)$, with more diverse points $x$ being assigned identical probability values as a consequence. %

On the level of individual data sets -- refer to \autoref{fig:adversarial_individual_datasets} -- we observe noticeable differences when comparing the adversarial and DARE (\autoref{fig:dare_individual_datasets}) ensembles (as was the case when evaluating the latter next to the deep ensemble). %
This is especially visible for the \emph{similarity} and \emph{plausibility} metrics on the \emph{home} data set. %
We can also see that for some metrics both instantiations of our uncertainty-aware explainer exhibit quite high variance across different data sets; %
note, however, that this behaviour is markedly less pronounced for the uncertainty plus distance flavour of our approach when using the DARE ensemble. %

\section*{Funding}
This research was supported by the 2024 DAAD \emph{Postdoc-NeT-AI -- Short-term Research Stay} programme. %
Additional funding was provided by the Hasler Foundation (2024-10-14-189) %
as well as the Swiss National Science Foundation through the TRUST-ME project (205121L 214991). %

\section*{Conflict of Interest}%
We declare no competing interests. %

\section*{Data Availability}
This research relies exclusively on publicly available data sets. %

\section*{Code Availability}
\github{}%

\printcredits

\bibliographystyle{cas-model2-names}

\bibliography{bib}

\begin{thebibliography}{64}
\expandafter\ifx\csname natexlab\endcsname\relax\def\natexlab#1{#1}\fi
\providecommand{\url}[1]{\texttt{#1}}
\providecommand{\href}[2]{#2}
\providecommand{\path}[1]{#1}
\providecommand{\DOIprefix}{doi:}
\providecommand{\ArXivprefix}{arXiv:}
\providecommand{\URLprefix}{URL: }
\providecommand{\Pubmedprefix}{pmid:}
\providecommand{\doi}[1]{\href{http://dx.doi.org/#1}{\path{#1}}}
\providecommand{\Pubmed}[1]{\href{pmid:#1}{\path{#1}}}
\providecommand{\bibinfo}[2]{#2}
\ifx\xfnm\relax \def\xfnm[#1]{\unskip,\space#1}\fi
%Type = Article
\bibitem[{Abdar et~al.(2021)Abdar, Pourpanah, Hussain, Rezazadegan, Liu, Ghavamzadeh, Fieguth, Cao, Khosravi, Acharya, Makarenkov and Nahavandi}]{abdar2021review}
\bibinfo{author}{Abdar, M.}, \bibinfo{author}{Pourpanah, F.}, \bibinfo{author}{Hussain, S.}, \bibinfo{author}{Rezazadegan, D.}, \bibinfo{author}{Liu, L.}, \bibinfo{author}{Ghavamzadeh, M.}, \bibinfo{author}{Fieguth, P.}, \bibinfo{author}{Cao, X.}, \bibinfo{author}{Khosravi, A.}, \bibinfo{author}{Acharya, U.R.}, \bibinfo{author}{Makarenkov, V.}, \bibinfo{author}{Nahavandi, S.}, \bibinfo{year}{2021}.
\newblock \bibinfo{title}{A review of uncertainty quantification in deep learning: {Techniques}, applications and challenges}.
\newblock \bibinfo{journal}{Information Fusion} \bibinfo{volume}{76}, \bibinfo{pages}{243--297}.
%Type = Inproceedings
\bibitem[{Antor{\'a}n et~al.(2021)Antor{\'a}n, Bhatt, Adel, Weller and Hern{\'a}ndez-Lobato}]{antoran2020getting}
\bibinfo{author}{Antor{\'a}n, J.}, \bibinfo{author}{Bhatt, U.}, \bibinfo{author}{Adel, T.}, \bibinfo{author}{Weller, A.}, \bibinfo{author}{Hern{\'a}ndez-Lobato, J.M.}, \bibinfo{year}{2021}.
\newblock \bibinfo{title}{Getting a {CLUE}: {A} method for explaining uncertainty estimates}, in: \bibinfo{booktitle}{Proceedings of the 9th International Conference on Learning Representations}, \bibinfo{publisher}{Curran Associates, Inc.}, \bibinfo{address}{Red Hook, NY, USA}. pp. \bibinfo{pages}{632--665}.
%Type = Inproceedings
\bibitem[{Bhatt et~al.(2021)Bhatt, Antor{\'a}n, Zhang, Liao, Sattigeri, Fogliato, Melan{\c{c}}on, Krishnan, Stanley, Tickoo, Nachman, Chunara, Srikumar, Weller and Xiang}]{bhatt2021uncertainty}
\bibinfo{author}{Bhatt, U.}, \bibinfo{author}{Antor{\'a}n, J.}, \bibinfo{author}{Zhang, Y.}, \bibinfo{author}{Liao, Q.V.}, \bibinfo{author}{Sattigeri, P.}, \bibinfo{author}{Fogliato, R.}, \bibinfo{author}{Melan{\c{c}}on, G.}, \bibinfo{author}{Krishnan, R.}, \bibinfo{author}{Stanley, J.}, \bibinfo{author}{Tickoo, O.}, \bibinfo{author}{Nachman, L.}, \bibinfo{author}{Chunara, R.}, \bibinfo{author}{Srikumar, M.}, \bibinfo{author}{Weller, A.}, \bibinfo{author}{Xiang, A.}, \bibinfo{year}{2021}.
\newblock \bibinfo{title}{Uncertainty as a form of transparency: {Measuring}, communicating, and using uncertainty}, in: \bibinfo{booktitle}{Proceedings of the AAAI/ACM Conference on AI, Ethics, and Society}, \bibinfo{publisher}{ACM}, \bibinfo{address}{New York, NY, USA}. pp. \bibinfo{pages}{401--413}.
%Type = Article
\bibitem[{Breiman(2001)}]{breiman2001statistical}
\bibinfo{author}{Breiman, L.}, \bibinfo{year}{2001}.
\newblock \bibinfo{title}{Statistical modeling: {The} two cultures}.
\newblock \bibinfo{journal}{Statistical Science} \bibinfo{volume}{16}, \bibinfo{pages}{199--231}.
%Type = Inproceedings
\bibitem[{Chau et~al.(2023)Chau, Muandet and Sejdinovic}]{chau2023explaining}
\bibinfo{author}{Chau, S.L.}, \bibinfo{author}{Muandet, K.}, \bibinfo{author}{Sejdinovic, D.}, \bibinfo{year}{2023}.
\newblock \bibinfo{title}{Explaining the uncertain: {Stochastic} {Shapley} values for {Gaussian} process models}, in: \bibinfo{booktitle}{Advances in Neural Information Processing Systems}, \bibinfo{publisher}{Curran Associates, Inc.}, \bibinfo{address}{Red Hook, NY, USA}. pp. \bibinfo{pages}{50769--50795}.
%Type = Article
\bibitem[{Chiaburu et~al.(2024)Chiaburu, Hau{\ss}er and Bie{\ss}mann}]{chiaburu2024uncertainty}
\bibinfo{author}{Chiaburu, T.}, \bibinfo{author}{Hau{\ss}er, F.}, \bibinfo{author}{Bie{\ss}mann, F.}, \bibinfo{year}{2024}.
\newblock \bibinfo{title}{Uncertainty in {XAI}: {Human} perception and modeling approaches}.
\newblock \bibinfo{journal}{Machine Learning and Knowledge Extraction} \bibinfo{volume}{6}, \bibinfo{pages}{1170--1192}.
%Type = Article
\bibitem[{Christodoulou and Sun(2026)}]{christodoulou2026impact}
\bibinfo{author}{Christodoulou, L.}, \bibinfo{author}{Sun, C.}, \bibinfo{year}{2026}.
\newblock \bibinfo{title}{The impact of machine learning uncertainty on the robustness of counterfactual explanations}.
\newblock \bibinfo{journal}{Expert Systems with Applications} \bibinfo{volume}{309}, \bibinfo{pages}{131198}.
%Type = Article
\bibitem[{Delaney et~al.(2021)Delaney, Greene and Keane}]{delaney2021uncertainty}
\bibinfo{author}{Delaney, E.}, \bibinfo{author}{Greene, D.}, \bibinfo{author}{Keane, M.T.}, \bibinfo{year}{2021}.
\newblock \bibinfo{title}{Uncertainty estimation and out-of-distribution detection for counterfactual explanations: {Pitfalls} and solutions}.
\newblock \bibinfo{journal}{ICML Workshop on Algorithmic Recourse} .
%Type = Inproceedings
\bibitem[{Depeweg et~al.(2018)Depeweg, Hernandez-Lobato, Doshi-Velez and Udluft}]{depeweg2018decomposition}
\bibinfo{author}{Depeweg, S.}, \bibinfo{author}{Hernandez-Lobato, J.M.}, \bibinfo{author}{Doshi-Velez, F.}, \bibinfo{author}{Udluft, S.}, \bibinfo{year}{2018}.
\newblock \bibinfo{title}{Decomposition of uncertainty in {Bayesian} deep learning for efficient and risk-sensitive learning}, in: \bibinfo{booktitle}{International Conference on Machine Learning}, \bibinfo{publisher}{PMLR}, \bibinfo{address}{Breckenridge, CO, USA}. pp. \bibinfo{pages}{1184--1193}.
%Type = Book
\bibitem[{Dubois and Prade(1988)}]{dubois1988possibility}
\bibinfo{author}{Dubois, D.}, \bibinfo{author}{Prade, H.}, \bibinfo{year}{1988}.
\newblock \bibinfo{title}{Possibility Theory}.
\newblock \bibinfo{publisher}{Plenum Press}.
%Type = Inproceedings
\bibitem[{Duell et~al.(2024)Duell, Seisenberger, Fu and Fan}]{duell2024quce}
\bibinfo{author}{Duell, J.}, \bibinfo{author}{Seisenberger, M.}, \bibinfo{author}{Fu, H.}, \bibinfo{author}{Fan, X.}, \bibinfo{year}{2024}.
\newblock \bibinfo{title}{{QUCE}: {The} minimisation and quantification of path-based uncertainty for generative counterfactual explanations}, in: \bibinfo{booktitle}{Proceedings of the 24th IEEE International Conference on Data Mining}, \bibinfo{publisher}{IEEE}, \bibinfo{address}{Piscataway, NJ, USA}. pp. \bibinfo{pages}{693--698}.
%Type = Inproceedings
\bibitem[{Dwork et~al.(2012)Dwork, Hardt, Pitassi, Reingold and Zemel}]{dwork2012fairness}
\bibinfo{author}{Dwork, C.}, \bibinfo{author}{Hardt, M.}, \bibinfo{author}{Pitassi, T.}, \bibinfo{author}{Reingold, O.}, \bibinfo{author}{Zemel, R.}, \bibinfo{year}{2012}.
\newblock \bibinfo{title}{Fairness through awareness}, in: \bibinfo{booktitle}{Proceedings of the 3rd Innovations in Theoretical Computer Science Conference}, \bibinfo{publisher}{ACM}, \bibinfo{address}{New York, NY, USA}. pp. \bibinfo{pages}{214--226}.
%Type = Article
\bibitem[{Freiesleben(2022)}]{freiesleben2022intriguing}
\bibinfo{author}{Freiesleben, T.}, \bibinfo{year}{2022}.
\newblock \bibinfo{title}{The intriguing relation between counterfactual explanations and adversarial examples}.
\newblock \bibinfo{journal}{Minds and Machines} \bibinfo{volume}{32}, \bibinfo{pages}{77--109}.
%Type = Article
\bibitem[{Gawlikowski et~al.(2023)Gawlikowski, Tassi, Ali, Lee, Humt, Feng, Kruspe, Triebel, Jung, Roscher, Shahzad, Yang, Bamler and Zhu}]{gawlikowski2023survey}
\bibinfo{author}{Gawlikowski, J.}, \bibinfo{author}{Tassi, C.R.N.}, \bibinfo{author}{Ali, M.}, \bibinfo{author}{Lee, J.}, \bibinfo{author}{Humt, M.}, \bibinfo{author}{Feng, J.}, \bibinfo{author}{Kruspe, A.}, \bibinfo{author}{Triebel, R.}, \bibinfo{author}{Jung, P.}, \bibinfo{author}{Roscher, R.}, \bibinfo{author}{Shahzad, M.}, \bibinfo{author}{Yang, W.}, \bibinfo{author}{Bamler, R.}, \bibinfo{author}{Zhu, X.X.}, \bibinfo{year}{2023}.
\newblock \bibinfo{title}{A survey of uncertainty in deep neural networks}.
\newblock \bibinfo{journal}{Artificial Intelligence Review} \bibinfo{volume}{56}, \bibinfo{pages}{1513--1589}.
%Type = Article
\bibitem[{Gigerenzer(2023)}]{gigerenzer2023psychological}
\bibinfo{author}{Gigerenzer, G.}, \bibinfo{year}{2023}.
\newblock \bibinfo{title}{Psychological {AI}: {Designing} algorithms informed by human psychology}.
\newblock \bibinfo{journal}{Perspectives on Psychological Science} \bibinfo{volume}{19}, \bibinfo{pages}{839--848}.
%Type = Inproceedings
\bibitem[{Goodfellow et~al.(2015)Goodfellow, Shlens and Szegedy}]{goodfellow2014explaining}
\bibinfo{author}{Goodfellow, I.J.}, \bibinfo{author}{Shlens, J.}, \bibinfo{author}{Szegedy, C.}, \bibinfo{year}{2015}.
\newblock \bibinfo{title}{Explaining and harnessing adversarial examples}, in: \bibinfo{booktitle}{Proceedings of the 3rd International Conference on Learning Representations}, \bibinfo{publisher}{Curran Associates, Inc.}, \bibinfo{address}{Red Hook, NY, USA}. pp. \bibinfo{pages}{1--11}.
%Type = Article
\bibitem[{Guidotti(2024)}]{guidotti2022counterfactual}
\bibinfo{author}{Guidotti, R.}, \bibinfo{year}{2024}.
\newblock \bibinfo{title}{Counterfactual explanations and how to find them: {Literature} review and benchmarking}.
\newblock \bibinfo{journal}{Data Mining and Knowledge Discovery} \bibinfo{volume}{38}, \bibinfo{pages}{2770--2824}.
%Type = Article
\bibitem[{H{\"u}llermeier and Waegeman(2021)}]{hullermeier2021aleatoric}
\bibinfo{author}{H{\"u}llermeier, E.}, \bibinfo{author}{Waegeman, W.}, \bibinfo{year}{2021}.
\newblock \bibinfo{title}{Aleatoric and epistemic uncertainty in machine learning: {An} introduction to concepts and methods}.
\newblock \bibinfo{journal}{Machine Learning} \bibinfo{volume}{110}, \bibinfo{pages}{457--506}.
%Type = Article
\bibitem[{Jin et~al.(2023)Jin, Li and Hamarneh}]{jin2023rethinking}
\bibinfo{author}{Jin, W.}, \bibinfo{author}{Li, X.}, \bibinfo{author}{Hamarneh, G.}, \bibinfo{year}{2023}.
\newblock \bibinfo{title}{Why is plausibility surprisingly problematic as an {XAI} criterion?}
\newblock \bibinfo{journal}{arXiv preprint arXiv:2303.17707} .
%Type = Inproceedings
\bibitem[{Kanamori et~al.(2024)Kanamori, Takagi, Kobayashi and Ike}]{kanamori2024learning}
\bibinfo{author}{Kanamori, K.}, \bibinfo{author}{Takagi, T.}, \bibinfo{author}{Kobayashi, K.}, \bibinfo{author}{Ike, Y.}, \bibinfo{year}{2024}.
\newblock \bibinfo{title}{Learning decision trees and forests with algorithmic recourse}, in: \bibinfo{booktitle}{International Conference on Machine Learning}, \bibinfo{publisher}{PMLR}, \bibinfo{address}{Breckenridge, CO, USA}. pp. \bibinfo{pages}{22936--22962}.
%Type = Article
\bibitem[{Karimi et~al.(2022)Karimi, Barthe, Sch{\"o}lkopf and Valera}]{karimi2022survey}
\bibinfo{author}{Karimi, A.H.}, \bibinfo{author}{Barthe, G.}, \bibinfo{author}{Sch{\"o}lkopf, B.}, \bibinfo{author}{Valera, I.}, \bibinfo{year}{2022}.
\newblock \bibinfo{title}{A survey of algorithmic recourse: {Contrastive} explanations and consequential recommendations}.
\newblock \bibinfo{journal}{ACM Computing Surveys} \bibinfo{volume}{55}, \bibinfo{pages}{1--29}.
%Type = Inproceedings
\bibitem[{Kingma and Ba(2015)}]{kingma2015adam}
\bibinfo{author}{Kingma, D.P.}, \bibinfo{author}{Ba, J.}, \bibinfo{year}{2015}.
\newblock \bibinfo{title}{Adam: {A} method for stochastic optimization}, in: \bibinfo{booktitle}{Proceedings of the 3rd International Conference on Learning Representations}, \bibinfo{publisher}{Curran Associates, Inc.}, \bibinfo{address}{Red Hook, NY, USA}. pp. \bibinfo{pages}{1--15}.
%Type = Inproceedings
\bibitem[{Lakshminarayanan et~al.(2017)Lakshminarayanan, Pritzel and Blundell}]{lakshminarayanan2017simple}
\bibinfo{author}{Lakshminarayanan, B.}, \bibinfo{author}{Pritzel, A.}, \bibinfo{author}{Blundell, C.}, \bibinfo{year}{2017}.
\newblock \bibinfo{title}{Simple and scalable predictive uncertainty estimation using deep ensembles}, in: \bibinfo{booktitle}{Advances in Neural Information Processing Systems}, \bibinfo{publisher}{Curran Associates, Inc.}, \bibinfo{address}{Red Hook, NY, USA}. pp. \bibinfo{pages}{6402--6413}.
%Type = Inproceedings
\bibitem[{Laugel et~al.(2018)Laugel, Lesot, Marsala, Renard and Detyniecki}]{laugel2018comparison}
\bibinfo{author}{Laugel, T.}, \bibinfo{author}{Lesot, M.J.}, \bibinfo{author}{Marsala, C.}, \bibinfo{author}{Renard, X.}, \bibinfo{author}{Detyniecki, M.}, \bibinfo{year}{2018}.
\newblock \bibinfo{title}{Comparison-based inverse classification for interpretability in machine learning}, in: \bibinfo{booktitle}{Proceedings of the International Conference on Information Processing and Management of Uncertainty in Knowledge-Based Systems}, \bibinfo{publisher}{Springer}, \bibinfo{address}{Cham, Switzerland}. pp. \bibinfo{pages}{100--111}.
%Type = Article
\bibitem[{de~Mathelin et~al.(2025)de~Mathelin, Deheeger, Mougeot and Vayatis}]{mathelin2023deep}
\bibinfo{author}{de~Mathelin, A.}, \bibinfo{author}{Deheeger, F.}, \bibinfo{author}{Mougeot, M.}, \bibinfo{author}{Vayatis, N.}, \bibinfo{year}{2025}.
\newblock \bibinfo{title}{Deep out-of-distribution uncertainty quantification via weight entropy maximization}.
\newblock \bibinfo{journal}{Journal of Machine Learning Research} \bibinfo{volume}{26}, \bibinfo{pages}{1--68}.
%Type = Article
\bibitem[{Mehdiyev et~al.(2025)Mehdiyev, Majlatow and Fettke}]{mehdiyev2025integrating}
\bibinfo{author}{Mehdiyev, N.}, \bibinfo{author}{Majlatow, M.}, \bibinfo{author}{Fettke, P.}, \bibinfo{year}{2025}.
\newblock \bibinfo{title}{Integrating permutation feature importance with conformal prediction for robust explainable artificial intelligence in predictive process monitoring}.
\newblock \bibinfo{journal}{Engineering Applications of Artificial Intelligence} \bibinfo{volume}{149}, \bibinfo{pages}{110363}.
%Type = Article
\bibitem[{Miller(2019)}]{miller2019explanation}
\bibinfo{author}{Miller, T.}, \bibinfo{year}{2019}.
\newblock \bibinfo{title}{Explanation in artificial intelligence: {Insights} from the social sciences}.
\newblock \bibinfo{journal}{Artificial Intelligence} \bibinfo{volume}{267}, \bibinfo{pages}{1--38}.
%Type = Inproceedings
\bibitem[{Miller(2023)}]{miller2023explainable}
\bibinfo{author}{Miller, T.}, \bibinfo{year}{2023}.
\newblock \bibinfo{title}{Explainable {AI} is dead, long live explainable {AI}! {Hypothesis}-driven decision support using evaluative {AI}}, in: \bibinfo{booktitle}{Proceedings of the ACM Conference on Fairness, Accountability, and Transparency}, \bibinfo{publisher}{ACM}, \bibinfo{address}{New York, NY, USA}. pp. \bibinfo{pages}{333--342}.
%Type = Inproceedings
\bibitem[{Mothilal et~al.(2020)Mothilal, Sharma and Tan}]{mothilal2020explaining}
\bibinfo{author}{Mothilal, R.K.}, \bibinfo{author}{Sharma, A.}, \bibinfo{author}{Tan, C.}, \bibinfo{year}{2020}.
\newblock \bibinfo{title}{Explaining machine learning classifiers through diverse counterfactual explanations}, in: \bibinfo{booktitle}{Proceedings of the ACM Conference on Fairness, Accountability, and Transparency}, \bibinfo{publisher}{ACM}, \bibinfo{address}{New York, NY, USA}. pp. \bibinfo{pages}{607--617}.
%Type = Inproceedings
\bibitem[{Paszke et~al.(2019)Paszke, Gross, Massa, Lerer, Bradbury, Chanan, Killeen, Lin, Gimelshein, Antiga, Desmaison, Kopf, Yang, DeVito, Raison, Tejani, Chilamkurthy, Steiner, Fang, Bai and Chintala}]{paszke2019pytorch}
\bibinfo{author}{Paszke, A.}, \bibinfo{author}{Gross, S.}, \bibinfo{author}{Massa, F.}, \bibinfo{author}{Lerer, A.}, \bibinfo{author}{Bradbury, J.}, \bibinfo{author}{Chanan, G.}, \bibinfo{author}{Killeen, T.}, \bibinfo{author}{Lin, Z.}, \bibinfo{author}{Gimelshein, N.}, \bibinfo{author}{Antiga, L.}, \bibinfo{author}{Desmaison, A.}, \bibinfo{author}{Kopf, A.}, \bibinfo{author}{Yang, E.}, \bibinfo{author}{DeVito, Z.}, \bibinfo{author}{Raison, M.}, \bibinfo{author}{Tejani, A.}, \bibinfo{author}{Chilamkurthy, S.}, \bibinfo{author}{Steiner, B.}, \bibinfo{author}{Fang, L.}, \bibinfo{author}{Bai, J.}, \bibinfo{author}{Chintala, S.}, \bibinfo{year}{2019}.
\newblock \bibinfo{title}{{PyTorch}: {An} imperative style, high-performance deep learning library}, in: \bibinfo{booktitle}{Advances in Neural Information Processing Systems}, \bibinfo{publisher}{Curran Associates, Inc.}, \bibinfo{address}{Red Hook, NY, USA}. pp. \bibinfo{pages}{8024--8035}.
%Type = Inproceedings
\bibitem[{Pawelczyk et~al.(2021)Pawelczyk, Bielawski, van~den Heuvel, Richter and Kasneci}]{pawelczyk2021carla}
\bibinfo{author}{Pawelczyk, M.}, \bibinfo{author}{Bielawski, S.}, \bibinfo{author}{van~den Heuvel, J.}, \bibinfo{author}{Richter, T.}, \bibinfo{author}{Kasneci, G.}, \bibinfo{year}{2021}.
\newblock \bibinfo{title}{{CARLA}: {A} {Python} library to benchmark algorithmic recourse and counterfactual explanation algorithms}, in: \bibinfo{booktitle}{Proceedings of the Neural Information Processing Systems Track on Datasets and Benchmarks}, \bibinfo{publisher}{Curran Associates, Inc.}, \bibinfo{address}{Red Hook, NY, USA}. pp. \bibinfo{pages}{1--13}.
%Type = Inproceedings
\bibitem[{Pawelczyk et~al.(2020)Pawelczyk, Broelemann and Kasneci}]{pawelczyk2020learning}
\bibinfo{author}{Pawelczyk, M.}, \bibinfo{author}{Broelemann, K.}, \bibinfo{author}{Kasneci, G.}, \bibinfo{year}{2020}.
\newblock \bibinfo{title}{Learning model-agnostic counterfactual explanations for tabular data}, in: \bibinfo{booktitle}{Proceedings of the ACM Web Conference}, \bibinfo{publisher}{ACM}, \bibinfo{address}{New York, NY, USA}. pp. \bibinfo{pages}{3126--3132}.
%Type = Inproceedings
\bibitem[{Perello-Nieto et~al.(2016)Perello-Nieto, de~Menezes~e Silva Filho Silva~Filho, Kull and Flach}]{perello2016background}
\bibinfo{author}{Perello-Nieto, M.}, \bibinfo{author}{de~Menezes~e Silva Filho Silva~Filho, T.}, \bibinfo{author}{Kull, M.}, \bibinfo{author}{Flach, P.}, \bibinfo{year}{2016}.
\newblock \bibinfo{title}{Background check: {A} general technique to build more reliable and versatile classifiers}, in: \bibinfo{booktitle}{Proceedings of the 16th IEEE International Conference on Data Mining}, \bibinfo{publisher}{IEEE}, \bibinfo{address}{Piscataway, NJ, USA}. pp. \bibinfo{pages}{1143--1148}.
%Type = Inproceedings
\bibitem[{Poyiadzi et~al.(2020)Poyiadzi, Sokol, Santos-Rodriguez, De~Bie and Flach}]{poyiadzi2020face}
\bibinfo{author}{Poyiadzi, R.}, \bibinfo{author}{Sokol, K.}, \bibinfo{author}{Santos-Rodriguez, R.}, \bibinfo{author}{De~Bie, T.}, \bibinfo{author}{Flach, P.}, \bibinfo{year}{2020}.
\newblock \bibinfo{title}{{FACE}: {Feasible} and actionable counterfactual explanations}, in: \bibinfo{booktitle}{Proceedings of the AAAI/ACM Conference on AI, Ethics, and Society}, \bibinfo{publisher}{ACM}, \bibinfo{address}{New York, NY, USA}. pp. \bibinfo{pages}{344--350}.
%Type = Inproceedings
\bibitem[{Romashov et~al.(2022)Romashov, Gjoreski, Sokol, Martinez and Langheinrich}]{romashov2022baycon}
\bibinfo{author}{Romashov, P.}, \bibinfo{author}{Gjoreski, M.}, \bibinfo{author}{Sokol, K.}, \bibinfo{author}{Martinez, M.V.}, \bibinfo{author}{Langheinrich, M.}, \bibinfo{year}{2022}.
\newblock \bibinfo{title}{{BayCon}: {Model}-agnostic {Bayesian} counterfactual generator}, in: \bibinfo{booktitle}{Proceedings of the 31st International Joint Conference on Artificial Intelligence}, \bibinfo{publisher}{IJCAI}, \bibinfo{address}{Vienna, Austria}. pp. \bibinfo{pages}{740--746}.
%Type = Article
\bibitem[{Rudin(2019)}]{rudin2019stop}
\bibinfo{author}{Rudin, C.}, \bibinfo{year}{2019}.
\newblock \bibinfo{title}{Stop explaining black box machine learning models for high stakes decisions and use interpretable models instead}.
\newblock \bibinfo{journal}{Nature Machine Intelligence} \bibinfo{volume}{1}, \bibinfo{pages}{206--215}.
%Type = Article
\bibitem[{Rudin et~al.(2022)Rudin, Chen, Chen, Huang, Semenova and Zhong}]{rudin2022interpretable}
\bibinfo{author}{Rudin, C.}, \bibinfo{author}{Chen, C.}, \bibinfo{author}{Chen, Z.}, \bibinfo{author}{Huang, H.}, \bibinfo{author}{Semenova, L.}, \bibinfo{author}{Zhong, C.}, \bibinfo{year}{2022}.
\newblock \bibinfo{title}{Interpretable machine learning: {Fundamental} principles and 10 grand challenges}.
\newblock \bibinfo{journal}{Statistic Surveys} \bibinfo{volume}{16}, \bibinfo{pages}{1--85}.
%Type = Inproceedings
\bibitem[{Rudin et~al.(2024)Rudin, Zhong, Semenova, Seltzer, Parr, Liu, Katta, Donnelly, Chen and Boner}]{rudin2024amazing}
\bibinfo{author}{Rudin, C.}, \bibinfo{author}{Zhong, C.}, \bibinfo{author}{Semenova, L.}, \bibinfo{author}{Seltzer, M.}, \bibinfo{author}{Parr, R.}, \bibinfo{author}{Liu, J.}, \bibinfo{author}{Katta, S.}, \bibinfo{author}{Donnelly, J.}, \bibinfo{author}{Chen, H.}, \bibinfo{author}{Boner, Z.}, \bibinfo{year}{2024}.
\newblock \bibinfo{title}{Position: {Amazing} things come from having many good models}, in: \bibinfo{booktitle}{International Conference on Machine Learning}, \bibinfo{publisher}{PMLR}, \bibinfo{address}{Breckenridge, CO, USA}. pp. \bibinfo{pages}{42783--42795}.
%Type = Article
\bibitem[{Ryan(2020)}]{ryan2020ai}
\bibinfo{author}{Ryan, M.}, \bibinfo{year}{2020}.
\newblock \bibinfo{title}{In {AI} we trust: {Ethics}, artificial intelligence, and reliability}.
\newblock \bibinfo{journal}{Science and Engineering Ethics} \bibinfo{volume}{26}, \bibinfo{pages}{2749--2767}.
%Type = Article
\bibitem[{Salvi et~al.(2025)Salvi, Seoni, Campagner, Gertych, Acharya, Molinari and Cabitza}]{salvi2025explainability}
\bibinfo{author}{Salvi, M.}, \bibinfo{author}{Seoni, S.}, \bibinfo{author}{Campagner, A.}, \bibinfo{author}{Gertych, A.}, \bibinfo{author}{Acharya, U.R.}, \bibinfo{author}{Molinari, F.}, \bibinfo{author}{Cabitza, F.}, \bibinfo{year}{2025}.
\newblock \bibinfo{title}{Explainability and uncertainty: {Two} sides of the same coin for enhancing the interpretability of deep learning models in healthcare}.
\newblock \bibinfo{journal}{International Journal of Medical Informatics} \bibinfo{volume}{197}, \bibinfo{pages}{105846}.
%Type = Inproceedings
\bibitem[{Schut et~al.(2021)Schut, Key, Mc~Grath, Costabello, Sacaleanu and Gal}]{schut2021generating}
\bibinfo{author}{Schut, L.}, \bibinfo{author}{Key, O.}, \bibinfo{author}{Mc~Grath, R.}, \bibinfo{author}{Costabello, L.}, \bibinfo{author}{Sacaleanu, B.}, \bibinfo{author}{Gal, Y.}, \bibinfo{year}{2021}.
\newblock \bibinfo{title}{Generating interpretable counterfactual explanations by implicit minimisation of epistemic and aleatoric uncertainties}, in: \bibinfo{booktitle}{Proceedings of the 24th International Conference on Artificial Intelligence and Statistics}, \bibinfo{publisher}{PMLR}, \bibinfo{address}{Breckenridge, CO, USA}. pp. \bibinfo{pages}{1756--1764}.
%Type = Book
\bibitem[{Shafer(1976)}]{shafer1976mathematical}
\bibinfo{author}{Shafer, G.}, \bibinfo{year}{1976}.
\newblock \bibinfo{title}{A Mathematical Theory of Evidence}.
\newblock \bibinfo{publisher}{Princeton University Press}.
%Type = Article
\bibitem[{Shwartz-Ziv and Armon(2022)}]{shwartz2022tabular}
\bibinfo{author}{Shwartz-Ziv, R.}, \bibinfo{author}{Armon, A.}, \bibinfo{year}{2022}.
\newblock \bibinfo{title}{Tabular data: {Deep} learning is not all you need}.
\newblock \bibinfo{journal}{Information Fusion} \bibinfo{volume}{81}, \bibinfo{pages}{84--90}.
%Type = Article
\bibitem[{Silva~Filho et~al.(2023)Silva~Filho, Song, Perello-Nieto, Santos-Rodriguez, Kull and Flach}]{silva2023classifier}
\bibinfo{author}{Silva~Filho, T.}, \bibinfo{author}{Song, H.}, \bibinfo{author}{Perello-Nieto, M.}, \bibinfo{author}{Santos-Rodriguez, R.}, \bibinfo{author}{Kull, M.}, \bibinfo{author}{Flach, P.}, \bibinfo{year}{2023}.
\newblock \bibinfo{title}{Classifier calibration: {A} survey on how to assess and improve predicted class probabilities}.
\newblock \bibinfo{journal}{Machine Learning} \bibinfo{volume}{112}, \bibinfo{pages}{3211--3260}.
%Type = Article
\bibitem[{Small et~al.(2025)Small, Clark, McWilliams, Ambler, Sokol, Chan, Salim and Santos-Rodriguez}]{small2023counterfactual}
\bibinfo{author}{Small, E.A.}, \bibinfo{author}{Clark, J.N.}, \bibinfo{author}{McWilliams, C.J.}, \bibinfo{author}{Ambler, M.}, \bibinfo{author}{Sokol, K.}, \bibinfo{author}{Chan, J.}, \bibinfo{author}{Salim, F.D.}, \bibinfo{author}{Santos-Rodriguez, R.}, \bibinfo{year}{2025}.
\newblock \bibinfo{title}{Counterfactual explanations via locally-guided sequential algorithmic recourse}.
\newblock \bibinfo{journal}{ACM Transactions on Intelligent Systems and Technology} .
%Type = Article
\bibitem[{Smets and Kennes(1994)}]{smets1994transferable}
\bibinfo{author}{Smets, P.}, \bibinfo{author}{Kennes, R.}, \bibinfo{year}{1994}.
\newblock \bibinfo{title}{The transferable belief model}.
\newblock \bibinfo{journal}{Artificial Intelligence} \bibinfo{volume}{66}, \bibinfo{pages}{191--234}.
%Type = Article
\bibitem[{Sokol et~al.(2025a)Sokol, Fackler and Vogt}]{sokol2025artificial}
\bibinfo{author}{Sokol, K.}, \bibinfo{author}{Fackler, J.}, \bibinfo{author}{Vogt, J.E.}, \bibinfo{year}{2025}a.
\newblock \bibinfo{title}{Artificial intelligence should genuinely support clinical reasoning and decision making to bridge the translational gap}.
\newblock \bibinfo{journal}{npj Digital Medicine} \bibinfo{volume}{8}, \bibinfo{pages}{345}.
%Type = Inproceedings
\bibitem[{Sokol and Flach(2020a)}]{sokol2020explainability}
\bibinfo{author}{Sokol, K.}, \bibinfo{author}{Flach, P.}, \bibinfo{year}{2020}a.
\newblock \bibinfo{title}{Explainability fact sheets: {A} framework for systematic assessment of explainable approaches}, in: \bibinfo{booktitle}{Proceedings of the ACM Conference on Fairness, Accountability, and Transparency}, \bibinfo{publisher}{ACM}, \bibinfo{address}{New York, NY, USA}. pp. \bibinfo{pages}{56--67}.
%Type = Article
\bibitem[{Sokol and Flach(2020b)}]{sokol2020one}
\bibinfo{author}{Sokol, K.}, \bibinfo{author}{Flach, P.}, \bibinfo{year}{2020}b.
\newblock \bibinfo{title}{One explanation does not fit all: {The} promise of interactive explanations for machine learning transparency}.
\newblock \bibinfo{journal}{KI-K{\"u}nstliche Intelligenz} \bibinfo{volume}{34}, \bibinfo{pages}{235--250}.
%Type = Article
\bibitem[{Sokol and Flach(2025)}]{sokol2020limetree}
\bibinfo{author}{Sokol, K.}, \bibinfo{author}{Flach, P.}, \bibinfo{year}{2025}.
\newblock \bibinfo{title}{{LIMEtree}: {Consistent} and faithful surrogate explanations of multiple classes}.
\newblock \bibinfo{journal}{Electronics} \bibinfo{volume}{14}, \bibinfo{pages}{929}.
%Type = Article
\bibitem[{Sokol et~al.(2022)Sokol, Hepburn, Santos-Rodriguez and Flach}]{sokol2022what}
\bibinfo{author}{Sokol, K.}, \bibinfo{author}{Hepburn, A.}, \bibinfo{author}{Santos-Rodriguez, R.}, \bibinfo{author}{Flach, P.}, \bibinfo{year}{2022}.
\newblock \bibinfo{title}{What and how of machine learning transparency: {Building} bespoke explainability tools with interoperable algorithmic components}.
\newblock \bibinfo{journal}{Journal of Open Source Education} \bibinfo{volume}{5}, \bibinfo{pages}{175}.
%Type = Article
\bibitem[{Sokol et~al.(2024)Sokol, Kull, Chan and Salim}]{sokol2024cross}
\bibinfo{author}{Sokol, K.}, \bibinfo{author}{Kull, M.}, \bibinfo{author}{Chan, J.}, \bibinfo{author}{Salim, F.}, \bibinfo{year}{2024}.
\newblock \bibinfo{title}{Cross-model fairness: {Empirical} study of fairness and ethics under model multiplicity}.
\newblock \bibinfo{journal}{ACM Journal on Responsible Computing} \bibinfo{volume}{1}, \bibinfo{pages}{1--27}.
%Type = Article
\bibitem[{Sokol et~al.(2025b)Sokol, Small and Xuan}]{sokol2023navigating}
\bibinfo{author}{Sokol, K.}, \bibinfo{author}{Small, E.}, \bibinfo{author}{Xuan, Y.}, \bibinfo{year}{2025}b.
\newblock \bibinfo{title}{Navigating explanatory multiverse through counterfactual path geometry}.
\newblock \bibinfo{journal}{Machine Learning} \bibinfo{volume}{114}, \bibinfo{pages}{1--33}.
%Type = Article
\bibitem[{Sokol and Vogt(2023)}]{sokol2023reasonable}
\bibinfo{author}{Sokol, K.}, \bibinfo{author}{Vogt, J.E.}, \bibinfo{year}{2023}.
\newblock \bibinfo{title}{{(Un)reasonable} allure of ante-hoc interpretability for high-stakes domains: {Transparency} is necessary but insufficient for comprehensibility}.
\newblock \bibinfo{journal}{ICML Workshop on Interpretable Machine Learning in Healthcare} .
%Type = Inproceedings
\bibitem[{Sokol and Vogt(2024)}]{sokol2024what}
\bibinfo{author}{Sokol, K.}, \bibinfo{author}{Vogt, J.E.}, \bibinfo{year}{2024}.
\newblock \bibinfo{title}{What does evaluation of explainable artificial intelligence actually tell us? {A} case for compositional and contextual validation of {XAI} building blocks}, in: \bibinfo{booktitle}{Extended Abstracts of the ACM CHI Conference on Human Factors in Computing Systems}, \bibinfo{publisher}{ACM}, \bibinfo{address}{New York, NY, USA}. pp. \bibinfo{pages}{1--8}.
%Type = Inproceedings
\bibitem[{Teufel et~al.(2026)Teufel, Leinweber and Friederich}]{teufel2025improving}
\bibinfo{author}{Teufel, J.}, \bibinfo{author}{Leinweber, A.}, \bibinfo{author}{Friederich, P.}, \bibinfo{year}{2026}.
\newblock \bibinfo{title}{Improving counterfactual truthfulness for molecular property prediction through uncertainty quantification}, in: \bibinfo{booktitle}{Proceedings of the 3rd World Conference on Explainable Artificial Intelligence}, \bibinfo{publisher}{Springer}, \bibinfo{address}{Cham, Switzerland}. pp. \bibinfo{pages}{317--339}.
%Type = Article
\bibitem[{Thiagarajan et~al.(2022)Thiagarajan, Thopalli, Rajan and Turaga}]{thiagarajan2022training}
\bibinfo{author}{Thiagarajan, J.J.}, \bibinfo{author}{Thopalli, K.}, \bibinfo{author}{Rajan, D.}, \bibinfo{author}{Turaga, P.}, \bibinfo{year}{2022}.
\newblock \bibinfo{title}{Training calibration-based counterfactual explainers for deep learning models in medical image analysis}.
\newblock \bibinfo{journal}{Scientific Reports} \bibinfo{volume}{12}, \bibinfo{pages}{597}.
%Type = Article
\bibitem[{Tomsett et~al.(2020)Tomsett, Preece, Braines, Cerutti, Chakraborty, Srivastava, Pearson and Kaplan}]{tomsett2020rapid}
\bibinfo{author}{Tomsett, R.}, \bibinfo{author}{Preece, A.}, \bibinfo{author}{Braines, D.}, \bibinfo{author}{Cerutti, F.}, \bibinfo{author}{Chakraborty, S.}, \bibinfo{author}{Srivastava, M.}, \bibinfo{author}{Pearson, G.}, \bibinfo{author}{Kaplan, L.}, \bibinfo{year}{2020}.
\newblock \bibinfo{title}{Rapid trust calibration through interpretable and uncertainty-aware {AI}}.
\newblock \bibinfo{journal}{Patterns} \bibinfo{volume}{1}, \bibinfo{pages}{100049}.
%Type = Article
\bibitem[{Wachter et~al.(2017)Wachter, Mittelstadt and Russell}]{wachter2017counterfactual}
\bibinfo{author}{Wachter, S.}, \bibinfo{author}{Mittelstadt, B.}, \bibinfo{author}{Russell, C.}, \bibinfo{year}{2017}.
\newblock \bibinfo{title}{Counterfactual explanations without opening the black box: {Automated} decisions and the {GDPR}}.
\newblock \bibinfo{journal}{Harvard Journal of Law \& Technology} \bibinfo{volume}{31}, \bibinfo{pages}{841}.
%Type = Book
\bibitem[{Walley(1991)}]{walley1991statistical}
\bibinfo{author}{Walley, P.}, \bibinfo{year}{1991}.
\newblock \bibinfo{title}{Statistical Reasoning with Imprecise Probabilities}.
\newblock Monographs on Statistics \& Applied Probability, \bibinfo{publisher}{Chapman \& Hall}.
%Type = Article
\bibitem[{Wang and Lin(2021)}]{wang2021hybrid}
\bibinfo{author}{Wang, T.}, \bibinfo{author}{Lin, Q.}, \bibinfo{year}{2021}.
\newblock \bibinfo{title}{Hybrid predictive models: {When} an interpretable model collaborates with a black-box model}.
\newblock \bibinfo{journal}{Journal of Machine Learning Research} \bibinfo{volume}{22}, \bibinfo{pages}{1--38}.
%Type = Inproceedings
\bibitem[{Wielopolski et~al.(2024)Wielopolski, Furman, Stefanowski and Zi{\k{e}}ba}]{wielopolski2024probabilistically}
\bibinfo{author}{Wielopolski, P.}, \bibinfo{author}{Furman, O.}, \bibinfo{author}{Stefanowski, J.}, \bibinfo{author}{Zi{\k{e}}ba, M.}, \bibinfo{year}{2024}.
\newblock \bibinfo{title}{Probabilistically plausible counterfactual explanations with normalizing flows}, in: \bibinfo{booktitle}{Proceedings of the 27th European Conference on Artificial Intelligence}, \bibinfo{publisher}{IOS Press}, \bibinfo{address}{Amsterdam, The Netherlands}. pp. \bibinfo{pages}{954--961}.
%Type = Article
\bibitem[{Wood et~al.(2024)Wood, Papamarkou, Benatan and Allmendinger}]{wood2024model}
\bibinfo{author}{Wood, D.}, \bibinfo{author}{Papamarkou, T.}, \bibinfo{author}{Benatan, M.}, \bibinfo{author}{Allmendinger, R.}, \bibinfo{year}{2024}.
\newblock \bibinfo{title}{Model-agnostic variable importance for predictive uncertainty: {An} entropy-based approach}.
\newblock \bibinfo{journal}{Data Mining and Knowledge Discovery} \bibinfo{volume}{38}, \bibinfo{pages}{4184--4216}.
%Type = Article
\bibitem[{Xuan et~al.(2025)Xuan, Sokol, Sanderson and Chan}]{xuan2024perfect}
\bibinfo{author}{Xuan, Y.}, \bibinfo{author}{Sokol, K.}, \bibinfo{author}{Sanderson, M.}, \bibinfo{author}{Chan, J.}, \bibinfo{year}{2025}.
\newblock \bibinfo{title}{Perfect counterfactuals in imperfect worlds: {Modelling} noisy implementation of actions in sequential algorithmic recourse}.
\newblock \bibinfo{journal}{Machine Learning} \bibinfo{volume}{114}, \bibinfo{pages}{187}.

\end{thebibliography}

\end{document}